\pdfoutput=1
\documentclass{article}

\usepackage{iclr2027_conference,times}

\usepackage{amsmath,amsfonts,bm}

\def\eqref#1{equation~\ref{#1}}
\def\Eqref#1{Equation~\ref{#1}}

\def\1{\bm{1}}

\DeclareMathAlphabet{\mathsfit}{\encodingdefault}{\sfdefault}{m}{sl}
\SetMathAlphabet{\mathsfit}{bold}{\encodingdefault}{\sfdefault}{bx}{n}

\newcommand{\E}{\mathbb{E}}

\newcommand{\Var}{\mathrm{Var}}

\newcommand{\Cov}{\mathrm{Cov}}

\DeclareMathOperator*{\argmax}{arg\,max}

\usepackage{amssymb}
\usepackage{amsthm}
\usepackage{array}
\usepackage{booktabs}
\usepackage{longtable}
\usepackage{float}
\usepackage{graphicx}
\usepackage{hyperref}
\usepackage{url}

\newcommand{\Normal}{\mathcal{N}}
\newcommand{\method}{\textsc{RHOPE}}
\newcommand{\delayedmethod}{\textsc{RHOPE-Delayed}}
\newcommand{\sem}[1]{\,{\scriptsize$\pm$#1}}
\newcommand{\mathmethod}{\mathrm{RHOPE}}
\newcommand{\mathdelayedmethod}{\mathrm{RHOPE\text{-}Delayed}}
\newcommand{\RHO}{\mathrm{RHO}}

\newcommand{\diag}{\operatorname{diag}}
\newcommand{\res}{\mathrm{res}}
\newcommand{\ind}{\mathbf{1}}
\newcommand{\Fcal}{\mathcal{F}}
\newcommand{\DeltaK}{\Delta_K}
\newcolumntype{L}[1]{>{\raggedright\arraybackslash}p{#1}}

\newtheorem{theorem}{Theorem}
\newtheorem{definition}{Definition}
\newtheorem{assumption}{Assumption}
\newtheorem{lemma}{Lemma}
\newtheorem{proposition}{Proposition}

\newcounter{algorithm}
\renewcommand{\thealgorithm}{\arabic{algorithm}}
\newenvironment{algorithmbox}[2]{%
  \refstepcounter{algorithm}%
  \par\medskip
  \begin{center}
  \begin{minipage}{0.96\linewidth}
  \hrule
  \vspace{3pt}
  \textbf{Algorithm \thealgorithm: #2}\label{#1}
  \par\smallskip\hrule\smallskip
}{%
  \par\smallskip\hrule
  \end{minipage}
  \end{center}
}

\title{RHOPE: LLM-Augmented\\Adaptive Experimental Design with\\Complex Contextual Information}

\iclrfinalcopy

\author{
Bowen Shi\\
Tsinghua University\\
\texttt{sbw22@mails.tsinghua.edu.cn}
\And
Bo Li\\
Tsinghua University\\
\texttt{libo@sem.tsinghua.edu.cn}
}

\begin{document}

\maketitle
\lhead{Preprint}

\begin{abstract}
Experimental design allocates limited measurement effort across candidate
treatments to learn their expected outcomes. Adaptive designs refine
allocations as evidence accumulates, improving estimation and treatment
selection. Contextual information can inform these decisions, but exploiting
unstructured or high-dimensional context
remains challenging. We use large language models (LLMs) to transform complex
context into proxy outcomes and propose
Residual Horizon Optimization for Prediction-Powered Experimentation
(\method{}), a principled Bayesian adaptive experimentation framework that
combines true outcomes with LLM-generated proxy outcomes and formulates
finite-batch design as a posterior Markov decision process (MDP).
Our methodology uses proxy outcomes in two ways: sample augmentation improves
estimation precision, while observing proxies before each allocation decision
provides advance information.
Synthetic experiments and semi-synthetic experiments using Genomics of Drug
Sensitivity in Cancer (GDSC) data suggest that estimation precision drives most gains, with smaller
advance-information gains, and highlight the importance of proxy
informativeness and moment-estimation accuracy.
\end{abstract}

\section{Introduction}
\label{sec:introduction}

Experimental design allocates experimental units among candidate treatments
to estimate their mean outcomes under resource constraints. Unlike static
designs, adaptive designs revise allocations as outcomes accumulate, improving
estimation and treatment selection within a fixed budget. Adaptive experimental design is used in
clinical trials, online platforms, recommender systems, and scientific screening
\citep{villar2015multi,burtini2015survey,wu2018learning,perez2020adaptivebandit,liang2023experimental}.
We study fixed-budget simple-regret best-arm identification (BAI):
researchers sequentially collect data to identify the arm with the highest
expected outcome. A fixed budget reflects a predetermined sample size or
horizon; simple regret measures the terminal recommendation's expected-outcome
shortfall, not merely whether the best arm is identified
\citep{audibert2010best,bubeck2011pure,russo2016simple,
zhao2023revisiting,komiyama2024rateoptimal}.

Prior work on fixed-budget simple-regret BAI has developed methods from both
theoretical and algorithmic perspectives
\citep{che2023adaptive,che2024optimization,komiyama2024rateoptimal,
kato2025minimax}. The broader BAI literature considers objectives and settings
beyond fixed-budget simple regret
\citep{audibert2010best,karnin2013almost,russo2016simple,jourdan2022top,
kato2022smallgap}. Despite this substantial literature, two main issues remain
insufficiently discussed.

The first issue concerns context. Contextual information explains part of
the variation in a unit's outcome and can inform experimental design \citep{kato2021contextual},
but its use remains understudied in fixed-budget simple-regret BAI.
Other BAI-related work estimates conditional moments of potential outcomes
from experimental data \citep{kato2020efficient,kato2022smallgap}.
Under limited budgets, this approach scales poorly and is brittle with
unstructured or high-dimensional context, such as text, images, genomic
profiles, and multimodal records
\citep{yang2013gdsc,rindtorff2019precision,dhawan2024natural}.

The second issue is objective mismatch. Most of the BAI literature relies on
heuristic or mimic algorithms \citep{karnin2013almost,russo2016simple,
jourdan2022top,kato2022smallgap}, whose allocation decisions are driven by
surrogate objectives based on confidence bounds, posterior misidentification
probabilities, or asymptotic error bounds. Our final objective, by contrast,
is simple regret. Even when these algorithms attain optimal regret or error
rates, unfavorable constants can impair finite-budget performance. In a
one-armed bandit experiment, \citet{adusumilli2021risk} shows that Thompson
sampling, as a heuristic algorithm, can incur twice the optimal regret despite
having a favorable rate. Thus, optimizing a surrogate objective can weaken
adaptive experimental design in practice.

To address these issues, we propose Residual Horizon Optimization for
Prediction-Powered Experimentation (\method{}). Inspired by prediction-powered inference (PPI)
\citep{angelopoulos2023ppi,angelopoulos2023ppiplus} and its Bayesian
counterpart \citep{hofer2024bayesian}, it uses LLMs as general-purpose
generative models to turn rich context into auxiliary proxy outcomes,
avoiding conditional-moment estimation from scarce experimental data.
Knowledge encoded in LLMs can also support cold-start experiments.
Building on \citet{che2023adaptive}, \method{} formulates a posterior
Markov decision process (MDP) and applies residual horizon optimization (RHO) to directly optimize simple
regret. Proxies contribute information for arm-mean estimation and an early
look at upcoming-batch outcomes before allocation.

\noindent\textbf{Contributions.} Our contributions are threefold.
\textit{First,} we introduce an adaptive experimentation framework for
fixed-budget simple-regret BAI that uses LLM-generated proxy outcomes
without requiring a high-dimensional response model to be estimated from
scarce experimental data.
\textit{Second,} we establish value comparisons for proxy-based estimation
and advance information under the Gaussian sequential experiment and
quantify both gaps: the estimation gain removes a fixed fraction of the
regret guarantee that grows with the proxy--outcome correlation, whereas the
advance-information gain decays over the horizon. Experiments confirm that
improved arm-mean estimation drives most gains, with a small additional
benefit from advance information. When moments are estimated, the gain
persists only if the proxy--outcome correlation exceeds a threshold that
shrinks with the estimation sample size.
\textit{Third,} we show that, with known moments and a flat prior, the
finite-batch working-posterior mean for each arm has an exact PPI++ representation
\citep{angelopoulos2023ppi,angelopoulos2023ppiplus}. This algebraic connection
applies arm by arm in a setting with multiple potential outcomes
\citep{debartolomeis2025efficient,shankar2025augmenting}, connecting PPI++
to adaptive experimental design.

\section{Related Work}
\label{sec:related}

\paragraph{Adaptive experimental design and bandits.}
Adaptive experimental design is often modeled as a multi-armed bandit problem.
Depending on the objective, adaptive designs can be broadly categorized into
variance-minimizing designs \citep{neyman1934optimal,zhao2024adaptive,
kato2020efficient}, cumulative-regret bandits \citep{lai1985asymptotically,
auer2002finite,russo2018tutorial,lattimore2020bandit}, and pure-exploration
problems such as BAI \citep{audibert2010best,
bubeck2011pure,russo2016simple,kato2021contextual}. A recent line of work
instead uses diffusion limits to directly optimize experiment-level risk
\citep{kuang2021weak,fan2021diffusion,adusumilli2021risk,che2023adaptive}.

\paragraph{Best-arm identification.}
BAI has fixed-confidence
\citep{garivier2016optimal,kaufmann2016complexity,kato2021contextual} and
fixed-budget settings. Within the fixed-budget setting, common objectives are
the probability of error \citep{audibert2010best,karnin2013almost,
carpentier2016tight,kato2022smallgap} and simple regret
\citep{audibert2010best,bubeck2011pure,russo2016simple,zhao2023revisiting,
kato2025minimax}.

\paragraph{Prediction-powered inference.}
PPI uses predictions from machine learning
systems to improve statistical inference \citep{angelopoulos2023ppi}.
Subsequent work develops power-tuned variants and multi-predictor
extensions \citep{angelopoulos2023ppiplus,ji2025surrogates,miao2025postprediction,
shan2025sada,carlson2025unifying,xu2025semisupervised,kluger2025imputed,
vanderlaan2026calibeating}. Bayesian PPI and Bayes-assisted frequentist PPI
provide Bayesian and hybrid formulations of prediction-powered inference,
respectively \citep{hofer2024bayesian,cortinovis2025fabppi}. Recent studies
extend PPI ideas to causal inference
\citep{debartolomeis2025efficient,shankar2025augmenting} and to adaptive
experimentation \citep{ji2025surrogate,ma2026proxy,ao2026textual}.

\section{Problem Setup}
\label{sec:setup}

Following the setup of BAI in multi-armed bandits, we seek the highest-mean arm among
\([K]:=\{1,\ldots,K\}\) over \(T\) batches. Batch \(t=0,\ldots,T-1\) contains
\(b_tn\) units, where \(n\) is the total \mbox{budget}, \(b_t>0\) is fixed and
known, \(\sum_{t=0}^{T-1}b_t=1\), and \(b_tn\) is an integer.
We focus on this finite-batch structure for two reasons. First, updating the
allocation policy entails nontrivial operational costs and engineering
difficulties, particularly in medical and platform experiments; consequently,
the number of reallocations is often finite
\citep{bakshy2018ae,chow2008adaptive}.
Second, under the finite-batch structure, each batch's outcome aggregates can
be approximated by Gaussian signals without imposing parametric assumptions on
individual-outcome distributions. This representation underpins the MDP
formulation in Section~\ref{sec:method}.

For unit \(j\) in batch \(t\), context \(X_{t,j}\) may include
high-dimensional covariates and unstructured information such as images and
text. Let \(Y_{t,j,a}\) be its true potential outcome under arm \(a\).
Querying an LLM with \(X_{t,j}\) and arm information produces the proxy
\(\tilde Y_{t,j,a}\). We model these outcomes as
\begin{align}
Y_{t,j,a}&=h_a/\sqrt{n}+\epsilon_{t,j,a},
\qquad
\tilde Y_{t,j,a}=\tilde h_a/\sqrt{n}+\tilde\epsilon_{t,j,a}.
\end{align}
For the centered noise terms \(\epsilon_{t,j,a}\) and
\(\tilde\epsilon_{t,j,a}\), we define
\(\Var(\epsilon_{t,j,a})=s_a^2\),
\(\Var(\tilde\epsilon_{t,j,a})=\tilde s_a^2\), and
\(\Cov(\epsilon_{t,j,a},\tilde\epsilon_{t,j,a})
=\varrho_a s_a\tilde s_a\), where \(\varrho_a\) is the arm-specific
correlation between the true outcome and the proxy outcome. We assume
\(s_a^2>0\), \(\tilde s_a^2>0\), and \(|\varrho_a|<1\) for every arm
\(a\), excluding degenerate noise distributions. The
\(1/\sqrt n\) local scaling and the treatment of variances and
covariances as known are standard in diffusion-limit analyses
\citep{kuang2021weak,adusumilli2021risk}. In practice, unknown outcome variances
and covariances can instead be estimated from an initial batch; we discuss
the impact of estimation error through the ablation experiments in
Appendix~\ref{app:unknown-moments}.

At the beginning of batch \(t\), the researcher observes all contexts in the
batch and generates \(\tilde Y_{t,j,a}\) from \(X_{t,j}\) for every unit--arm
pair. After observing these proxy outcomes, the researcher chooses an
allocation policy
\(\pi_t=(\pi_{t,1},\ldots,\pi_{t,K})\in\DeltaK\), where
\(\DeltaK:=\{p\in[0,1]^K:\sum_a p_a=1\}\), and assigns units in accordance
with it.
Let \(A_{t,j}\) denote the treatment arm assigned to unit \(j\); only the
corresponding true potential outcome \(Y_{t,j,A_{t,j}}\) is observed.

Throughout, we adopt a Bayesian perspective and treat the unknown local
parameters \(h=(h_1,\ldots,h_K)\) and
\(\tilde h=(\tilde h_1,\ldots,\tilde h_K)\) as random, with joint prior
\((h,\tilde h)\sim\nu\). Let
\(a^*(h)\in\argmax_a h_a\) be the best arm at the local scale, let
\(\hat a_T\) be the terminal recommendation, and let
\(\pi=(\pi_0,\ldots,\pi_{T-1})\) denote the complete adaptive allocation
policy. Our final optimization target is Bayes simple regret,
\begin{align}
\label{eq:bayes-simple-regret}
\mathfrak R_T(\pi,\nu)
&=
\E^\pi_{(h,\tilde h)\sim\nu}
\left[h_{a^*(h)}-h_{\hat a_T}\right].
\end{align}
Intuitively, the proposed algorithm should therefore perform well on average
over problem instances drawn from this joint prior.

\section{LLM-Augmented Adaptive Experimental Design}
\label{sec:method}

We combine orthogonal mean estimation, a Gaussian posterior MDP, and RHO
to obtain tractable batch-level allocation policies.

\subsection{Orthogonal Decomposition for Mean Estimation}
\label{subsec:orthogonal}

For each arm \(a\), we estimate \(h_a\) and \(\tilde h_a\)
through the orthogonal decomposition
\begin{align}
\theta_a^{(1)}&:=h_a-\lambda_a\tilde h_a,\qquad
\theta_a^{(2)}:=\tilde h_a.
\end{align}
The arm-wise projection coefficient is
\(\lambda_a:=\Cov(\epsilon_{t,j,a},\tilde\epsilon_{t,j,a})/
\Var(\tilde\epsilon_{t,j,a})
=\varrho_a s_a/\tilde s_a\).
Since \(h_a=\theta_a^{(1)}+\lambda_a\theta_a^{(2)}\) and
\(\tilde h_a=\theta_a^{(2)}\), these transformed parameters determine the
original means and are the targets of our estimation and posterior updates.

To estimate \(\theta_a^{(1)}\) and \(\theta_a^{(2)}\), define
\(\xi_{t,j,a}:=\ind\{A_{t,j}=a\}\). The corresponding aggregate estimators are
\begin{align}
\hat J_{t,a}
&:=
\frac1{b_t\sqrt n}\sum_{j=1}^{b_tn}\tilde Y_{t,j,a},
\qquad
\hat G_{t,a}
:=
\frac1{b_t\sqrt n}
\sum_{j=1}^{b_tn}\xi_{t,j,a}
\left(Y_{t,j,a}-\lambda_a\tilde Y_{t,j,a}\right).
\end{align}
Write \(\hat J_t=(\hat J_{t,a})_{a=1}^K\) and
\(\hat G_t=(\hat G_{t,a})_{a=1}^K\) for the all-arm vectors.
Let \(s_{a,\res}^2:=s_a^2-\lambda_a^2\tilde s_a^2
=s_a^2(1-\varrho_a^2)\).
For fixed local parameters and a fixed allocation vector \(\pi_t\), as
\(n\to\infty\), the central limit theorem gives
\begin{align*}
\hat J_{t,a}&\xrightarrow{d}
\Normal\left(\theta_a^{(2)},\tilde s_a^2/b_t\right),
\qquad
\hat G_{t,a}\xrightarrow{d}
\Normal\left(\pi_{t,a}\theta_a^{(1)},
\pi_{t,a}s_{a,\res}^2/b_t\right).
\end{align*}
Under the stated moment conditions, these aggregates admit Gaussian
approximations without Gaussian individual outcomes, following the
diffusion-limit perspective of
\citet{kuang2021weak,fan2021diffusion,adusumilli2021risk,che2023adaptive}.

\paragraph{Why use the orthogonal decomposition?}
First, the normalized aggregates correspond to efficient generalized method
of moments (GMM) estimators.

\begin{proposition}[GMM form of the orthogonal aggregates]
\label{prop:gmm}
Fix an arm \(a\) and a batch \(t\) with fixed allocation probability
\(\pi_{t,a}>0\). Consider the GMM moments constructed from the labeled true
outcome, the labeled proxy outcome, and the all-unit proxy
outcome:
\begin{align*}
&\left(
\frac{\xi_{t,j,a}}{\pi_{t,a}}
\{\sqrt n\,Y_{t,j,a}-\theta_a^{(1)}-\lambda_a\theta_a^{(2)}\},\;
\frac{\xi_{t,j,a}}{\pi_{t,a}}
\{\sqrt n\,\tilde Y_{t,j,a}-\theta_a^{(2)}\},\;
\sqrt n\,\tilde Y_{t,j,a}-\theta_a^{(2)}
\right).
\end{align*}
Under the optimal weighting matrix, the GMM estimators satisfy
\(\hat\theta_{t,a}^{(2)}=\hat J_{t,a}+o_p(1)\) and
\(\hat\pi_{t,a}\hat\theta_{t,a}^{(1)}=\hat G_{t,a}\),
where \(\hat\pi_{t,a}:=(b_tn)^{-1}\sum_j\xi_{t,j,a}\) is the realized allocation proportion.
On the event \(\hat\pi_{t,a}>0\), whose probability tends to one, the
normalized orthogonal aggregates
\((\hat G_{t,a}/\hat\pi_{t,a},\hat J_{t,a})\) are asymptotically equivalent
to the efficient GMM estimators
\((\hat\theta_{t,a}^{(1)},\hat\theta_{t,a}^{(2)})\).
\end{proposition}

The proof is in Appendix~\ref{app:gmm-proof}. These efficient estimators
facilitate downstream allocation decisions.

Second, from a computational perspective, conditional on fixed local
parameters and for a fixed allocation probability \(\pi_{t,a}\), the two aggregate estimators
\((\hat G_{t,a},\hat J_{t,a})\) are uncorrelated and, by the joint central
limit theorem, asymptotically independent. This motivates the independent
priors on \(\theta_a^{(1)}\) and \(\theta_a^{(2)}\) adopted in
Section~\ref{subsec:bayes}. Together with the Gaussian likelihood, the
independent-prior assumption decouples the estimation of
\(\boldsymbol\theta^{(1)}\) and \(\boldsymbol\theta^{(2)}\), providing
substantial computational convenience for the Bayesian posterior updates
introduced in Section~\ref{subsec:bayes}. A similar factorization of the
posteriors for the prediction and correction components is used in Bayesian
PPI \citep{hofer2024bayesian}.

\subsection{Bayesian Posterior MDP}
\label{subsec:gaussian}
\label{subsec:bayes}

The arm-wise Gaussian approximations in Section~\ref{subsec:orthogonal}
motivate the following Gaussian sequential experiment and its posterior
updates.

\begin{definition}[Gaussian sequential experiment]
\label{def:gse}
A Gaussian sequential experiment, conditional on fixed local parameters, is
characterized by observations
\((J_0,G_0,\ldots,J_{T-1},G_{T-1})\) satisfying
\begin{align}
J_t&\sim
\Normal\left(
\boldsymbol\theta^{(2)},
\diag\left\{\tilde s_a^2/b_t\right\}_{a=1}^K
\right),
\end{align}
and
\begin{align}
G_t\mid J_0,G_0,\ldots,J_{t-1},G_{t-1},J_t
&\sim
\Normal\left(
\pi_t\odot\boldsymbol\theta^{(1)},
\diag\left\{\pi_{t,a}s_{a,\res}^2/b_t\right\}_{a=1}^K
\right).
\end{align}
The proxy signal \(J_t\) is independent of the preceding observations,
\(\pi_t\) is evaluated on the displayed history, and \(\odot\) denotes
elementwise multiplication.
\end{definition}

For the aggregate-based policies in Assumption~\ref{ass:gse}, the next
theorem establishes the sequential Gaussian limit.

\begin{theorem}[Gaussian sequential experiment limit]
\label{thm:gaussian-limit}
Under Assumption~\ref{ass:gse}, for any fixed \(T\) and any fixed local
parameter pair \((h,\tilde h)\),
\begin{align}
(\hat J_0,\hat G_0,\ldots,\hat J_{T-1},\hat G_{T-1})
&\xrightarrow{d}
(J_0,G_0,\ldots,J_{T-1},G_{T-1}),
\end{align}
where the limiting process is the Gaussian sequential experiment defined
in Definition~\ref{def:gse}.
\end{theorem}

Assumption~\ref{ass:gse} and the proof are given in
Appendices~\ref{app:gaussian-assumptions} and~\ref{app:gaussian-proof},
respectively. This limit motivates analyzing the Gaussian sequential
experiment as the asymptotic framework underlying
Algorithms~\ref{alg:rhope}--\ref{alg:rhope-delayed}; their finite-batch
posterior updates use realized allocation proportions.

For the Gaussian experiment, we specify the joint prior \(\nu\) through
independent conjugate priors
\begin{align*}
\theta_a^{(\ell)}
&\sim\Normal\left(\mu_{0,a}^{(\ell)},(\sigma_{0,a}^{(\ell)})^2\right),
\qquad a\in[K],\quad \ell\in\{1,2\}.
\end{align*}
The prior variances are finite and positive. This within- and across-arm
independence mirrors the orthogonal decomposition and yields two scalar
posterior updates per arm, simplifying computation and optimization.

Suppose that the posterior means and standard deviations upon completion of
batch \(t-1\) are
\((\boldsymbol\mu_t^{(1)},\boldsymbol\sigma_t^{(1)},
\boldsymbol\mu_t^{(2)},\boldsymbol\sigma_t^{(2)})\).
Given the Gaussian signals \(J_t\) followed by \(G_t\), the standard posterior
updates are, for each arm \(a\),
\begin{align}
&(\sigma_{t+1,a}^{(2)})^{-2}
=
(\sigma_{t,a}^{(2)})^{-2}
+\frac{b_t}{\tilde s_a^2},
\qquad
\mu_{t+1,a}^{(2)}
=
(\sigma_{t+1,a}^{(2)})^2
\left[
(\sigma_{t,a}^{(2)})^{-2}\mu_{t,a}^{(2)}
+\frac{b_t}{\tilde s_a^2}J_{t,a}
\right].
\label{eq:posterior-update-proxy}
\\
&(\sigma_{t+1,a}^{(1)})^{-2}
=
(\sigma_{t,a}^{(1)})^{-2}
+\frac{\pi_{t,a}b_t}{s_{a,\res}^2},
\qquad
\mu_{t+1,a}^{(1)}
=
(\sigma_{t+1,a}^{(1)})^2
\left[
(\sigma_{t,a}^{(1)})^{-2}\mu_{t,a}^{(1)}
+\frac{b_t}{s_{a,\res}^2}G_{t,a}
\right].
\label{eq:posterior-update-residual}
\end{align}

The posterior update induces the transition between the pre-LLM state
\(x_t^-\) and post-LLM state \(x_t^+\):
\begin{align*}
x_t^-
&=(\boldsymbol\mu_t^{(1)},\boldsymbol\sigma_t^{(1)},
\boldsymbol\mu_t^{(2)},\boldsymbol\sigma_t^{(2)})
\xrightarrow{\ J_t\ }
x_t^+=(\boldsymbol\mu_t^{(1)},\boldsymbol\sigma_t^{(1)},
\boldsymbol\mu_{t+1}^{(2)},\boldsymbol\sigma_{t+1}^{(2)})
\\
&\xrightarrow{\ G_t\ }
x_{t+1}^-=(\boldsymbol\mu_{t+1}^{(1)},\boldsymbol\sigma_{t+1}^{(1)},
\boldsymbol\mu_{t+1}^{(2)},\boldsymbol\sigma_{t+1}^{(2)}).
\end{align*}

Because \(h_a=\theta_a^{(1)}+\lambda_a\theta_a^{(2)}\) and the two posterior
components remain conditionally independent given the observations, after
batch \(t-1\) the local-scale mean \(h_a\) has
a normal posterior distribution with mean and variance
\begin{align*}
\mu_{t,a}
&:=
\mu_{t,a}^{(1)}+\lambda_a\mu_{t,a}^{(2)},
\qquad
\sigma_{t,a}^2
:=
(\sigma_{t,a}^{(1)})^2
+\lambda_a^2(\sigma_{t,a}^{(2)})^2.
\end{align*}

At the terminal stage, minimizing the Bayes simple regret in
\Eqref{eq:bayes-simple-regret} is equivalent to recommending the arm with the
highest posterior mean. Under this default recommendation rule, optimizing the
allocation policy is therefore equivalent to solving the following
problem, where \(V_t^{\pi,-}\) and \(V_t^{\pi,+}\) distinguish values before
and after observing \(J_t\), respectively.\footnote{Here
\(\E^\pi\) denotes expectation over the random variables
\(J_0,G_0,\ldots,J_{T-1},G_{T-1}\) under the prior predictive law.}
\begin{align}
\label{eq:bayes-posterior-objective}
&\max_{\pi_0,\ldots,\pi_{T-1}}
\left\{
V_0^{\pi,-}(
\boldsymbol\mu_0^{(1)},\boldsymbol\sigma_0^{(1)},
\boldsymbol\mu_0^{(2)},\boldsymbol\sigma_0^{(2)})
:=
\E^\pi\left[
\max_{a\in[K]}
\left\{\mu_{T,a}^{(1)}+\lambda_a\mu_{T,a}^{(2)}\right\}
\right]
\right\}.
\end{align}

To express \Eqref{eq:bayes-posterior-objective} in terms of each batch's
allocation policy, we reparameterize the Gaussian posterior updates,
connecting the current states \(x_t^-\) and \(x_t^+\) to the terminal
state \(x_T^-\).

\begin{lemma}[Posterior random-walk representation]
\label{lem:random-walk}
For every batch \(t\) and arm \(a\), the posterior means satisfy
\begin{align}
\label{eq:posterior-random-walk}
\mu_{t+1,a}^{(2)}
&=
\mu_{t,a}^{(2)}
+
\sqrt{
\frac{(\sigma_{t,a}^{(2)})^4 b_t}
{\tilde s_a^2+(\sigma_{t,a}^{(2)})^2 b_t}
}
Z_{t,a}^J,
\qquad
\mu_{t+1,a}^{(1)}
=
\mu_{t,a}^{(1)}
+
\sqrt{
\frac{(\sigma_{t,a}^{(1)})^4\pi_{t,a}b_t}
{s_{a,\res}^2+(\sigma_{t,a}^{(1)})^2\pi_{t,a}b_t}
}
Z_{t,a}^G,
\end{align}
where \(Z_{t,a}^J\) and \(Z_{t,a}^G\) are independent standard normal
innovations across arms, batches, and signal types.
\end{lemma}

From a post-LLM state, the value under policy \(\pi\) is
\begin{align}
\label{eq:value-expansion}
V_t^{\pi,+}(x_t^+)
&:=
\E^\pi\left[
\max_{a\in[K]}
\left\{\mu_{T,a}^{(1)}+\lambda_a\mu_{T,a}^{(2)}\right\}
\mid x_t^+
\right]
.
\end{align}
Lemma~\ref{lem:random-walk} expands this value in future Gaussian innovations;
the full expression is given in Appendix~\ref{app:random-walk-proof}.

We have now transformed the Bayes simple regret objective in
\Eqref{eq:bayes-simple-regret} into a dynamic-programming problem in
\Eqref{eq:value-expansion}. Exact
dynamic programming is nevertheless intractable because the state and action
spaces are continuous. RHO addresses this
challenge by reducing the remaining-horizon policy optimization to a tractable
single-simplex problem in each batch \citep{che2023adaptive}.

\subsection{Residual Horizon Optimization}
\label{subsec:rho}

In batch \(t\), RHO jointly optimizes \((\pi_t,\ldots,\pi_{T-1})\) using only
\(x_t^+\), implements the current allocation, and replans after the next
posterior update. The Gaussian posterior structure reduces this
remaining-horizon optimization to a single simplex.

\begin{lemma}[RHO reduction to one simplex]
\label{lem:rho}
Let \(B_t:=\sum_{v=t}^{T-1}b_v\) for \(t=0,\ldots,T\), with an empty sum
equal to zero. If the sequence of
future allocation policies is optimized using only the currently available
information \(x_t^+\), then the RHO optimization is equivalent to the
single-simplex maximization problem
\begin{align}
\label{eq:rho-single-simplex}
\rho_t
&\in
\argmax_{\bar\rho\in\DeltaK}
\E_{Z^G,Z^J}\Bigg[
\max_{a\in[K]}
\Bigg\{
\begin{aligned}
&
\mu_{t,a}^{(1)}
+\sqrt{
\frac{(\sigma_{t,a}^{(1)})^4\bar\rho_a B_t}
{s_{a,\res}^2+(\sigma_{t,a}^{(1)})^2\bar\rho_aB_t}
}Z_a^G
\\
&
+\lambda_a\mu_{t+1,a}^{(2)}
+\lambda_a
\sqrt{
\frac{(\sigma_{t+1,a}^{(2)})^4B_{t+1}}
{\tilde s_a^2+(\sigma_{t+1,a}^{(2)})^2B_{t+1}}
}Z_a^J
\end{aligned}
\Bigg\}
\Bigg].
\end{align}
Here \(Z_a^G\) and \(Z_a^J\) are independent standard normal random
variables.
\end{lemma}

We approximate the expectation in \Eqref{eq:rho-single-simplex} by Monte Carlo
and use gradient-based methods, such as SGD or Adam, to obtain a numerical
allocation policy \(\rho_t\).
The complete finite-batch procedure is given in
Algorithm~\ref{alg:rhope} (Appendix~\ref{app:rhope-algorithm}).
Relative to the RHO formulation of \citet{che2023adaptive}, proxy
outcomes provide advance information about the upcoming batch (hence the use
of \(\mu_{t+1,a}^{(2)}\) rather than \(\mu_{t,a}^{(2)}\)) and reduce
posterior uncertainty about arm means, supporting better allocation policies;
Section~\ref{subsec:llm-benefits}
provides a detailed theoretical analysis.

\section{Theoretical Analysis}
\label{sec:theory}

We provide a theoretical analysis of \method{}, examining its flat-prior connection to PPI++
(Section~\ref{subsec:ppi}) and the two benefits of LLM-based sample
augmentation: improved estimation precision and advance information for
upcoming-batch allocation (Section~\ref{subsec:llm-benefits}).

\subsection{Relationship with PPI++}
\label{subsec:ppi}

PPI++ combines labeled outcomes with predictions through a tuned correction
\citep{angelopoulos2023ppiplus}. We relate its mean-estimation formula to our
finite-batch working-posterior mean, accounting for pooled predictions and realized
sample counts.

\begin{theorem}[Flat-prior PPI++ representation]
\label{thm:ppi}
Consider the finite-batch Gaussian working-posterior updates in
Algorithms~\ref{alg:rhope} and~\ref{alg:rhope-delayed}, with fixed known
moments, \(\tilde s_a^2>0\), \(s_{a,\res}^2>0\), and zero initial
precisions. Fix \(t\in\{1,\ldots,T\}\) and an arm \(a\)
such that \(\sum_{v=0}^{t-1}\hat\pi_{v,a}b_v>0\).
Let \(\bar Y_{0:t-1,a}^{\mathrm{obs}}\) and
\(\bar{\tilde Y}_{0:t-1,a}^{\mathrm{obs}}\) be the true-outcome and proxy
sample means among units assigned to arm \(a\) in batches
\(0,\ldots,t-1\), normalized by their actual count, and let
\(\bar{\tilde Y}_{0:t-1,a}^{\mathrm{all}}\) be the corresponding all-unit
proxy mean. Then the working-posterior mean
\(\mu_{n,t,a}:=\mu_{n,t,a}^{(1)}+\lambda_a\mu_{n,t,a}^{(2)}\) for \(h_a\) satisfies
\begin{align}
\mu_{n,t,a}
&=
\sqrt n\left(
\bar Y_{0:t-1,a}^{\mathrm{obs}}
-
\lambda_a\bar{\tilde Y}_{0:t-1,a}^{\mathrm{obs}}
\right)
+
\lambda_a\sqrt n\,\bar{\tilde Y}_{0:t-1,a}^{\mathrm{all}}.
\end{align}
Up to the \(\sqrt n\) scaling, this is the pooled-sample PPI++ formula,
with the tuning-coefficient conversion given in the proof.
\end{theorem}

Theorem~\ref{thm:ppi} establishes an exact algebraic identity, without
transferring i.i.d.-sampling guarantees to adaptive data. With a nonflat
prior, the orthogonal aggregates are combined with prior information through
Gaussian posterior updates. Whereas conventional PPI centers on a single
outcome, our BAI setting has one potential outcome per arm, so the connection
holds arm by arm. Under the theorem's conditions, proxy outcomes
reduce working-posterior variance for the same realized allocation, as shown
in the proof. This connection extends PPI++ from statistical
inference to adaptive experimental design.

\subsection{Benefits of LLM-Generated Proxy Outcomes}
\label{subsec:llm-benefits}

Within the Gaussian sequential experiment, we compare three methods to
isolate the estimation and advance-information benefits of proxy outcomes.

\paragraph{RHO.}
RHO \citep{che2023adaptive} uses only true outcomes, without proxy outcomes
or their contextual information.

\paragraph{\delayedmethod{}.}
\delayedmethod{} adds proxy outcomes and their contextual information to RHO
but uses the current batch's proxies only after allocation. This
delayed-information ablation isolates the value of observing proxies before
allocation.
The complete procedure is given in Algorithm~\ref{alg:rhope-delayed}.

\paragraph{\method{}.}
\method{} is our proposed method. In addition to using proxy outcomes
and their contextual information, it uses the current batch's proxy
outcomes when optimizing the current allocation policy.
The complete procedure is given in Algorithm~\ref{alg:rhope}.

Under the RHO approximation, denote the fixed-allocation expected
returns of RHO, \delayedmethod{}, and \method{} by
\(\widetilde V_t^{\mathrm{RHO}}(\bar\rho,x_t^-)\),
\(\widetilde V_t^{\mathdelayedmethod}(\bar\rho,x_t^-)\), and
\(\widetilde V_t^{\mathmethod}(\bar\rho,x_t^+)\), respectively.
Here \(\bar\rho\in\DeltaK\) is fixed, and \(x_t^-\) and \(x_t^+\) are the pre-LLM and post-LLM states.

We compare returns at a common pre-LLM state, initializing RHO with the same
marginal posterior for \(h\) as the two proxy-based methods.

\begin{theorem}[RHOPE return comparison]
\label{thm:rho-return-comparison}
For any fixed pre-LLM state \(x_t^-\),
\begin{align}
\max_{\bar\rho\in\DeltaK}
\widetilde V_t^{\mathrm{RHO}}(\bar\rho,x_t^-)
&\le
\max_{\bar\rho\in\DeltaK}
\widetilde V_t^{\mathdelayedmethod}(\bar\rho,x_t^-)
\le
\E_{J_t}
\left[
\max_{\bar\rho\in\DeltaK}
\widetilde V_t^{\mathmethod}(\bar\rho,x_t^+)
\mid x_t^-
\right].
\end{align}
\end{theorem}

Theorem~\ref{thm:rho-return-comparison}
(proved in Appendix~\ref{app:proof-rho-return-comparison}) identifies two
benefits of sample augmentation, which supplies proxy information for every
arm, including those receiving no units. First, combining proxies with true
outcomes improves estimation precision by reducing posterior uncertainty
about arm means. Second, observing current-batch proxies before allocation
provides advance information for better allocation decisions.

Building on the ordering in Theorem~\ref{thm:rho-return-comparison}, we now
quantify the two gaps. Proposition~\ref{prop:estimation-gap} turns the first
inequality into a comparison of the Bayes simple regret guarantees of RHO and
\delayedmethod{}, and Proposition~\ref{prop:advance-gap} bounds the gap
between \delayedmethod{} and \method{}, the second inequality, directly.

For the estimation benefit we consider exchangeable arms: the prior
parameters \((\mu_{0,a}^{(\ell)},\sigma_{0,a}^{(\ell)})\) and the moments
\((s_a,\tilde s_a,\varrho_a)\) do not depend on \(a\), so we drop the arm
index on \(s,\tilde s,\varrho,\lambda,s_{\res}\), and the \(h_a\) are
i.i.d.\ a priori with variance
\(\tau^2:=(\sigma_0^{(1)})^2+\lambda^2(\sigma_0^{(2)})^2\). Let
\(c_K:=\E[\max_{a\le K}Z_a]\) for i.i.d.\ standard normal
\(Z_1,\ldots,Z_K\), let
\(\mathcal R_K(\sigma^2;\tau):=c_K\bigl(\tau-\sqrt{\tau^2-\sigma^2}\bigr)\)
for \(0\le\sigma^2\le\tau^2\), and let
\begin{align*}
\sigma_{T,\mathrm{PE}}^2
:=\Bigl[(\sigma_0^{(1)})^{-2}+\tfrac{1}{Ks_{\res}^2}\Bigr]^{-1}
+\lambda^2\Bigl[(\sigma_0^{(2)})^{-2}+\tfrac{1}{\tilde s^2}\Bigr]^{-1},
\qquad
\sigma_{T,0}^2:=\Bigl[\tau^{-2}+\tfrac{1}{Ks^2}\Bigr]^{-1}
\end{align*}
be the terminal posterior variances of \(h_a\) under uniform allocation
with and without proxies, respectively.

\begin{proposition}[Estimation benefit as a regret guarantee]
\label{prop:estimation-gap}
Assume exchangeable arms. For \textup{\delayedmethod{}} and RHO,
\begin{align*}
\mathfrak R_T(\mathdelayedmethod,\nu)
\le\mathcal R_K(\sigma_{T,\mathrm{PE}}^2;\tau),
\qquad
\mathfrak R_T(\RHO,\nu)\le\mathcal R_K(\sigma_{T,0}^2;\tau).
\end{align*}
Moreover, \(\sigma_{T,\mathrm{PE}}^2\le\sigma_{T,0}^2\), and under a flat
prior the ratio of the two bounds converges to the ratio of the limiting
terminal variances,
\begin{align*}
\frac{\mathcal R_K(\sigma_{T,\mathrm{PE}}^2;\tau)}
{\mathcal R_K(\sigma_{T,0}^2;\tau)}
\;\to\;
\frac{Ks_{\res}^2+\lambda^2\tilde s^2}{Ks^2}
\;=\;
1-\varrho^2\Bigl(1-\frac1K\Bigr).
\end{align*}
\end{proposition}

The proof is in Appendix~\ref{app:proof-estimation-gap}. The gap between
the two guarantees is thus controlled by the ratio of the terminal
posterior variances: under a flat prior, proxies reduce the regret
guarantee by the fraction \(\varrho^2(1-1/K)\). This reduction comes from
the improvement of the terminal posterior variance from \(\sigma_{T,0}^2\)
to \(\sigma_{T,\mathrm{PE}}^2\) once proxy information is added, and it
makes precise the first benefit identified after
Theorem~\ref{thm:rho-return-comparison}: combining proxies with true
outcomes reduces posterior uncertainty about the arm means.

\begin{proposition}[Advance-information benefit]
\label{prop:advance-gap}
For any pre-LLM state \(x_t^-\) with \(t\ge1\), the gap in the second
inequality of Theorem~\ref{thm:rho-return-comparison},
\begin{align*}
\Delta_t^{\mathrm{adv}}(x_t^-)
:=
\E_{J_t}\Bigl[\max_{\bar\rho\in\DeltaK}
\widetilde V_t^{\mathmethod}(\bar\rho,x_t^+)\mid x_t^-\Bigr]
-\max_{\bar\rho\in\DeltaK}
\widetilde V_t^{\mathdelayedmethod}(\bar\rho,x_t^-),
\end{align*}
satisfies, under a flat prior,
\begin{align*}
0\le\Delta_t^{\mathrm{adv}}(x_t^-)
\le
2\sqrt{2\log(2K)}\,
\max_{a\in[K]}|\lambda_a|(\sigma_{t,a}^{(2)})^2
\sqrt{\frac{b_t}{\tilde s_a^2+(\sigma_{t,a}^{(2)})^2b_t}}
\;\asymp\;
\frac{\sqrt{T\log K}}{t},
\end{align*}
where \(\asymp\) holds for equal batches up to a constant that does not
depend on \(t\), \(T\), or \(K\).
\end{proposition}

The proof is in Appendix~\ref{app:proof-advance-gap}. Under a flat prior,
the advance-information gap at batch \(t\) is therefore of order
\(\sqrt{T\log K}/t\) and decays with \(t\), whereas the estimation channel
removes the fixed fraction \(\varrho^2(1-1/K)\) of the posterior variance in
every batch (Proposition~\ref{prop:estimation-gap}). Estimation precision is
therefore the dominant channel, consistent with the experiments in
Section~\ref{sec:experiments}.

\begin{table}[t]
\centering
\caption{Final relative simple regret (\%) for (a) Bernoulli and (b) Gumbel
with \(\sigma^2=1\); uniform is 100 in every column, and lower is better.
Entries are mean \(\pm\) standard error of the mean over 5,000
replications, both as percentages of the mean simple regret of uniform.
Boldface marks column minima; dashes denote no-proxy results unchanged
across \(\varrho\).}
\label{tab:synthetic}
\small\setlength{\tabcolsep}{3pt}
\begin{minipage}{\textwidth}
\centering
\textbf{(a) Bernoulli setting}\par
\begin{tabular}{lrrrrrr}
\toprule
& \multicolumn{3}{c}{\(b_tn=100\)} & \multicolumn{3}{c}{\(b_tn=1000\)} \\
\cmidrule(lr){2-4}\cmidrule(lr){5-7}
Policy / \(\varrho\) & \(0.2\) & \(0.4\) & \(0.6\) & \(0.2\) & \(0.4\) & \(0.6\) \\
\midrule
\multicolumn{7}{l}{\underline{\text{Without proxy}}} \\
uniform & 100.00\sem{1.60} & -- & -- & 100.00\sem{1.55} & -- & -- \\
TS & 79.78\sem{1.44} & -- & -- & 77.74\sem{1.37} & -- & -- \\
TTTS & 78.64\sem{1.41} & -- & -- & 76.89\sem{1.37} & -- & -- \\
Sequential Halving & 77.17\sem{1.37} & -- & -- & 75.00\sem{1.32} & -- & -- \\
TS-SPAS & 108.89\sem{1.66} & -- & -- & 107.31\sem{1.63} & -- & -- \\
myopic & 76.87\sem{1.39} & -- & -- & 74.11\sem{1.33} & -- & -- \\
RHO & 74.48\sem{1.34} & -- & -- & 71.37\sem{1.29} & -- & -- \\
\addlinespace[0pt]
\multicolumn{7}{l}{\underline{\text{With proxy}}} \\
uniform-LLM & 100.45\sem{1.59} & 95.48\sem{1.55} & 83.46\sem{1.43} & 94.18\sem{1.51} & 87.73\sem{1.45} & 79.15\sem{1.37} \\
TS-LLM & 77.51\sem{1.39} & 73.78\sem{1.35} & 63.58\sem{1.24} & 73.91\sem{1.32} & 69.16\sem{1.28} & 59.96\sem{1.18} \\
TTTS-LLM & 77.33\sem{1.39} & 70.36\sem{1.32} & 62.85\sem{1.23} & 75.93\sem{1.37} & 70.28\sem{1.31} & 61.37\sem{1.21} \\
MLA-UCB & 97.09\sem{1.58} & 92.35\sem{1.54} & 75.62\sem{1.40} & 92.45\sem{1.51} & 86.39\sem{1.45} & 77.24\sem{1.35} \\
myopic-LLM & 74.93\sem{1.35} & 70.43\sem{1.29} & 61.64\sem{1.19} & 74.39\sem{1.32} & 68.79\sem{1.26} & 59.23\sem{1.16} \\
\delayedmethod{} & 71.68\sem{1.31} & \textbf{66.43}\sem{1.27} & 59.37\sem{1.17} & 69.95\sem{1.27} & 65.86\sem{1.24} & 57.89\sem{1.15} \\
\method{} & \textbf{71.43}\sem{1.31} & 67.05\sem{1.27} & \textbf{57.88}\sem{1.14} & \textbf{69.82}\sem{1.30} & \textbf{65.47}\sem{1.25} & \textbf{56.66}\sem{1.13} \\
\bottomrule
\end{tabular}
\end{minipage}\par\medskip
\begin{minipage}{\textwidth}
\centering
\textbf{(b) Gumbel setting (\(\sigma^2=1\))}\par
\begin{tabular}{lrrrrrr}
\toprule
& \multicolumn{3}{c}{\(b_tn=100\)} & \multicolumn{3}{c}{\(b_tn=1000\)} \\
\cmidrule(lr){2-4}\cmidrule(lr){5-7}
Policy / \(\varrho\) & \(0.2\) & \(0.4\) & \(0.6\) & \(0.2\) & \(0.4\) & \(0.6\) \\
\midrule
\multicolumn{7}{l}{\underline{\text{Without proxy}}} \\
uniform & 100.00\sem{1.85} & -- & -- & 100.00\sem{1.87} & -- & -- \\
TS & 63.76\sem{1.42} & -- & -- & 70.18\sem{1.53} & -- & -- \\
TTTS & 66.48\sem{1.46} & -- & -- & 71.85\sem{1.57} & -- & -- \\
Sequential Halving & 64.90\sem{1.43} & -- & -- & 69.35\sem{1.50} & -- & -- \\
TS-SPAS & 125.62\sem{2.18} & -- & -- & 125.24\sem{2.14} & -- & -- \\
myopic & 63.82\sem{1.41} & -- & -- & 68.75\sem{1.46} & -- & -- \\
RHO & 61.63\sem{1.41} & -- & -- & 65.06\sem{1.43} & -- & -- \\
\addlinespace[0pt]
\multicolumn{7}{l}{\underline{\text{With proxy}}} \\
uniform-LLM & 97.04\sem{1.81} & 89.79\sem{1.74} & 76.82\sem{1.57} & 96.23\sem{1.83} & 88.72\sem{1.72} & 76.41\sem{1.58} \\
TS-LLM & 62.59\sem{1.39} & 56.32\sem{1.32} & 49.40\sem{1.22} & 66.91\sem{1.48} & 63.62\sem{1.38} & 51.97\sem{1.22} \\
TTTS-LLM & 65.57\sem{1.45} & 59.76\sem{1.35} & 46.89\sem{1.16} & 68.17\sem{1.49} & 62.11\sem{1.38} & 52.56\sem{1.25} \\
MLA-UCB & 91.15\sem{1.76} & 89.08\sem{1.75} & 73.42\sem{1.56} & 98.73\sem{1.82} & 89.78\sem{1.72} & 74.20\sem{1.52} \\
myopic-LLM & 61.97\sem{1.39} & 55.60\sem{1.32} & 47.95\sem{1.19} & 67.72\sem{1.45} & 61.42\sem{1.34} & 50.87\sem{1.22} \\
\delayedmethod{} & 60.02\sem{1.36} & \textbf{51.36}\sem{1.22} & 44.94\sem{1.12} & \textbf{62.83}\sem{1.40} & 58.43\sem{1.33} & 49.24\sem{1.19} \\
\method{} & \textbf{58.34}\sem{1.34} & 52.91\sem{1.26} & \textbf{44.45}\sem{1.12} & 63.11\sem{1.42} & \textbf{56.23}\sem{1.29} & \textbf{48.45}\sem{1.18} \\
\bottomrule
\end{tabular}
\end{minipage}
\end{table}

\begin{table}[t]
\centering
\caption{GDSC relative simple regret (\%) with known moments. Entries are
mean \(\pm\) standard error of the mean over 10,000 replications, both as
percentages of the mean simple regret of uniform; lower is better. Boldface
marks column minima among the displayed values.}
\label{tab:semi}
\fontsize{8}{9.6}\selectfont\setlength{\tabcolsep}{3pt}
\begin{tabular}{@{}lrrrrrr@{}}
\toprule
Policy / proxy & Claude Opus 5.5 & GPT-6.1 Sol & DeepSeek V4 Pro & DeepSeek V4 Flash & GBT & Ridge \\
\midrule
\multicolumn{7}{l}{\underline{\text{Without proxy}}} \\
uniform & 100.00\sem{1.87} & -- & -- & -- & -- & -- \\
TS & 51.92\sem{1.27} & -- & -- & -- & -- & -- \\
TTTS & 45.84\sem{1.22} & -- & -- & -- & -- & -- \\
Sequential Halving & 60.79\sem{1.38} & -- & -- & -- & -- & -- \\
TS-SPAS & 215.91\sem{3.80} & -- & -- & -- & -- & -- \\
myopic & 45.34\sem{1.20} & -- & -- & -- & -- & -- \\
RHO & 42.55\sem{1.17} & -- & -- & -- & -- & -- \\
\addlinespace[1pt]
\multicolumn{7}{l}{\underline{\text{With proxy}}} \\
uniform-LLM & 87.89\sem{1.68} & 88.41\sem{1.68} & 89.80\sem{1.72} & 91.65\sem{1.73} & 96.53\sem{1.82} & 95.86\sem{1.78} \\
TS-LLM & 44.67\sem{1.18} & 45.47\sem{1.21} & 47.35\sem{1.21} & 48.28\sem{1.22} & 49.90\sem{1.25} & 55.09\sem{1.33} \\
TTTS-LLM & 43.18\sem{1.14} & 41.18\sem{1.11} & 44.19\sem{1.16} & 44.22\sem{1.18} & 45.71\sem{1.20} & 51.49\sem{1.26} \\
MLA-UCB & 50.13\sem{1.23} & 49.38\sem{1.23} & 51.23\sem{1.25} & 51.34\sem{1.25} & 48.04\sem{1.22} & \textbf{49.02}\sem{1.22} \\
myopic-LLM & 44.23\sem{1.17} & 43.93\sem{1.17} & 44.51\sem{1.18} & 43.79\sem{1.16} & 47.42\sem{1.22} & 55.43\sem{2.16} \\
\delayedmethod{} & \textbf{38.54}\sem{1.10} & 39.53\sem{1.12} & 39.39\sem{1.11} & \textbf{39.55}\sem{1.12} & 42.68\sem{1.15} & 54.12\sem{2.23} \\
\method{} & 38.70\sem{1.11} & \textbf{38.91}\sem{1.11} & \textbf{39.26}\sem{1.11} & 40.69\sem{1.14} & \textbf{41.82}\sem{1.14} & 54.39\sem{2.22} \\
\bottomrule
\end{tabular}
\end{table}

\section{Experiments}
\label{sec:experiments}

Synthetic and GDSC-based semi-synthetic experiments evaluate finite-sample
decision gains and their dependence on proxy informativeness.

\subsection{Synthetic Data Experiments}
\label{subsec:synthetic}

The main synthetic experiments pair true and proxy outcomes with Bernoulli
or Gumbel marginals, using equal-sized batches of \(b_tn\in\{100,1000\}\)
and Pearson correlation \(\varrho\in\{0.2,0.4,0.6\}\). The main Gumbel
variance is \(\sigma^2=1\); policies use known variances and covariances.
Appendix~\ref{app:experiments} specifies the distributions and parameter
priors and reports the more complex Bernoulli--Gumbel setting and Gumbel
variance ablations with \(\sigma^2\in\{1/3,3\}\).

We compare \method{} with uniform allocation, Thompson sampling (TS),
top-two Thompson sampling (TTTS) \citep{russo2016simple}, MLA-UCB
\citep{ji2025surrogate}, sequential halving
\citep{karnin2013almost,zhao2023revisiting}, TS-SPAS
\citep{kato2025minimax}, myopic one-step optimization, RHO
\citep{che2023adaptive}, and LLM-assisted counterparts where available.
Uniform allocation provides the 100\% simple-regret reference in each setting.
Table~\ref{tab:synthetic} presents the main results;
Appendix~\ref{app:synthetic-results} provides experimental details and
Appendix~\ref{app:unknown-moments} presents ablation experiments with
estimated moments.

Across Tables~\ref{tab:synthetic} and~\ref{tab:synthetic-bernoulli-gumbel},
either \method{} or \delayedmethod{} attains the lowest relative simple
regret in every setting, showing that directly optimizing Bayes simple regret
with informative proxies can outperform heuristic algorithms. Their similar
performance suggests that estimation precision drives most gains, with a
modest additional benefit from advance information. LLM-assisted methods
generally improve on their non-LLM counterparts, increasingly so at larger
\(\varrho\). These findings are consistent with Section~\ref{sec:theory}.

\subsection{Semi-Synthetic Data Experiment: GDSC Dataset}
\label{subsec:semi}

The GDSC-based benchmark uses seven cancer drugs and 627 cell lines (cell
lines with imputed entries in the response matrix are excluded, and 22 of
the remaining ones form a pool from which 16 in-context examples are drawn
at random) \citep{yang2013gdsc}. Proxy
outcomes are generated from each cell line's features by four LLMs, Claude
Opus 5.5, GPT-6.1 Sol, DeepSeek V4 Pro, and DeepSeek V4 Flash, through
16-shot prompts, and by two traditional machine-learning predictors,
gradient-boosted trees (GBT) and ridge regression, trained on the same 16
in-context cell lines. The proxy--outcome correlations, averaged over arms,
decrease across these proxies: 0.31 for Claude Opus 5.5, 0.29 for GPT-6.1
Sol, 0.26 for DeepSeek V4 Pro, 0.20 for DeepSeek V4 Flash, 0.19 for GBT,
and 0.16 for ridge, so proxy informativeness decreases in the same order.
Table~\ref{tab:semi} presents the main results;
Appendix~\ref{app:semi-synthetic-details} provides experimental details and
Appendix~\ref{app:unknown-moments} presents ablation experiments with
estimated moments.

According to Table~\ref{tab:semi}, \method{} or \delayedmethod{} attains
the lowest relative simple regret with every proxy except ridge, the
weakest predictor. With the three stronger LLMs, Claude Opus 5.5, GPT-6.1
Sol, and DeepSeek V4 Pro, both are significantly better than RHO, which uses
no proxies, showing that proxy information generated by a strong LLM leads to better adaptive
experimental designs. Comparing across proxies, the LLM-generated proxies
have higher proxy--outcome correlations than those of the traditional
machine-learning predictors, and the performance of \method{} and
\delayedmethod{} deteriorates as this correlation decreases. LLMs are
therefore better suited than traditional techniques to this setting, which
requires generated data of high quality.

\section{Conclusion}
\label{sec:conclusion}

We introduced \method, a framework incorporating contextual LLM-generated
proxy outcomes into posterior updating and allocation for finite-batch BAI.
Our analysis distinguishes estimation and advance-information benefits, while
experiments suggest estimation precision drives most gains.
Both synthetic and GDSC-based semi-synthetic experiments highlight the
importance of proxy informativeness and moment-estimation accuracy.

\subsection*{AI use statement}

Claude Opus 5.5, GPT-6.1 Sol, DeepSeek V4 Pro, and DeepSeek V4 Flash were used
as part of the experimental procedure to generate the proxy outcomes evaluated
in the semi-synthetic experiment.
Generative AI tools were also used to assist with writing proofs, writing code,
generating figures, reviewing methodological choices, interpreting results,
translating and polishing text, and organizing references. The
authors manually reviewed and verified all citations, proofs, code, data
processing, and reported results. The authors take full responsibility for the
final content of this work, including all text, claims, code, results, figures,
and other artifacts produced with the assistance of generative AI.

\subsection*{Reproducibility statement}

The assumptions and complete proofs for the theoretical results are provided in
Appendix~\ref{app:proofs}. Appendix~\ref{app:experiments} documents the
experimental settings, data construction, prompting procedure, benchmark
policies, and additional results. The complete finite-batch implementations of
\method{} and \delayedmethod{} are specified in Appendix~\ref{app:rhope-algorithm}.
Appendix~\ref{app:unknown-moments} reports ablations with estimated moments for both
synthetic and semi-synthetic experiments, estimating the required outcome
moments from the first batch.

\bibliography{references}
\bibliographystyle{iclr2027_conference}

\appendix
\raggedbottom

\section{Extended Related Work}
\label{app:extended-related-work}

\paragraph{Adaptive experimental design and bandits.}
Adaptive experimental design is often modeled as a multi-armed bandit problem
because both involve sequential sampling under partial feedback. However, the
appropriate design and algorithm can differ substantially across optimization
objectives. Variance-minimizing designs allocate samples to reduce uncertainty
in target estimators \citep{neyman1934optimal,zhao2024adaptive,kato2020efficient}.
Cumulative-regret bandits trade off exploration and exploitation during the
sampling process \citep{lai1985asymptotically,auer2002finite,russo2018tutorial,lattimore2020bandit}.
Pure-exploration objectives, including BAI, evaluate the final recommendation, not
the cumulative outcomes collected during sampling \citep{audibert2010best,bubeck2011pure,
russo2016simple,kato2021contextual}. Most algorithms in these areas are driven
by confidence bounds, posterior sampling, or elimination heuristics.

A recent line of work instead uses diffusion limits to directly optimize
experiment-level risk. \citet{kuang2021weak} study weak-signal asymptotics for
sequentially randomized experiments, and \citet{fan2021diffusion} derive
diffusion approximations for Thompson sampling. \citet{adusumilli2021risk}
formulates bandit experiments as Bayes and minimax risk problems in a diffusion
regime, defines asymptotic Bayes risk as expected cumulative regret, and
characterizes the minimal Bayes risk through the associated partial differential
equation, the continuous-time limit of the dynamic program. \citet{che2023adaptive}
approximate batched adaptive experiments by Gaussian sequential experiments and
use residual horizon optimization (RHO) to optimize the allocation policy. Our
algorithm follows this line: rather than using a classical bandit heuristic, it
formulates a Gaussian posterior MDP and optimizes the Bayes simple regret
objective.

\paragraph{Best-arm identification.}
BAI is a pure-exploration problem that asks researchers to identify the arm
with the largest mean. It is closely related to ranking-and-selection problems in
simulation optimization \citep{swisher2003survey}. BAI is broadly divided into
fixed-confidence and fixed-budget settings. In the fixed-confidence setting,
the goal is to stop as quickly as possible while controlling the probability
of error \citep{garivier2016optimal,kaufmann2016complexity,kato2021contextual}.
In the fixed-budget setting, common objectives are to minimize either the
probability of error \citep{audibert2010best,karnin2013almost,
carpentier2016tight,kato2022smallgap} or simple regret
\citep{audibert2010best,bubeck2011pure,russo2016simple,zhao2023revisiting,
kato2025minimax}. The probability-of-error criterion considers only whether the selected arm is
optimal, whereas simple regret also accounts for the expected-outcome gap between the
selected and optimal arms. This paper studies fixed-budget simple-regret BAI.

\paragraph{Prediction-powered inference.}
Prediction-powered inference (PPI) uses predictions from machine learning
systems to improve statistical inference when labels are expensive and
unlabeled covariates are abundant \citep{angelopoulos2023ppi}. PPI++ improves
efficiency by tuning the coefficient of a prediction-based correction
\citep{angelopoulos2023ppiplus}. Recent work studies additional variants and
multi-predictor extensions \citep{ji2025surrogates,miao2025postprediction,
shan2025sada,carlson2025unifying,xu2025semisupervised,kluger2025imputed,
vanderlaan2026calibeating}. Bayesian and Bayes-assisted frequentist
formulations have also been developed
\citep{hofer2024bayesian,cortinovis2025fabppi}. Recent studies extend PPI ideas
to causal inference \citep{debartolomeis2025efficient,shankar2025augmenting}.
Adaptive experimentation is naturally a missing-data
problem: each sampled unit reveals only the outcome under the assigned
arm. Thus, although our objective is decision quality rather than efficient
inference alone, PPI ideas can also be used in adaptive
experimentation \citep{ji2025surrogate,ma2026proxy,ao2026textual}. Among
these, \citet{ma2026proxy} also study best-arm identification with
generative proxies, but in the fixed-confidence setting: an elimination
procedure with a \(\delta\)-PAC guarantee minimizes the number of true
samples needed to certify the best arm, and it has no budget parameter.
Our setting is fixed-budget BAI with Bayes simple regret as the
objective, so the two methods address different problems and are not
directly comparable. We use
prediction-powered aggregates in posterior MDP state transitions and, on this
basis, construct an optimization framework for adaptive experimentation with
Bayes simple regret as its objective.

\section{Proofs}
\label{app:proofs}

\subsection{Proof of Proposition~\ref{prop:gmm}}
\label{app:gmm-proof}

Fix the local parameters, one arm, and one batch, and suppress the arm and
batch indices. Let the batch size be
\(M=b_tn\), let \(S_i=\ind\{A_i=a\}\), let
\(\pi=\Pr(S_i=1)\), and write
\[
R_i=\sqrt n\,Y_i,\qquad \tilde R_i=\sqrt n\,\tilde Y_i.
\]
Then \(\E[R_i]=h\), \(\E[\tilde R_i]=\tilde h\),
\(\Var(R_i)=ns^2\), \(\Var(\tilde R_i)=n\tilde s^2\), and
\(\Cov(R_i,\tilde R_i)=nc\), where \(c:=\varrho s\tilde s\). The factor
\(n\) appears because the GMM moments use the local-scale variables
\(\sqrt n\,Y_i\) and \(\sqrt n\,\tilde Y_i\). This is a fixed-\(\pi\), one-batch
GMM calculation; under randomized assignment, \(S_i\) is independent of
\((R_i,\tilde R_i)\). The raw GMM construction uses the three moments
\[
\begin{pmatrix}
\frac{S_i}{\pi}(R_i-h)\\
\frac{S_i}{\pi}(\tilde R_i-\tilde h)\\
\tilde R_i-\tilde h
\end{pmatrix}.
\]
They correspond respectively to the labeled true outcome, the labeled
proxy outcome, and the all-unit proxy outcome. Their
covariance matrix is
\[
F=
n\begin{pmatrix}
\pi^{-1}s^2 & \pi^{-1}c & c\\
\pi^{-1}c & \pi^{-1}\tilde s^2 & \tilde s^2\\
c & \tilde s^2 & \tilde s^2
\end{pmatrix}.
\]
For \(0<\pi<1\), the nondegeneracy conditions \(s^2>0\),
\(\tilde s^2>0\), and \(|\varrho|<1\) make \(F\) positive definite.
We first consider this case. The optimally weighted GMM criterion uses
weighting matrix \(F^{-1}\):
\[
Q(h,\tilde h)=
\bar m(h,\tilde h)^\top F^{-1}\bar m(h,\tilde h),
\]
where \(\bar m(h,\tilde h)\) is the sample average of the displayed moment
vector. The common factor \(n\) in \(F\) only rescales the optimal GMM
criterion and does not change its minimizer.

Set \(\lambda=c/\tilde s^2=\varrho s/\tilde s\), and define the transformed
parameters
\[
\theta^{(1)}=h-\lambda\tilde h,\qquad \theta^{(2)}=\tilde h.
\]
Because GMM with the optimal weighting matrix is invariant to nonsingular
linear transformations of the moment vector, we apply
\[
L^{-1}
=
\begin{pmatrix}
1&-\lambda&0\\
0&1&-1\\
0&0&1
\end{pmatrix}.
\]
Then
\[
L^{-1}
\begin{pmatrix}
\frac{S_i}{\pi}(R_i-h)\\
\frac{S_i}{\pi}(\tilde R_i-\tilde h)\\
\tilde R_i-\tilde h
\end{pmatrix}
=
\begin{pmatrix}
\frac{S_i}{\pi}
\{R_i-\lambda\tilde R_i-\theta^{(1)}\}\\
\left(\frac{S_i}{\pi}-1\right)(\tilde R_i-\theta^{(2)})\\
\tilde R_i-\theta^{(2)}
\end{pmatrix}.
\]
The first transformed moment is the residual score. It is orthogonal to the
proxy score because
\[
\Cov(R_i-\lambda\tilde R_i,\tilde R_i)
=
n\left(c-\lambda\tilde s^2\right)=0.
\]
Since \(S_i\) is independent of the outcomes, the transformed
covariance matrix is diagonal:
\[
L^{-1}F(L^{-1})^\top
=
n\begin{pmatrix}
\pi^{-1}s_{\res}^2&0&0\\
0&(\pi^{-1}-1)\tilde s^2&0\\
0&0&\tilde s^2
\end{pmatrix},
\qquad
s_{\res}^2=s^2-\lambda^2\tilde s^2.
\]
Thus the optimal weighting matrix in the transformed coordinates simply
weights three mutually orthogonal components. The second transformed moment,
\((S_i/\pi-1)(\tilde R_i-\theta^{(2)})\), has expectation zero for every
value of \((\theta^{(1)},\theta^{(2)})\). Its population derivative with
respect to \(\theta^{(2)}\) is also zero because
\[
-\E\left[\frac{S_i}{\pi}-1\right]=0.
\]
It therefore contributes no first-order information about the target means.
Write \(N=\sum_{i=1}^M S_i\), \(\hat\pi=N/M\), and
\(\hat J=M^{-1}\sum_{i=1}^M\tilde R_i\). Because \(\pi>0\),
\(\Pr(N=0)=(1-\pi)^M\to0\); the following estimator calculations are on
the event \(N>0\).
More explicitly, if \(\bar D_M\) denotes the sample Jacobian of the
transformed sample moments with respect to
\((\theta^{(1)},\theta^{(2)})\), then
\[
\bar D_M
=
\begin{pmatrix}
-\hat\pi/\pi&0\\
0&-(\hat\pi/\pi-1)\\
0&-1
\end{pmatrix}
\xrightarrow{p}
D
:=
\begin{pmatrix}
-1&0\\
0&0\\
0&-1
\end{pmatrix}.
\]
The optimally weighted GMM first-order condition is
\[
\bar D_M^\top
\{L^{-1}F(L^{-1})^\top\}^{-1}
\bar m_L(\theta^{(1)},\theta^{(2)})=0,
\]
where \(\bar m_L\) is the sample average of the transformed moment vector.
The first component of this condition gives the exact identity
\[
\hat\theta^{(1)}
=\frac1N\sum_{i:S_i=1}(R_i-\lambda\tilde R_i).
\]
The second component gives
\[
\hat\theta^{(2)}-\hat J
=\frac{\hat\pi/\pi-1}{\pi^{-1}-1}
\frac1M\sum_{i=1}^M
\left(\frac{S_i}{\pi}-1\right)
(\tilde R_i-\hat\theta^{(2)}).
\]
Subtracting and adding \(\hat J\) inside the final factor and solving for
\(\hat\theta^{(2)}-\hat J\) yields
\[
\hat\theta^{(2)}-\hat J
=
\frac{\hat\pi/\pi-1}
{(\pi^{-1}-1)+(\hat\pi/\pi-1)^2}
\frac1M\sum_{i=1}^M
\left(\frac{S_i}{\pi}-1\right)(\tilde R_i-\hat J).
\]
Thus the finite-sample GMM estimator need not equal \(\hat J\).
To bound their difference, note that
\(\hat\pi/\pi-1=O_p(M^{-1/2})\) and
\[
\begin{aligned}
&\frac1M\sum_{i=1}^M
\left(\frac{S_i}{\pi}-1\right)(\tilde R_i-\hat J)
\\
&\quad=
\frac1M\sum_{i=1}^M
\left(\frac{S_i}{\pi}-1\right)(\tilde R_i-\theta^{(2)})
-\left(\frac{\hat\pi}{\pi}-1\right)(\hat J-\theta^{(2)})
=O_p(1).
\end{aligned}
\]
Indeed, independence of assignment and outcomes implies that the first term
on the right has mean zero and variance
\(n(\pi^{-1}-1)\tilde s^2/M\), while
\(\Var(\hat J)=n\tilde s^2/M\). Both variances are bounded because
\(M=b_tn\). The denominator in the exact difference formula converges in
probability to \(\pi^{-1}-1>0\), so
\[
\hat\theta^{(2)}-\hat J=O_p(M^{-1/2})=o_p(1).
\]
Returning to the notation of batch \(t\), these relations imply
\[
\hat\theta_{t,a}^{(2)}
=
\frac1{b_t\sqrt n}\sum_{j=1}^{b_tn}\tilde Y_{t,j,a}
+o_p(1)
=\hat J_{t,a}+o_p(1),
\]
and
\[
\hat\pi_{t,a}\hat\theta_{t,a}^{(1)}
=
\frac1{b_tn}\sum_{j=1}^{b_tn}
\xi_{t,j,a}\sqrt n
\left(Y_{t,j,a}-\lambda_a\tilde Y_{t,j,a}\right)
=
\frac1{b_t\sqrt n}\sum_{j=1}^{b_tn}
\xi_{t,j,a}
\left(Y_{t,j,a}-\lambda_a\tilde Y_{t,j,a}\right)
=\hat G_{t,a}.
\]
Consequently, on \(\hat\pi_{t,a}>0\),
\[
\left(\frac{\hat G_{t,a}}{\hat\pi_{t,a}},\hat J_{t,a}\right)
-\left(\hat\theta_{t,a}^{(1)},\hat\theta_{t,a}^{(2)}\right)
\xrightarrow{p}(0,0).
\]
Thus the normalized orthogonal aggregates are asymptotically equivalent to
the optimally weighted GMM estimators and inherit their asymptotic efficiency.
If \(\pi=1\), the labeled and all-unit proxy moments are
identical. Removing the duplicate moment leaves two exactly identified
equations, which give \(\hat\theta^{(2)}=\hat J\) and the same residual-mean
estimator of \(\theta^{(1)}\), with \(\hat\pi=1\). The same conclusion
therefore holds in this case, with exact equality.
This proves the proposition.

\subsection{Regularity Conditions for Theorem~\ref{thm:gaussian-limit}}
\label{app:gaussian-assumptions}

Theorem~\ref{thm:gaussian-limit} concerns the aggregate-based policy class
specified below. It establishes the Gaussian limit used as the asymptotic
framework for Algorithms~\ref{alg:rhope}--\ref{alg:rhope-delayed}.
The finite-batch implementations, whose posterior updates also use realized
allocation proportions, are specified separately in
Appendix~\ref{app:rhope-algorithm}.

\begin{assumption}[Regularity for the finite-batch experiment]
\label{ass:gse}
Conditional on fixed local parameters, for every batch \(t\) and unit \(j\), let
\(\xi_{t,j}=(\xi_{t,j,1},\ldots,\xi_{t,j,K})\),
\(\epsilon_{t,j}=(\epsilon_{t,j,1},\ldots,\epsilon_{t,j,K})\), and
\(\tilde\epsilon_{t,j}
=(\tilde\epsilon_{t,j,1},\ldots,\tilde\epsilon_{t,j,K})\). The noise-vector
pairs \((\epsilon_{t,j},\tilde\epsilon_{t,j})\) are identically distributed
and independent across units and batches. The following conditions hold.

1. \emph{(Ignorability)} Given the aggregate history available before physical
assignment, the assignment vectors are conditionally independent categorical
draws with probability vector \(\pi_t\), and the collection of assignment
vectors is conditionally independent of all true and proxy outcomes in batch \(t\):
\[
\{\xi_{t,j}:1\le j\le b_tn\}\perp
\{Y_{t,j,a},\tilde Y_{t,j,a}:1\le j\le b_tn,\ a\in[K]\}
\mid
\hat J_0,\hat G_0,\ldots,\hat J_{t-1},\hat G_{t-1},\hat J_t.
\]

2. \emph{(Moment restriction)} There exists a constant
\(C>0\) such that \(\E\|\epsilon_{t,j}\|_2^4\le C\) and
\(\E\|\tilde\epsilon_{t,j}\|_2^4\le C\). With
\(\Lambda=\diag(\lambda_1,\ldots,\lambda_K)\), the joint covariance structure
satisfies \(\Cov(\tilde\epsilon_{t,j})
=\diag\{\tilde s_a^2\}_{a=1}^K\) and
\(\Cov(\tilde\epsilon_{t,j},
\epsilon_{t,j}-\Lambda\tilde\epsilon_{t,j})=0\).

3. \emph{(Sufficiency)} The policy depends on the experimental data only through
the aggregate estimators:
\[
\pi_t
=
\pi_t(\hat J_0,\hat G_0,\ldots,\hat J_{t-1},\hat G_{t-1},\hat J_t).
\]
4. \emph{(Continuity)} The set of discontinuity points of \(\pi_t(\cdot)\)
has measure zero under the distribution of
\((J_0,G_0,\ldots,J_{t-1},G_{t-1},J_t)\).
\end{assumption}

Ignorability specifies the randomization mechanism: given
the aggregate history used by the policy, physical arm assignment follows the
chosen allocation probabilities and is independent of the realized potential
outcomes. The joint orthogonality condition extends the arm-wise decomposition
in Section~\ref{subsec:orthogonal} to the full vectors: it rules out residual
cross-covariance between the all-arm proxy signal and the all-arm residual
signal. Appendix~\ref{app:covariance-robustness} motivates this condition
and the diagonal proxy covariance, and examines the posterior when they
fail. Sufficiency states that the adaptive allocation rule uses the aggregate
estimators rather than the unit-level outcomes. The fourth-moment bound and the continuity
condition are standard regularity conditions needed to pass from batch-level
central limit approximations to a sequential weak-convergence result.

\subsection{Proof of Theorem~\ref{thm:gaussian-limit}}
\label{app:gaussian-proof}

For probability laws \(\mu\) and \(\nu\) on \(\mathbb R^m\), define the
bounded-Lipschitz distance by
\[
d_{\mathrm{BL}}(\mu,\nu)
:=
\sup_{\varphi\in\mathcal F_{\mathrm{BL}}}
\left|
\int \varphi\,d\mu-\int \varphi\,d\nu
\right|,
\qquad
\mathcal F_{\mathrm{BL}}
:=
\left\{
\varphi:\mathbb R^m\to\mathbb R:\|\varphi\|_\infty\le1,\ 
\mathrm{Lip}(\varphi)\le1
\right\},
\]
where
\[
\mathrm{Lip}(\varphi)
:=
\sup_{x\ne y}
\frac{|\varphi(x)-\varphi(y)|}{\|x-y\|_2}.
\]
For random vectors \(U\) and \(V\), we write
\(d_{\mathrm{BL}}(U,V)\) as shorthand for
\(d_{\mathrm{BL}}(\mathcal L(U),\mathcal L(V))\).
This distance metrizes weak convergence on every finite-dimensional Euclidean
space.

The proof follows the induction structure of
\citet[Appendix B.1]{che2023adaptive}, but the one-batch argument must be
modified. In our experiment, the current allocation is selected after
observing the current all-unit proxy aggregate estimator \(\hat J_t\). Consequently, conditioning
on \(\hat J_t\) need not leave the current residuals independent, even though
the residuals and proxy outcomes are orthogonal. We therefore prove a
joint functional central limit theorem for processes representing \(\hat J_t\)
and \(\hat G_t\), indexed by all possible allocation vectors.

Let \(\Lambda=\diag(\lambda_1,\ldots,\lambda_K)\), and, within a fixed batch,
write
\[
V_j:=\tilde\epsilon_{t,j},
\qquad
R_j:=\epsilon_{t,j}-\Lambda\tilde\epsilon_{t,j}.
\]
Assumption~\ref{ass:gse} and the definition of \(\lambda_a\) imply
\[
\Cov(V_j)=\diag\{\tilde s_a^2\}_{a=1}^K,
\qquad
\Cov(V_j,R_j)=0,
\qquad
\Var(R_{j,a})=s_{a,\res}^2.
\]

\paragraph{One-batch joint limit.}
Fix batch \(t\), and consider deterministic pre-batch histories
\(r_{0:t-1}^{(n)}\to r_{0:t-1}\); for \(t=0\), the history is absent. We
show that the finite-batch transition from \(r_{0:t-1}^{(n)}\) converges to
the Gaussian sequential-experiment transition from \(r_{0:t-1}\), provided that
\(\pi_t\) is continuous at \((r_{0:t-1},J_t)\) with probability one.

Set \(m=b_tn\). By the conditional categorical randomization in
Assumption~\ref{ass:gse}, the assignments can be represented using i.i.d.\
\(U_j\sim\operatorname{Unif}(0,1)\), independent of the current-batch noise
vectors. For \(p\in\DeltaK\), define
\[
c_0(p):=0,
\qquad
c_a(p):=\sum_{\ell=1}^a p_\ell,
\qquad
\xi_{j,a}(p):=\ind\{c_{a-1}(p)<U_j\le c_a(p)\}.
\]
Also define
\[
\mathbb V_m
:=
\frac1{\sqrt m}\sum_{j=1}^m V_j,
\qquad
\mathbb W_m(q)
:=
\frac1{\sqrt m}\sum_{j=1}^m R_j\ind\{U_j\le q\},
\quad 0\le q\le1,
\]
and \(F_m(q):=m^{-1}\sum_{j=1}^m\ind\{U_j\le q\}\).

To justify the required joint functional limit, consider the class consisting
of the coordinate functions \((v,r,u)\mapsto v_a\) and the threshold-indexed
functions
\((v,r,u)\mapsto r_a\ind\{u\le q\}\), where
\(a\in[K]\) and \(q\in[0,1]\). This is a VC-type class with an
\(L^2\) envelope under Assumption~\ref{ass:gse}; hence it is Donsker
\citep[Sections 2.5--2.6]{vanderVaart1996weak}. Its empirical-process limit,
together with the uniform Glivenko--Cantelli theorem for \(F_m\), gives
\[
\bigl(\mathbb V_m,\mathbb W_m(\cdot),F_m(\cdot)\bigr)
\xrightarrow{d}
\bigl(\mathbb V,\mathbb W(\cdot),q\mapsto q\bigr)
\]
in
\(\mathbb R^K\times\ell^\infty([0,1])^K\times\ell^\infty([0,1])\).
The limiting variables satisfy
\[
\mathbb V
\sim
\Normal\left(0,\diag\{\tilde s_a^2\}_{a=1}^K\right),
\]
and \(\mathbb W\) is a continuous mean-zero Gaussian process with
\[
\Cov\!\left(\mathbb W(q),\mathbb W(q')\right)
=(q\wedge q')\Cov(R_j),
\qquad
\Cov\!\left(\mathbb V,\mathbb W(q)\right)
=q\Cov(V_j,R_j)=0.
\]
Because the joint limit is Gaussian, \(\mathbb V\) and the entire process
\(\mathbb W\) are independent. The Glivenko--Cantelli part of the same
display also gives
\[
\sup_{0\le q\le1}|F_m(q)-q|\xrightarrow{p}0.
\]

For the actual allocation
\[
p_m
:=
\pi_t(r_{0:t-1}^{(n)},\hat J_t),
\]
where the pre-batch-history argument is omitted when \(t=0\), the definitions
of the aggregate estimators give the exact representations
\[
\hat J_t
=
\boldsymbol\theta^{(2)}+\frac1{\sqrt{b_t}}\mathbb V_m
\]
and
\[
\hat G_{t,a}
=
\frac{N_{t,a}}m\theta_a^{(1)}
+
\frac1{\sqrt{b_t}}
\left\{
\mathbb W_{m,a}(c_a(p_m))
-\mathbb W_{m,a}(c_{a-1}(p_m))
\right\},
\]
where
\[
\frac{N_{t,a}}m
=
F_m(c_a(p_m))-F_m(c_{a-1}(p_m)).
\]
Uniformly over \(p_m\in\DeltaK\),
\[
\left|\frac{N_{t,a}}m-p_{m,a}\right|
\le
2\sup_{0\le q\le1}|F_m(q)-q|
\xrightarrow{p}0.
\]

It follows from the functional central limit theorem, the assumed continuity
of \(\pi_t\), and the extended continuous mapping theorem that
\[
\left(\hat J_t,p_m,\mathbb W_m(\cdot)\right)
\xrightarrow{d}
\left(J_t,\pi_t(r_{0:t-1},J_t),\mathbb W(\cdot)\right),
\]
where
\[
J_t
=
\boldsymbol\theta^{(2)}+\frac1{\sqrt{b_t}}\mathbb V.
\]
Because \(\mathbb W\) has continuous sample paths, evaluation at the
allocation-dependent interval endpoints is continuous. Hence
\[
(\hat J_t,\hat G_t)
\xrightarrow{d}
(J_t,G_t)
\]
under the corresponding deterministic histories, where, writing
\(p=\pi_t(r_{0:t-1},J_t)\),
\[
G_{t,a}
=
p_a\theta_a^{(1)}
+
\frac1{\sqrt{b_t}}
\left\{
\mathbb W_a(c_a(p))-\mathbb W_a(c_{a-1}(p))
\right\}.
\]

It remains to verify that this is exactly the transition in
Definition~\ref{def:gse}. Conditional on \(J_t\), the vector \(p\) is fixed
and \(\mathbb W\) remains independent of \(J_t\). For any \(a,b\in[K]\),
the covariance of the two Gaussian increments in the preceding display is
\[
\begin{aligned}
&\Cov\Big(
\mathbb W_a(c_a(p))-\mathbb W_a(c_{a-1}(p)),
\mathbb W_b(c_b(p))-\mathbb W_b(c_{b-1}(p))
\Big)
\\
&\qquad=
\Cov(R_{j,a},R_{j,b})
\left|
(c_{a-1}(p),c_a(p)]
\cap
(c_{b-1}(p),c_b(p)]
\right|
\\
&\qquad=
\ind\{a=b\}\,p_a s_{a,\res}^2.
\end{aligned}
\]
Therefore
\[
G_t\mid r_{0:t-1},J_t
\sim
\Normal\left(
p\odot\boldsymbol\theta^{(1)},
\diag\left\{\frac{p_as_{a,\res}^2}{b_t}\right\}_{a=1}^K
\right).
\]
This calculation also covers \(p_a=0\): the corresponding interval is empty
and the Gaussian increment is identically zero. This proves the one-batch joint
limit without conditioning the residual sample on the realized value of
\(\hat J_t\).

\paragraph{Induction over batches.}
For every fixed \(t=0,\ldots,T-1\), write
\[
H_t^{(n)}
:=
(\hat J_0,\hat G_0,\ldots,\hat J_t,\hat G_t),
\qquad
H_t
:=
(J_0,G_0,\ldots,J_t,G_t).
\]
We prove \(H_t^{(n)}\xrightarrow{d}H_t\) by induction. For \(t=0\), the
one-batch joint limit applies with no previous history.
Assumption~\ref{ass:gse} ensures that \(\pi_0\) is continuous at \(J_0\)
with probability one, so the base case follows.

Suppose that \(H_t^{(n)}\xrightarrow{d}H_t\). To prove convergence through
batch \(t+1\), fix a bounded Lipschitz function \(f\) satisfying
\[
\sup_{x\ne y}
\frac{|f(x)-f(y)|}{\|x-y\|_2}
\le1,
\qquad
\sup_{x,y}|f(x)-f(y)|\le1.
\]
It is enough to consider this normalization because multiplying \(f\) by a
positive constant and subtracting a constant do not affect whether the
difference of expectations converges to zero.

For a deterministic history \(r\in\mathbb R^{2K(t+1)}\), let
\(P_{n,t+1}(r,\cdot)\) denote the finite-batch adaptive-experiment transition law of
\((\hat J_{t+1},\hat G_{t+1})\) when the previous history supplied to the
policy is fixed at \(r\), and let \(P_{t+1}(r,\cdot)\) denote the corresponding
Gaussian transition law. Define
\[
g_n(r)
:=
\int f(r,z)\,P_{n,t+1}(r,dz),
\qquad
g(r)
:=
\int f(r,z)\,P_{t+1}(r,dz).
\]
This transition-kernel definition avoids conditioning on a probability-zero
event of the form \(H_t^{(n)}=r\).

Let \(D_{t+1}\) be the set of discontinuity points of \(\pi_{t+1}\), and let
\(Q_{t+1}^J\) be the Gaussian law of \(J_{t+1}\). Define
\[
\mathcal C_t
:=
\left\{
r:\
Q_{t+1}^J\!\left(
\{j:(r,j)\in D_{t+1}\}
\right)=0
\right\}.
\]
Assumption~\ref{ass:gse} and Fubini's theorem imply
\[
\Pr(H_t\in\mathcal C_t)=1.
\]
The one-batch joint limit established above gives the following
\emph{continuous convergence} of the transition kernels: for every
deterministic sequence \(r^{(n)}\to r\) with \(r\in\mathcal C_t\),
\[
P_{n,t+1}(r^{(n)},\cdot)
\Rightarrow
P_{t+1}(r,\cdot).
\]
Indeed, this is precisely the preceding one-batch argument with
\(r_{0:t}^{(n)}=r^{(n)}\) and \(r_{0:t}=r\). Moreover,
\[
\begin{aligned}
|g_n(r^{(n)})-g(r)|
&\le
\int
|f(r^{(n)},z)-f(r,z)|\,P_{n,t+1}(r^{(n)},dz)
\\
&\quad+
\left|
\int f(r,z)\,P_{n,t+1}(r^{(n)},dz)
-
\int f(r,z)\,P_{t+1}(r,dz)
\right|.
\end{aligned}
\]
The first term is at most \(\|r^{(n)}-r\|_2\) by the Lipschitz property of
\(f\), and the second converges to zero because \(z\mapsto f(r,z)\) is bounded
and continuous and the transition laws converge weakly. Hence
\begin{equation}
\label{eq:continuous-transition}
r^{(n)}\to r\in\mathcal C_t
\quad\Longrightarrow\quad
g_n(r^{(n)})\to g(r).
\end{equation}

Appendix B.1 of \citet{che2023adaptive} obtains the induction step by proving
uniform convergence of its one-step conditional expectations. Here,
the implication in \eqref{eq:continuous-transition}, which follows from the joint
functional limit, is sufficient. Because
\(\mathbb R^{2K(t+1)}\) is a Polish space, the Skorohod representation theorem
applied to the induction hypothesis guarantees the existence of random vectors
\(H_t^{(n),*}\) and \(H_t^*\) on a common probability space such that
\[
H_t^{(n),*}\stackrel{d}=H_t^{(n)},
\qquad
H_t^*\stackrel{d}=H_t,
\qquad
H_t^{(n),*}\longrightarrow H_t^*
\quad\text{almost surely}.
\]
The equalities in distribution are essential: the coupled variables need not
be the original histories, but every expectation below is unchanged by
replacing the original histories with these copies.

Let \(\Omega_0\) be the event on which
\(H_t^{(n),*}\to H_t^*\) and \(H_t^*\in\mathcal C_t\). The preceding displays
give \(\Pr(\Omega_0)=1\). For each \(\omega\in\Omega_0\), apply
\eqref{eq:continuous-transition} to the deterministic sequence
\(r^{(n)}=H_t^{(n),*}(\omega)\) and limit
\(r=H_t^*(\omega)\). This establishes that
\[
g_n(H_t^{(n),*})
\longrightarrow
g(H_t^*)
\quad\text{almost surely}.
\]
Because \(f\) is bounded, \(g_n\) and \(g\) are bounded uniformly in \(n\).
Dominated convergence and the equalities in distribution therefore give
\[
\E[g_n(H_t^{(n)})]
=
\E[g_n(H_t^{(n),*})]
\longrightarrow
\E[g(H_t^*)]
=
\E[g(H_t)].
\]
By the tower property, the first and last expectations are respectively
\[
\E[f(H_t^{(n)},\hat J_{t+1},\hat G_{t+1})]
\quad\text{and}\quad
\E[f(H_t,J_{t+1},G_{t+1})].
\]
Thus the expectations converge for every bounded Lipschitz \(f\), proving
\(H_{t+1}^{(n)}\xrightarrow{d}H_{t+1}\). This completes the induction step.
Since \(T\) is fixed, induction gives
\[
(\hat J_0,\hat G_0,\ldots,\hat J_{T-1},\hat G_{T-1})
\xrightarrow{d}
(J_0,G_0,\ldots,J_{T-1},G_{T-1}),
\]
which is the claimed Gaussian sequential experiment limit.

\subsection{Proof of Lemma~\ref{lem:random-walk}}
\label{app:random-walk-proof}

Consider a scalar Gaussian prior and likelihood:
\[
\theta\sim\Normal(\mu,\sigma^2),\qquad
Y\mid\theta\sim\Normal(\theta,v).
\]
The posterior precision and mean are
\[
\sigma_+^{-2}=\sigma^{-2}+v^{-1},\qquad
\mu_+=\sigma_+^2(\sigma^{-2}\mu+v^{-1}Y).
\]
Under the prior predictive distribution,
\[
Y=\theta+\eta,\qquad \theta=\mu+\sigma Z_1,\qquad
\eta=\sqrt v Z_2,
\]
with independent standard normals \(Z_1,Z_2\). Hence
\[
\mu_+-\mu
=
\sigma_+^2v^{-1}(Y-\mu)
=
\sigma_+^2v^{-1}(\sigma Z_1+\sqrt v Z_2).
\]
Its mean is zero and its variance is
\[
(\sigma_+^2v^{-1})^2(\sigma^2+v)
=
\left(\frac{\sigma^2}{\sigma^2+v}\right)^2(\sigma^2+v)
=
\frac{\sigma^4}{\sigma^2+v}.
\]
Therefore
\[
\mu_+=\mu+\sqrt{\frac{\sigma^4}{\sigma^2+v}}\,Z
\]
for a standard normal \(Z\). Substituting \(v=\tilde s_a^2/b_t\) for the proxy
signal gives the displayed recursion from \(\mu_{t,a}^{(2)}\) to
\(\mu_{t+1,a}^{(2)}\). Let \(\mathcal F_t^-\) be the sigma-field generated
by all observations before \(J_t\), and let
\(\mathcal F_t^+:=\sigma(\mathcal F_t^-,J_t)\); the allocation \(\pi_t\) is
\(\mathcal F_t^+\)-measurable. For the residual component, when
\(\pi_{t,a}>0\), the conditional likelihood
\(G_{t,a}\mid\theta_a^{(1)},\mathcal F_t^+\sim
\Normal(\pi_{t,a}\theta_a^{(1)},\pi_{t,a}s_{a,\res}^2/b_t)\) is equivalent to
\(G_{t,a}/\pi_{t,a}\mid\theta_a^{(1)},\mathcal F_t^+\sim
\Normal(\theta_a^{(1)},s_{a,\res}^2/(\pi_{t,a}b_t))\). Substituting this
variance gives the second formula.
When \(\pi_{t,a}=0\), \(G_{t,a}=0\) almost surely and its likelihood does
not depend on \(\theta_a^{(1)}\). The posterior mean remains
\(\mu_{t,a}^{(1)}\), and the random-walk coefficient is zero.
Thus the second formula also holds at the boundary
\(\pi_{t,a}=0\).

It remains to justify the joint independence claimed in the lemma when the
allocation is adaptive. Conditional on \(\mathcal F_t^-\), the vector
of standardized proxy posterior-mean increments is multivariate standard normal
with identity covariance, and its conditional law does not depend on
\(\mathcal F_t^-\). It is therefore independent of \(\mathcal F_t^-\), with
independent coordinates across arms. Likewise, conditional on
\(\mathcal F_t^+\), the standardized residual posterior-mean increments on
coordinates with \(\pi_{t,a}>0\) are independent standard normals. Complete
this vector on coordinates with \(\pi_{t,a}=0\) using independent standard
normals; their coefficients are zero, so they do not change the posterior-mean
process. The resulting \(K\)-dimensional vector has conditional law
\(\Normal(0,I_K)\), which does not depend on \(\mathcal F_t^+\).
It is therefore independent of the observed history, including the current
proxy innovations. Applying this argument successively over
\(t=0,\ldots,T-1\) proves mutual independence across arms, batches, and
signal types.

Iterating the recursions in Lemma~\ref{lem:random-walk} gives the full
expansion of \Eqref{eq:value-expansion}:
\begin{align*}
V_t^{\pi,+}(x_t^+)
&=\E^\pi\Bigg[
\max_{a\in[K]}\Bigg\{
\begin{aligned}
&\mu_{t,a}^{(1)}
+\sum_{v=t}^{T-1}
\sqrt{\frac{(\sigma_{v,a}^{(1)})^4\pi_{v,a}b_v}
{s_{a,\res}^2+(\sigma_{v,a}^{(1)})^2\pi_{v,a}b_v}}Z_{v,a}^G
\\
&+\lambda_a\mu_{t+1,a}^{(2)}
+\lambda_a\sum_{v=t+1}^{T-1}
\sqrt{\frac{(\sigma_{v,a}^{(2)})^4b_v}
{\tilde s_a^2+(\sigma_{v,a}^{(2)})^2b_v}}Z_{v,a}^J
\end{aligned}
\Bigg\}\,\Bigg|\,x_t^+\Bigg].
\end{align*}

\subsection{Proof of Lemma~\ref{lem:rho}}
\label{app:rho-proof}

Fix \(t\in\{0,\ldots,T-1\}\) and a post-LLM state \(x_t^+\). Under RHO, the future allocation sequence
\((\pi_t,\ldots,\pi_{T-1})\) is selected using only \(x_t^+\). Hence,
conditional on \(x_t^+\), it is a deterministic open-loop sequence. For each
arm \(a\), define its budget-weighted average allocation by
\[
\bar\rho_a
=
\frac1{B_t}\sum_{v=t}^{T-1}b_v\pi_{v,a}.
\]
Because every \(\pi_v\) belongs to \(\DeltaK\), \(\bar\rho_a\ge0\) and
\[
\sum_{a=1}^K\bar\rho_a
=
\frac1{B_t}\sum_{v=t}^{T-1}b_v\sum_{a=1}^K\pi_{v,a}
=
1.
\]
Thus \(\bar\rho\in\DeltaK\). For arm \(a\), the residual posterior
precision after the full future allocation sequence is
\[
(\sigma_{T,a}^{(1)})^{-2}
=
(\sigma_{t,a}^{(1)})^{-2}
+
\sum_{v=t}^{T-1}\frac{b_v\pi_{v,a}}{s_{a,\res}^2}.
\]
By the definition of \(\bar\rho\), this is
\[
(\sigma_{T,a}^{(1)})^{-2}
=
(\sigma_{t,a}^{(1)})^{-2}
+
\frac{B_t\bar\rho_a}{s_{a,\res}^2}.
\]

By Lemma~\ref{lem:random-walk}, the ex-ante variance of the terminal residual
posterior mean equals the current residual posterior variance minus the
terminal residual posterior variance:
\[
(\sigma_{t,a}^{(1)})^2
-
\left[
(\sigma_{t,a}^{(1)})^{-2}
+
\frac{B_t\bar\rho_a}{s_{a,\res}^2}
\right]^{-1}
=
\frac{(\sigma_{t,a}^{(1)})^4\bar\rho_a B_t}
{s_{a,\res}^2+(\sigma_{t,a}^{(1)})^2\bar\rho_aB_t}.
\]
Therefore, conditional on \(x_t^+\), the terminal residual posterior mean has
the representation
\[
\mu_{T,a}^{(1)}
=
\mu_{t,a}^{(1)}
+
\sqrt{
\frac{(\sigma_{t,a}^{(1)})^4\bar\rho_a B_t}
{s_{a,\res}^2+(\sigma_{t,a}^{(1)})^2\bar\rho_aB_t}
}
Z_a^G.
\]

The current proxy signal \(J_t\) has already been absorbed into \(x_t^+\), so
only batches \(t+1,\ldots,T-1\) contribute future proxy uncertainty. These
future proxy signals do not depend on the allocation sequence. The terminal
proxy posterior precision is therefore
\[
(\sigma_{T,a}^{(2)})^{-2}
=
(\sigma_{t+1,a}^{(2)})^{-2}
+
\frac{B_{t+1}}{\tilde s_a^2},
\]
and the same normal posterior calculation gives
\[
\mu_{T,a}^{(2)}
=
\mu_{t+1,a}^{(2)}
+
\sqrt{
\frac{(\sigma_{t+1,a}^{(2)})^4B_{t+1}}
{\tilde s_a^2+(\sigma_{t+1,a}^{(2)})^2B_{t+1}}
}
Z_a^J.
\]
The random variables \(Z_a^G\) and \(Z_a^J\) are mutually independent standard
normals across arms and signal types by the diagonal Gaussian transition
covariances and the independence of the residual and proxy
posterior components. Combining the two displays gives the terminal posterior mean of
the target arm value,
\[
\begin{aligned}
\mu_{T,a}^{(1)}+\lambda_a\mu_{T,a}^{(2)}
&=
\mu_{t,a}^{(1)}+\lambda_a\mu_{t+1,a}^{(2)}
\\
&\quad+
\sqrt{
\frac{(\sigma_{t,a}^{(1)})^4\bar\rho_a B_t}
{s_{a,\res}^2+(\sigma_{t,a}^{(1)})^2\bar\rho_aB_t}
}
Z_a^G
\\
&\quad+
\lambda_a
\sqrt{
\frac{(\sigma_{t+1,a}^{(2)})^4B_{t+1}}
{\tilde s_a^2+(\sigma_{t+1,a}^{(2)})^2B_{t+1}}
}
Z_a^J.
\end{aligned}
\]
Thus every open-loop sequence selected from \(x_t^+\) induces a terminal
predictive distribution that depends on that sequence only through
\(\bar\rho\). Conversely, for every \(\bar\rho\in\DeltaK\), the constant
sequence \(\pi_v=\bar\rho\) for \(v=t,\ldots,T-1\) realizes the same
budget-weighted average allocation. Hence the set of terminal predictive distributions
attainable by jointly optimizing the \(T-t\) future allocation policies is
exactly the set indexed by \(\bar\rho\in\DeltaK\). Taking the expectation of
the terminal posterior maximum over this set yields the single-simplex
optimization problem displayed in Lemma~\ref{lem:rho}.

\subsection{Proof of Theorem~\ref{thm:ppi}}

We use the finite-batch updates in Algorithms~\ref{alg:rhope}
and~\ref{alg:rhope-delayed}. Their ordering differs, but both incorporate
\(\hat J_v\) and \(\hat G_v\) by the end of batch \(v\).
Throughout this proof, the moments and \(\lambda_a\) are fixed and known.
By the definition of the realized allocation proportion,
\begin{align*}
n\sum_{v=0}^{t-1}\hat\pi_{v,a}b_v
&=\sum_{v=0}^{t-1}\sum_{j=1}^{b_vn}\xi_{v,j,a}.
\end{align*}
Thus the three means in the theorem are
\begin{align*}
\bar Y_{0:t-1,a}^{\mathrm{obs}}
&:=
\frac{
\sum_{v=0}^{t-1}\sum_{j=1}^{b_vn}\xi_{v,j,a}Y_{v,j,a}
}{
n\sum_{v=0}^{t-1}\hat\pi_{v,a}b_v
},
\\
\bar{\tilde Y}_{0:t-1,a}^{\mathrm{obs}}
&:=
\frac{
\sum_{v=0}^{t-1}\sum_{j=1}^{b_vn}\xi_{v,j,a}\tilde Y_{v,j,a}
}{
n\sum_{v=0}^{t-1}\hat\pi_{v,a}b_v
},
\\
\bar{\tilde Y}_{0:t-1,a}^{\mathrm{all}}
&:=
\frac{
\sum_{v=0}^{t-1}\sum_{j=1}^{b_vn}\tilde Y_{v,j,a}
}{
n\sum_{v=0}^{t-1}b_v
}.
\end{align*}
The condition in the theorem requires at least one observed outcome for arm
\(a\), ensuring that the residual working posterior is proper.

\paragraph{Accumulating the finite-batch updates.}
A flat prior has zero initial precision and zero initial
precision-weighted mean. Summing the residual precision and
precision-weighted-mean increments in the algorithms gives
\begin{align*}
(\sigma_{n,t,a}^{(1)})^{-2}
&=\frac{\sum_{v=0}^{t-1}b_v\hat\pi_{v,a}}{s_{a,\res}^2},
\\
(\sigma_{n,t,a}^{(1)})^{-2}\mu_{n,t,a}^{(1)}
&=\frac{\sum_{v=0}^{t-1}b_v\hat G_{v,a}}{s_{a,\res}^2}.
\end{align*}
Consequently, the definition of \(\hat G_{v,a}\) yields
\begin{align*}
\mu_{n,t,a}^{(1)}
&=
\frac{\sum_{v=0}^{t-1}b_v\hat G_{v,a}}
{\sum_{v=0}^{t-1}\hat\pi_{v,a}b_v}
\\
&=
\frac{
\sum_{v=0}^{t-1}\sum_{j=1}^{b_vn}
\xi_{v,j,a}(Y_{v,j,a}-\lambda_a\tilde Y_{v,j,a})
}{
\sqrt n\sum_{v=0}^{t-1}\hat\pi_{v,a}b_v
}
\\
&=
\sqrt n\left(
\bar Y_{0:t-1,a}^{\mathrm{obs}}
-\lambda_a\bar{\tilde Y}_{0:t-1,a}^{\mathrm{obs}}
\right).
\end{align*}
Similarly, summing the proxy update increments gives
\begin{align*}
(\sigma_{n,t,a}^{(2)})^{-2}
&=\frac{\sum_{v=0}^{t-1}b_v}{\tilde s_a^2},
\\
(\sigma_{n,t,a}^{(2)})^{-2}\mu_{n,t,a}^{(2)}
&=\frac{\sum_{v=0}^{t-1}b_v\hat J_{v,a}}{\tilde s_a^2}.
\end{align*}
Using the definition of \(\hat J_{v,a}\), we obtain
\begin{align*}
\mu_{n,t,a}^{(2)}
&=
\frac{\sum_{v=0}^{t-1}b_v\hat J_{v,a}}
{\sum_{v=0}^{t-1}b_v}
=
\sqrt n\,\bar{\tilde Y}_{0:t-1,a}^{\mathrm{all}}.
\end{align*}
Recombining the two posterior means gives
\begin{align*}
\mu_{n,t,a}
&=
\mu_{n,t,a}^{(1)}+\lambda_a\mu_{n,t,a}^{(2)}
=
\sqrt n\left(
\bar Y_{0:t-1,a}^{\mathrm{obs}}
-\lambda_a\bar{\tilde Y}_{0:t-1,a}^{\mathrm{obs}}
\right)
+\lambda_a\sqrt n\,\bar{\tilde Y}_{0:t-1,a}^{\mathrm{all}}.
\end{align*}
This is an exact identity for the finite-batch working-posterior updates,
using realized allocation proportions throughout; no replacement of
\(\hat\pi_{v,a}\) by \(\pi_{v,a}\) is made.

\paragraph{Relation to the usual PPI++ formula.}
For this comparison, write
\(m:=n\sum_{v=0}^{t-1}\hat\pi_{v,a}b_v>0\) and
\(u:=n\sum_{v=0}^{t-1}(1-\hat\pi_{v,a})b_v\) for the numbers of units
with and without an observed true outcome for arm \(a\), respectively.
If \(u>0\), define the unlabeled-only proxy mean by
\begin{align*}
\bar{\tilde Y}_{0:t-1,a}^{\mathrm{unobs}}
&:=
\frac1u\sum_{v=0}^{t-1}\sum_{j=1}^{b_vn}
(1-\xi_{v,j,a})\tilde Y_{v,j,a}.
\end{align*}
The pooled proxy mean decomposes exactly as
\begin{align*}
\bar{\tilde Y}_{0:t-1,a}^{\mathrm{all}}
&=\frac{m}{m+u}\bar{\tilde Y}_{0:t-1,a}^{\mathrm{obs}}
+\frac{u}{m+u}\bar{\tilde Y}_{0:t-1,a}^{\mathrm{unobs}}.
\end{align*}
Substitution into the posterior-mean identity gives
\begin{align*}
\frac{\mu_{n,t,a}}{\sqrt n}
&=\bar Y_{0:t-1,a}^{\mathrm{obs}}
+\frac{u}{m+u}\lambda_a
\left(
\bar{\tilde Y}_{0:t-1,a}^{\mathrm{unobs}}
-\bar{\tilde Y}_{0:t-1,a}^{\mathrm{obs}}
\right).
\end{align*}
This is the usual disjoint-sample PPI++ mean-estimation formula
\citep{angelopoulos2023ppiplus}, with tuning coefficient
\(u\lambda_a/(m+u)\), rather than \(\lambda_a\).
If \(u=0\), the all-unit and observed proxy means coincide, so the identity
reduces to \(\mu_{n,t,a}/\sqrt n=\bar Y_{0:t-1,a}^{\mathrm{obs}}\)
without requiring an unlabeled-only mean.

To relate this coefficient to oracle tuning, consider separately a static
i.i.d.\ design with fixed sample sizes \(m,u>0\), independent labeled and
unlabeled samples, and known moments. For any fixed \(c\in\mathbb R\),
\begin{align*}
&\Var\left[
\bar Y_{0:t-1,a}^{\mathrm{obs}}
+c\left(
\bar{\tilde Y}_{0:t-1,a}^{\mathrm{unobs}}
-\bar{\tilde Y}_{0:t-1,a}^{\mathrm{obs}}
\right)\right]
\\
&\qquad=
\frac{s_a^2}{m}
-\frac{2c\lambda_a\tilde s_a^2}{m}
+c^2\tilde s_a^2\left(\frac1m+\frac1u\right).
\end{align*}
This strictly convex quadratic is minimized at
\(c=u\lambda_a/(m+u)\). Thus the coefficient is the unconstrained oracle
PPI++ coefficient in this static comparison. Under adaptive allocation, the
pooled-to-unlabeled identity remains exact for every realized dataset, but
this calculation does not establish oracle optimality, unbiasedness,
efficiency, or confidence-interval validity for adaptively collected data.
Nor does fixed known-moment tuning imply equality with a PPI++
implementation that estimates its tuning coefficient from data.

\paragraph{Working-posterior variance comparison.}
The same update calculation verifies the variance reduction invoked after
the theorem. Define
\(\sigma_{n,t,a}^2:=(\sigma_{n,t,a}^{(1)})^2+
\lambda_a^2(\sigma_{n,t,a}^{(2)})^2\), using the factorization of the
working posterior. Then
\begin{align*}
\sigma_{n,t,a}^2
&=
\frac{s_{a,\res}^2}{\sum_{v=0}^{t-1}\hat\pi_{v,a}b_v}
+\lambda_a^2\frac{\tilde s_a^2}{\sum_{v=0}^{t-1}b_v}.
\end{align*}
With a flat prior, known \(s_a^2\), and the same realized allocation,
the true-outcome-only Gaussian working-posterior variance is
\begin{align*}
\sigma_{n,t,a,\mathrm{obs}}^2
&=\frac{s_a^2}{\sum_{v=0}^{t-1}\hat\pi_{v,a}b_v}.
\end{align*}
Using \(s_{a,\res}^2=s_a^2-\lambda_a^2\tilde s_a^2\) and
\(\lambda_a^2\tilde s_a^2=\varrho_a^2s_a^2\) gives
\begin{align*}
\sigma_{n,t,a,\mathrm{obs}}^2-\sigma_{n,t,a}^2
&=
\varrho_a^2s_a^2
\left(
\frac1{\sum_{v=0}^{t-1}\hat\pi_{v,a}b_v}
-
\frac1{\sum_{v=0}^{t-1}b_v}
\right)
\ge0,
\end{align*}
because \(\sum_{v=0}^{t-1}b_v\ge
\sum_{v=0}^{t-1}\hat\pi_{v,a}b_v\).
Equality holds if \(\varrho_a=0\) or all units in batches
\(0,\ldots,t-1\) were assigned to arm \(a\); otherwise, the variance
reduction is strict.
This compares variances assigned by the two working posteriors under the
same realized allocation, not the frequentist variances of their estimators
under adaptive sampling.

\subsection{Lemma for Theorem~\ref{thm:rho-return-comparison}}
\label{app:lemma-rho-sample-augmentation}

\begin{lemma}[Sample augmentation under a fixed RHO allocation]
\label{lem:rho-sample-augmentation}
For any fixed pre-LLM state \(x_t^-\) and allocation
\(\bar\rho\in\DeltaK\),
\[
\widetilde V_t^{\mathdelayedmethod}(\bar\rho,x_t^-)
\ge
\widetilde V_t^{\mathrm{RHO}}(\bar\rho,x_t^-).
\]
\end{lemma}

\begin{proof}
Fix an arm \(a\) and an allocation \(\bar\rho\). For the proof, define
\[
\begin{gathered}
m_{t,a}:=\mu_{t,a}^{(1)}+\lambda_a\mu_{t,a}^{(2)},\qquad
q_a:=B_t\bar\rho_a,\\
P_a:=(\sigma_{t,a}^{(1)})^2,\qquad
Q_a:=\lambda_a^2(\sigma_{t,a}^{(2)})^2,\qquad
\tau_{t,a}^2:=P_a+Q_a,\\
r_a:=s_{a,\res}^2,\qquad
u_a:=\lambda_a^2\tilde s_a^2.
\end{gathered}
\]
Thus \(s_a^2=r_a+u_a\). We first consider the nondegenerate case
\(P_a,Q_a,r_a,u_a>0\); the case \(\lambda_a=0\), for which \(Q_a=u_a=0\),
follows by continuity of the
Gaussian posterior variances and the expected maximum in these variance
parameters.

Under RHO, the terminal posterior variance of \(h_a\) after allocating
the normalized budget \(q_a\) to arm \(a\) is
\[
v_a^{\mathrm{RHO}}
=
\left[
(\tau_{t,a}^2)^{-1}
+
\frac{q_a}{r_a+u_a}
\right]^{-1}.
\]
Under \delayedmethod{}, the residual component receives physical precision
\(q_a/r_a\), while the proxy component receives all-unit
precision \(B_t/u_a\). Its terminal posterior variance for \(h_a\) is therefore
\[
v_a^{\mathdelayedmethod}
=
\left(P_a^{-1}+\frac{q_a}{r_a}\right)^{-1}
+
\left(Q_a^{-1}+\frac{B_t}{u_a}\right)^{-1}.
\]
If \(q_a=0\), then
\[
v_a^{\mathdelayedmethod}
=
P_a+\left(Q_a^{-1}+\frac{B_t}{u_a}\right)^{-1}
\le P_a+Q_a
=v_a^{\mathrm{RHO}}.
\]
Suppose henceforth that \(q_a>0\). Since \(B_t\ge q_a\),
\[
\left(Q_a^{-1}+\frac{B_t}{u_a}\right)^{-1}
\le
\left(Q_a^{-1}+\frac{q_a}{u_a}\right)^{-1}.
\]
Consider an experiment that observes the residual and proxy
components separately with the same normalized information budget \(q_a\):
\[
Y_1=\theta_a^{(1)}+\eta_1,\qquad
Y_2=\lambda_a\theta_a^{(2)}+\eta_2,
\]
where \(\eta_1\sim\Normal(0,r_a/q_a)\) and
\(\eta_2\sim\Normal(0,u_a/q_a)\) are independent of each other and of the
parameters. The corresponding posterior
variance is
\[
\Var(h_a\mid Y_1,Y_2)
=
\left(P_a^{-1}+\frac{q_a}{r_a}\right)^{-1}
+
\left(Q_a^{-1}+\frac{q_a}{u_a}\right)^{-1}.
\]
The RHO observation is the sum
\[
Y_1+Y_2=h_a+\eta_1+\eta_2,
\]
whose noise variance is \((r_a+u_a)/q_a\), and hence
\[
\Var(h_a\mid Y_1+Y_2)=v_a^{\mathrm{RHO}}.
\]
Because \(Y_1+Y_2\) is a measurable function of \((Y_1,Y_2)\), the law of
total variance, conditional on \(Y_1+Y_2\), gives
\[
\begin{aligned}
\Var(h_a\mid Y_1+Y_2)
&=
\E\!\left[
\Var(h_a\mid Y_1,Y_2)
\mid Y_1+Y_2
\right]
\\
&\quad+
\Var\!\left(
\E[h_a\mid Y_1,Y_2]
\mid Y_1+Y_2
\right).
\end{aligned}
\]
The Gaussian conditional variances in this display are deterministic, and the
second term is nonnegative. Therefore
\[
\Var(h_a\mid Y_1,Y_2)\le \Var(h_a\mid Y_1+Y_2).
\]
Together with the previous precision inequality, this proves
\[
v_a^{\mathdelayedmethod}\le v_a^{\mathrm{RHO}}.
\]

Under a fixed RHO allocation, conditional on the current state, the terminal
posterior mean has variance equal to the reduction in Gaussian posterior variance.
Consequently, the two
fixed-allocation returns in Section~\ref{sec:theory} can be written as
\[
\widetilde V_t^{\mathrm{RHO}}(\bar\rho,x_t^-)
=
\E\left[\max_{a\in[K]}
\left\{m_{t,a}+\Gamma_a^{\mathrm{RHO}}\right\}\right],
\]
and
\[
\widetilde V_t^{\mathdelayedmethod}(\bar\rho,x_t^-)
=
\E\left[\max_{a\in[K]}
\left\{m_{t,a}+\Gamma_a^{\mathdelayedmethod}\right\}\right],
\]
where the innovations are independent across arms and satisfy
\[
\Gamma_a^{\mathrm{RHO}}
\sim
\Normal\left(0,\tau_{t,a}^2-v_a^{\mathrm{RHO}}\right),
\qquad
\Gamma_a^{\mathdelayedmethod}
\sim
\Normal\left(0,\tau_{t,a}^2-v_a^{\mathdelayedmethod}\right).
\]
Since \(v_a^{\mathdelayedmethod}\le v_a^{\mathrm{RHO}}\), construct a coupling
explicitly as follows. Let the \(\zeta_a\)'s be independent across arms and
independent of \(\Gamma^{\mathrm{RHO}}\), with
\[
\zeta_a
\sim
\Normal\left(
0,
v_a^{\mathrm{RHO}}-v_a^{\mathdelayedmethod}
\right).
\]
Then
\[
\Gamma_a^{\mathdelayedmethod}
\stackrel{d}{=}
\Gamma_a^{\mathrm{RHO}}+\zeta_a,
\]
jointly across arms. The map
\(z\mapsto\max_{a\in[K]}\{m_{t,a}+z_a\}\) is convex. Jensen's inequality then
gives
\[
\E_\zeta\left[
\max_{a\in[K]}
\left\{m_{t,a}+\Gamma_a^{\mathrm{RHO}}+\zeta_a\right\}
\ \middle|\ \Gamma^{\mathrm{RHO}}
\right]
\ge
\max_{a\in[K]}
\left\{m_{t,a}+\Gamma_a^{\mathrm{RHO}}\right\}.
\]
Taking expectation over \(\Gamma^{\mathrm{RHO}}\) proves the lemma.
\end{proof}

\subsection{Proof of Theorem~\ref{thm:rho-return-comparison}}
\label{app:proof-rho-return-comparison}

\begin{proof}
For a fixed \(\bar\rho\), the expected returns introduced in
Section~\ref{sec:theory} are
\begingroup

\[
\widetilde V_t^{\mathrm{RHO}}(\bar\rho,x_t^-)=\E\left[\max_{a\in[K]}\left\{\mu_{t,a}^{(1)}+\lambda_a\mu_{t,a}^{(2)}+\sqrt{\frac{\left((\sigma_{t,a}^{(1)})^2+\lambda_a^2(\sigma_{t,a}^{(2)})^2\right)^2\bar\rho_a B_t}{s_a^2+\left((\sigma_{t,a}^{(1)})^2+\lambda_a^2(\sigma_{t,a}^{(2)})^2\right)\bar\rho_aB_t}}Z_a\right\}\right].
\]
\[
\widetilde V_t^{\mathdelayedmethod}(\bar\rho,x_t^-)
=\E\Bigg[\max_{a\in[K]}\Bigg\{
\begin{aligned}
&\mu_{t,a}^{(1)}+\lambda_a\mu_{t,a}^{(2)}
+\sqrt{\frac{(\sigma_{t,a}^{(1)})^4\bar\rho_a B_t}
{s_{a,\res}^2+(\sigma_{t,a}^{(1)})^2\bar\rho_aB_t}}Z_a^G\\
&+\lambda_a\sqrt{\frac{(\sigma_{t,a}^{(2)})^4B_t}
{\tilde s_a^2+(\sigma_{t,a}^{(2)})^2B_t}}Z_a^J
\end{aligned}
\Bigg\}\Bigg].
\]
\[
\widetilde V_t^{\mathmethod}(\bar\rho,x_t^+)
=\E\Bigg[\max_{a\in[K]}\Bigg\{
\begin{aligned}
&\mu_{t,a}^{(1)}+\lambda_a\mu_{t+1,a}^{(2)}
+\sqrt{\frac{(\sigma_{t,a}^{(1)})^4\bar\rho_a B_t}
{s_{a,\res}^2+(\sigma_{t,a}^{(1)})^2\bar\rho_aB_t}}Z_a^G\\
&+\lambda_a\sqrt{\frac{(\sigma_{t+1,a}^{(2)})^4B_{t+1}}
{\tilde s_a^2+(\sigma_{t+1,a}^{(2)})^2B_{t+1}}}Z_a^J
\end{aligned}
\Bigg\}\Bigg].
\]
\endgroup
All Gaussian innovations in these expressions are independent standard
normals across arms and signal types. The RHO comparator starts from the
same marginal posterior of \(h\); it ignores future proxies, not any
information already encoded in that common starting state.
Lemma~\ref{lem:rho-sample-augmentation} holds for every fixed
\(\bar\rho\in\DeltaK\). Maximizing its two sides over \(\bar\rho\) yields the
first inequality in Theorem~\ref{thm:rho-return-comparison}.

For the second inequality, define
\[
F(\bar\rho,J_t)
:=
\widetilde V_t^{\mathmethod}
\bigl(\bar\rho,x_t^+(J_t)\bigr).
\]
We first show that, for every fixed \(\bar\rho\),
\begin{equation}
\label{eq:rhope-delayed-average}
\E_{J_t}\left[
F(\bar\rho,J_t)\mid x_t^-
\right]
=
\widetilde V_t^{\mathdelayedmethod}(\bar\rho,x_t^-).
\end{equation}
For this purpose, let
\[
L_{t,a}:=(\sigma_{t,a}^{(2)})^2,\qquad
d_a:=\tilde s_a^2.
\]
Conditional on \(x_t^-\), the current proxy update satisfies
\[
\mu_{t+1,a}^{(2)}
\stackrel{d}{=}
\mu_{t,a}^{(2)}
+
\sqrt{\frac{L_{t,a}^2b_t}{d_a+L_{t,a}b_t}}\,
Z_{t,a}^{J,0},
\]
where \(Z_{t,a}^{J,0}\) is standard normal, and
\[
L_{t+1,a}
:=
(\sigma_{t+1,a}^{(2)})^2
=
\frac{L_{t,a}d_a}{d_a+L_{t,a}b_t}.
\]
The current proxy innovation and the future proxy innovation in the
\method{} return are independent. Their combined variance, before multiplying
by \(\lambda_a^2\), is
\[
\begin{aligned}
&
\frac{L_{t,a}^2b_t}{d_a+L_{t,a}b_t}
+
\frac{L_{t+1,a}^2B_{t+1}}
{d_a+L_{t+1,a}B_{t+1}}
\\
&\qquad=
\frac{L_{t,a}^2B_t}{d_a+L_{t,a}B_t},
\end{aligned}
\]
where the equality follows by substituting the displayed expression for
\(L_{t+1,a}\) and using \(B_t=b_t+B_{t+1}\). Therefore,
\[
\begin{aligned}
&
\lambda_a\mu_{t+1,a}^{(2)}
+
\lambda_a
\sqrt{
\frac{(\sigma_{t+1,a}^{(2)})^4B_{t+1}}
{\tilde s_a^2+(\sigma_{t+1,a}^{(2)})^2B_{t+1}}
}\,Z_a^{J,+}
\\
&\qquad\stackrel{d}{=}
\lambda_a\mu_{t,a}^{(2)}
+
\lambda_a
\sqrt{
\frac{(\sigma_{t,a}^{(2)})^4B_t}
{\tilde s_a^2+(\sigma_{t,a}^{(2)})^2B_t}
}\,Z_a^J,
\end{aligned}
\]
where \(Z_a^{J,+}\) is independent of \(Z_{t,a}^{J,0}\), and \(Z_a^J\) is a
standard normal variable. This equality holds jointly across arms. The
residual innovation in \(F(\bar\rho,J_t)\) is independent of \(J_t\), so the
last display identifies the joint distribution inside the expected maximum
with that in
\(\widetilde V_t^{\mathdelayedmethod}(\bar\rho,x_t^-)\). This proves
\eqref{eq:rhope-delayed-average}.

For each \(J_t\), the map \(\bar\rho\mapsto F(\bar\rho,J_t)\) is continuous
on the compact simplex \(\DeltaK\). Hence the maximum over \(\DeltaK\) is
attained; moreover, the resulting maximum is measurable in \(J_t\) by the
measurable maximum theorem.

For every realization of \(J_t\) and every fixed \(\bar\rho\),
\[
\max_{\bar\rho'\in\DeltaK}F(\bar\rho',J_t)
\ge
F(\bar\rho,J_t).
\]
Taking the conditional expectation given \(x_t^-\) and then maximizing the
right-hand side over \(\bar\rho\) gives
\[
\begin{aligned}
\E_{J_t}\left[
\max_{\bar\rho'\in\DeltaK}F(\bar\rho',J_t)
\ \middle|\ x_t^-
\right]
&\ge
\max_{\bar\rho\in\DeltaK}
\E_{J_t}\left[
F(\bar\rho,J_t)\mid x_t^-
\right]
\\
&=
\max_{\bar\rho\in\DeltaK}
\widetilde V_t^{\mathdelayedmethod}(\bar\rho,x_t^-),
\end{aligned}
\]
where the equality follows from \eqref{eq:rhope-delayed-average}. Substituting the
definition of \(F\) yields the second inequality in the theorem.
\end{proof}

\subsection{Proof of Proposition~\ref{prop:estimation-gap}}
\label{app:proof-estimation-gap}

\paragraph{Notation.}
Let \(\Fcal_t^-\) be the sigma-field generated by the observations before
\(J_t\), let \(\Fcal_t^+:=\sigma(\Fcal_t^-,J_t)\), and let \(\Fcal_T\) be
the terminal information. For a replanning policy \(\pi\) with executed
allocations \(\rho_0,\ldots,\rho_{T-1}\), let \(\Fcal_t^\pi\) denote the
information available when \(\rho_t\) is chosen: \(\Fcal_t^+\) for
\method{} and \(\Fcal_t^-\) for \delayedmethod{} and RHO; if the policy
randomizes (for instance through optimizer initializations), the
randomization variables are included. For \(\bar\rho\in\DeltaK\), the
\emph{static value} of \(\pi\) at batch \(t\) is
\[
\widetilde V_t^\pi(\bar\rho,\Fcal_t^\pi)
:=
\E\bigl[h_{\hat a_T}\mid\Fcal_t^\pi;\ \rho_v=\bar\rho\ \text{for all } v\ge t\bigr],
\]
the conditional expected terminal reward, under the true data-generating
mechanism, when the allocation is held fixed at \(\bar\rho\) from batch \(t\)
onward and the terminal recommendation follows the rule of \(\pi\). Within
the Gaussian sequential experiment with prior \(\nu\) and the posterior-mean
recommendation, \(\E[h_{\hat a_T}\mid\Fcal_T]=\max_a\mu_{T,a}\), so the
static value coincides with the fixed-allocation return of
Section~\ref{subsec:llm-benefits},
\(\widetilde V_t^\pi(\bar\rho,\Fcal_t^\pi)=\widetilde V_t^\pi(\bar\rho,x_t^\pm)\)
(Lemma~\ref{lem:rho} and Appendix~\ref{app:rho-proof}). We write
\(\mathrm{unif}:=(1/K,\ldots,1/K)\) and let \(\pi^{\mathrm{unif}}\) be the
static design that allocates \(\mathrm{unif}\) in every batch and uses the
same estimator and recommendation rule as \(\pi\). For RHO, the outcome
signal is \(G^Y_{t,a}\sim\Normal(\pi_{t,a}h_a,\pi_{t,a}s_a^2/b_t)\),
independent across arms and batches given \(h\) and the allocation
\citep{che2023adaptive}, and the prior of RHO is the marginal
\(h_a\sim\Normal(\mu_{0,a},\sigma_{0,a}^2)\) of \(\nu\).

The proofs use only the following property of the executed allocations.

\begin{definition}[Monotone replanning policy]
\label{def:monotone}
A replanning policy \(\pi\) (RHO, \textup{\delayedmethod{}},
\textup{\method{}}, or a plug-in variant) is \emph{monotone} if
\(\rho_0=\mathrm{unif}\) and, for every \(t\ge1\),
\(\widetilde V_t^\pi(\rho_t,\Fcal_t^\pi)\ge\widetilde V_t^\pi(\rho_{t-1},\Fcal_t^\pi)\).
\end{definition}

In the Gaussian sequential experiment with prior \(\nu\), the static value
is the fixed-allocation return, so exact maximization in
\Eqref{eq:rho-single-simplex} (or of the corresponding objective for
\delayedmethod{} and RHO) implies the second condition, and a uniform
first batch implies the first. Algorithms~\ref{alg:rhope}
and~\ref{alg:rhope-delayed} optimize the first batch only for generality;
throughout the paper, and in every experiment, the first batch is uniform.
Hence the \delayedmethod{} and RHO policies of
Proposition~\ref{prop:estimation-gap} are monotone, and the definition
isolates what the proof uses. For a plug-in planner, which maximizes a
static value computed under estimated moments, the condition is an
assumption (Appendix~\ref{app:estimated-moments-theory}).

\begin{lemma}[Rollout improvement]
\label{lem:rollout}
Fix \(T\) and \((b_t)\), assume \(\E|h_a|<\infty\) for all \(a\), and let
\(\pi\) be monotone. Then
\[
\E^\pi[h_{\hat a_T}]
\;\ge\;
\E\bigl[\widetilde V_0^\pi(\mathrm{unif},\Fcal_0^\pi)\bigr],
\]
that is, \(\mathfrak R_T(\pi,\nu)\le\mathfrak R_T(\pi^{\mathrm{unif}},\nu)\).
\end{lemma}

\begin{proof}
For \(t\le T-1\) let \(M_t:=\widetilde V_t^\pi(\rho_t,\Fcal_t^\pi)\), and let
\(M_T:=\E[h_{\hat a_T}\mid\Fcal_T]\). Fix \(t\in\{1,\ldots,T\}\). The
open-loop continuation that holds \(\rho_{t-1}\) fixed from batch \(t-1\)
onward executes \(\rho_{t-1}\) in batch \(t-1\), and so does \(\pi\) itself,
because \(\rho_{t-1}\) is \(\Fcal_{t-1}^\pi\)-measurable. Hence, on the same
probability space, the information available to this continuation at the
decision time of batch \(t\) is \(\Fcal_t^\pi\) (for \method{} both contain
\(G_{t-1}\) and \(J_t\)), and from \(\Fcal_t^\pi\) onward it coincides with
the open-loop continuation of \(\rho_{t-1}\) started at batch \(t\). The
tower property therefore gives
\[
M_{t-1}
=
\widetilde V_{t-1}^\pi(\rho_{t-1},\Fcal_{t-1}^\pi)
=
\E\bigl[\widetilde V_t^\pi(\rho_{t-1},\Fcal_t^\pi)\mid\Fcal_{t-1}^\pi\bigr],
\]
where for \(t=T\) the inner quantity is \(\E[h_{\hat a_T}\mid\Fcal_T]=M_T\),
since no batches remain and the static value no longer depends on
\(\bar\rho\). For \(t\le T-1\), the monotonicity hypothesis gives
\(\widetilde V_t^\pi(\rho_{t-1},\Fcal_t^\pi)\le\widetilde V_t^\pi(\rho_t,\Fcal_t^\pi)=M_t\).
Thus \(M_{t-1}\le\E[M_t\mid\Fcal_{t-1}^\pi]\) for every
\(t\in\{1,\ldots,T\}\): \((M_t)_{t\le T}\) is a submartingale, integrable
because \(\E|h_a|<\infty\). Taking expectations and iterating,
\[
\E^\pi[h_{\hat a_T}]=\E M_T\ge\E M_0
=\E\,\widetilde V_0^\pi(\rho_0,\Fcal_0^\pi)
=\E\,\widetilde V_0^\pi(\mathrm{unif},\Fcal_0^\pi),
\]
where the last step uses \(\rho_0=\mathrm{unif}\). The right-hand side is the
expected terminal reward of \(\pi^{\mathrm{unif}}\); subtracting both sides
from \(\E[\max_ah_a]\) yields the regret inequality.
\end{proof}

\begin{lemma}[Exact regret of exchangeable static designs]
\label{lem:exchangeable-regret}
Let \(h_a\overset{\mathrm{iid}}{\sim}\Normal(m_0,\tau^2)\), \(a\in[K]\).
\textup{(a)} Suppose the data \(\mathcal D\) and \(h\) are jointly Gaussian,
the posterior means \(\mu_{T,a}:=\E[h_a\mid\mathcal D]\) are i.i.d.\ across
arms, and \(\Var(h_a\mid\mathcal D)=\sigma^2\in[0,\tau^2]\) is a constant
depending neither on \(a\) nor on the data. Then, for
\(\hat a\in\argmax_a\mu_{T,a}\),
\[
\E[\max_ah_a]-\E[h_{\hat a}]=\mathcal R_K(\sigma^2;\tau).
\]
\textup{(b)} Suppose \(\hat\mu_a=h_a+\eta_a\) with
\(\eta_a\overset{\mathrm{iid}}{\sim}\Normal(0,w)\) independent of \(h\), and
\(\hat a\in\argmax_a\hat\mu_a\). Then, with \(g(x):=1-(1+x)^{-1/2}\) for
\(x\ge0\),
\[
\E[\max_ah_a]-\E[h_{\hat a}]
=\tau c_K\,g(w/\tau^2)
=\mathcal R_K\bigl(\tau^2w/(\tau^2+w);\tau\bigr).
\]
\end{lemma}

\begin{proof}
(a) Joint Gaussianity makes \(\mu_{T,a}\) an affine function of the data,
hence Gaussian with \(\E\mu_{T,a}=m_0\). By the law of total variance and
the constancy of the posterior variance,
\(\Var\mu_{T,a}=\Var h_a-\E\Var(h_a\mid\mathcal D)=\tau^2-\sigma^2\).
Hence \(\mu_{T,a}\overset{\mathrm{iid}}{\sim}\Normal(m_0,\tau^2-\sigma^2)\)
and \(\E\max_a\mu_{T,a}=m_0+c_K\sqrt{\tau^2-\sigma^2}\) (for
\(\sigma^2=\tau^2\) the posterior means are degenerate and the formula still
holds). Since \(\hat a\) is \(\sigma(\mathcal D)\)-measurable,
\[
\E[h_{\hat a}]
=\sum_a\E\bigl[h_a\ind\{\hat a=a\}\bigr]
=\sum_a\E\bigl[\mu_{T,a}\ind\{\hat a=a\}\bigr]
=\E[\mu_{T,\hat a}]
=\E[\max_a\mu_{T,a}].
\]
Finally, \(\E\max_ah_a=m_0+c_K\tau\); subtracting gives
\(c_K(\tau-\sqrt{\tau^2-\sigma^2})=\mathcal R_K(\sigma^2;\tau)\).

(b) The pair \((h,\hat\mu)\) is jointly Gaussian with independent
coordinates across arms. Gaussian conditioning gives
\(\E[h_a\mid\hat\mu]=m_0+\frac{\tau^2}{\tau^2+w}(\hat\mu_a-m_0)\) and
\(\Var(h_a\mid\hat\mu)=\tau^2w/(\tau^2+w)\), a constant; the posterior means
are i.i.d.\ across arms, and
\(\argmax_a\E[h_a\mid\hat\mu]=\argmax_a\hat\mu_a\) because the shrinkage
factor is positive and common to all arms. Part~(a) then yields the regret
\(\mathcal R_K(\tau^2w/(\tau^2+w);\tau)\). Since
\(\tau^2-\tau^2w/(\tau^2+w)=\tau^4/(\tau^2+w)\),
\[
\mathcal R_K\bigl(\tau^2w/(\tau^2+w);\tau\bigr)
=c_K\Bigl(\tau-\frac{\tau^2}{\sqrt{\tau^2+w}}\Bigr)
=\tau c_K\bigl[1-(1+w/\tau^2)^{-1/2}\bigr]
=\tau c_K\,g(w/\tau^2).
\qedhere
\]
\end{proof}

\begin{proof}[Proof of Proposition~\ref{prop:estimation-gap}]
Throughout, arms are exchangeable, so
\(h_a\overset{\mathrm{iid}}{\sim}\Normal(m_0,\tau^2)\) with
\(m_0:=\mu_0^{(1)}+\lambda\mu_0^{(2)}\).

\emph{Step 1: terminal posterior variances and exact regret of the uniform
static designs.} Under \(\pi_{t,a}=1/K\) for all \(t\), the posterior
updates in Equations~\ref{eq:posterior-update-proxy}
and~\ref{eq:posterior-update-residual} and \(\sum_tb_t=1\) give
\[
(\sigma_{T,a}^{(1)})^{-2}
=(\sigma_0^{(1)})^{-2}+\sum_t\frac{b_t/K}{s_{\res}^2}
=(\sigma_0^{(1)})^{-2}+\frac{1}{Ks_{\res}^2},
\qquad
(\sigma_{T,a}^{(2)})^{-2}=(\sigma_0^{(2)})^{-2}+\frac{1}{\tilde s^2},
\]
so \(\sigma_{T,a}^2=\sigma_{T,\mathrm{PE}}^2\), a constant not depending on
the data. The posteriors of \(\theta^{(1)}\) and \(\theta^{(2)}\) are
independent (independent priors and signals),
\(h_a=\theta_a^{(1)}+\lambda\theta_a^{(2)}\) has a Gaussian posterior, the
arm posteriors are i.i.d.\ by exchangeability and the across-arm independence
of the signals, and the data and \(h\) are jointly Gaussian. By
Lemma~\ref{lem:exchangeable-regret}(a), the uniform static design with
proxies (the same for \method{} and \delayedmethod{}, since the allocation
is fixed) has Bayes simple regret \(\mathcal R_K(\sigma_{T,\mathrm{PE}}^2;\tau)\).
For RHO, the uniform static design observes
\(G^Y_{t,a}\sim\Normal(h_a/K,s^2/(Kb_t))\) with prior
\(h_a\sim\Normal(m_0,\tau^2)\); since the noise is independent of the
parameters, this model is the true law of the outcome signals, the terminal
posterior precision is \(\tau^{-2}+\sum_tb_t/(Ks^2)=\tau^{-2}+1/(Ks^2)\),
and the exact regret is \(\mathcal R_K(\sigma_{T,0}^2;\tau)\).

\emph{Step 2: upper bounds.} Within the Gaussian sequential experiment with
prior \(\nu\), each policy is well specified in its own information
structure (\delayedmethod{} uses \(\nu\); RHO uses the marginal prior of
\(h\), and the true law of the outcome signals is its model), so the static
value equals the fixed-allocation return; both policies start with a
uniform batch and replan by exact maximization, so they are monotone
(Definition~\ref{def:monotone}). Lemma~\ref{lem:rollout} gives
\(\mathfrak R_T(\pi,\nu)\le\mathfrak R_T(\pi^{\mathrm{unif}},\nu)\) for
\(\pi\in\{\mathdelayedmethod,\RHO\}\), and Step~1 evaluates the right-hand
sides.

\emph{Step 3: \(\sigma_{T,\mathrm{PE}}^2\le\sigma_{T,0}^2\).} Apply the
inequality \(v_a^{\mathdelayedmethod}\le v_a^{\mathrm{RHO}}\) established in
the proof of Lemma~\ref{lem:rho-sample-augmentation} at \(t=0\) with
\(\bar\rho=\mathrm{unif}\): there \(q_a=1/K\), \(P_a=(\sigma_0^{(1)})^2\),
\(Q_a=\lambda^2(\sigma_0^{(2)})^2\), \(r_a=s_{\res}^2\), and
\(u_a=\lambda^2\tilde s^2\), so that
\(v_a^{\mathdelayedmethod}=\sigma_{T,\mathrm{PE}}^2\) and
\(v_a^{\mathrm{RHO}}=\sigma_{T,0}^2\).

\emph{Step 4: ratio of the two bounds.} Since
\(\tau-\sqrt{\tau^2-\sigma^2}=\sigma^2/(\tau+\sqrt{\tau^2-\sigma^2})\),
\[
\frac{\mathcal R_K(\sigma_{T,\mathrm{PE}}^2;\tau)}
{\mathcal R_K(\sigma_{T,0}^2;\tau)}
=\frac{\sigma_{T,\mathrm{PE}}^2}{\sigma_{T,0}^2}\cdot
\frac{\tau+\sqrt{\tau^2-\sigma_{T,0}^2}}
{\tau+\sqrt{\tau^2-\sigma_{T,\mathrm{PE}}^2}}.
\]
Under a flat prior, that is, as
\((\sigma_0^{(1)})^{-2},(\sigma_0^{(2)})^{-2}\to0\), we have
\(\sigma_{T,\mathrm{PE}}^2\to Ks_{\res}^2+\lambda^2\tilde s^2\),
\(\sigma_{T,0}^2\to Ks^2\), and \(\tau^2\to\infty\), so the second
factor tends to one (its numerator and denominator are both between
\(\tau\) and \(2\tau\)) and the ratio tends to
\((Ks_{\res}^2+\lambda^2\tilde s^2)/(Ks^2)\). Finally,
\(Ks_{\res}^2+\lambda^2\tilde s^2=Ks^2(1-\varrho^2)+\varrho^2s^2
=Ks^2\bigl[1-\varrho^2(1-1/K)\bigr]\).
\end{proof}

\subsection{Proof of Proposition~\ref{prop:advance-gap}}
\label{app:proof-advance-gap}

\begin{lemma}[Gaussian maximal inequality]
\label{lem:gaussian-max}
Let \(X_1,\ldots,X_N\) be centered Gaussian random variables, not
necessarily independent, with standard deviations at most \(\sigma\). Then
\(\E\max_{i\le N}|X_i|\le\sigma\sqrt{2\log(2N)}\).
\end{lemma}

\begin{proof}
Let \((Y'_1,\ldots,Y'_{2N}):=(X_1,-X_1,\ldots,X_N,-X_N)\), so that
\(\max_i|X_i|=\max_jY'_j\). For any \(\vartheta>0\), Jensen's inequality and
the Gaussian moment generating function give
\[
\E\max_jY'_j
\le\frac1\vartheta\log\sum_{j\le2N}\E e^{\vartheta Y'_j}
\le\frac1\vartheta\log\bigl(2N\,e^{\vartheta^2\sigma^2/2}\bigr)
=\frac{\log(2N)}{\vartheta}+\frac{\vartheta\sigma^2}{2}.
\]
Choosing \(\vartheta=\sqrt{2\log(2N)}/\sigma\) gives the claim.
\end{proof}

\begin{proof}[Proof of Proposition~\ref{prop:advance-gap}]
Condition on \(x_t^-\) throughout and write, as in
Appendix~\ref{app:proof-rho-return-comparison},
\(L_{t,a}:=(\sigma_{t,a}^{(2)})^2\), \(d_a:=\tilde s_a^2\), and
\(L_{t+1,a}=L_{t,a}d_a/(d_a+L_{t,a}b_t)\), which does not depend on
\(J_t\). Let
\[
\mu_a^-:=\mu_{t,a}^{(1)}+\lambda_a\mu_{t,a}^{(2)},\qquad
\mu_a^+:=\mu_{t,a}^{(1)}+\lambda_a\mu_{t+1,a}^{(2)},\qquad
\Delta_a:=\mu_a^+-\mu_a^-=\lambda_a\bigl(\mu_{t+1,a}^{(2)}-\mu_{t,a}^{(2)}\bigr).
\]

\emph{Step 1: distribution of \(\Delta\).} By
\eqref{eq:posterior-update-proxy},
\(\mu_{t+1,a}^{(2)}-\mu_{t,a}^{(2)}=L_{t+1,a}\frac{b_t}{d_a}(J_{t,a}-\mu_{t,a}^{(2)})\).
Given \(x_t^-\), \(J_{t,a}\sim\Normal(\mu_{t,a}^{(2)},L_{t,a}+d_a/b_t)\)
independently across arms (the predictive law:
\(\theta_a^{(2)}\sim\Normal(\mu_{t,a}^{(2)},L_{t,a})\) plus independent
noise). Hence \(\Delta_a\) is centered Gaussian, independent across arms,
with
\[
\Var(\Delta_a)
=\lambda_a^2\frac{L_{t+1,a}^2b_t^2}{d_a^2}\Bigl(L_{t,a}+\frac{d_a}{b_t}\Bigr)
=\lambda_a^2\frac{L_{t,a}^2b_t}{d_a+L_{t,a}b_t}
=\lambda_a^2\bigl(L_{t,a}-L_{t+1,a}\bigr).
\]

\emph{Step 2: the two fixed-allocation returns.} By
\Eqref{eq:rho-single-simplex} and
Appendix~\ref{app:proof-rho-return-comparison},
\[
\begin{aligned}
\widetilde V_t^{\mathmethod}(\bar\rho,x_t^+)
&=\E\max_a\bigl\{\mu_a^++\varsigma_a^G(\bar\rho)Z_a^G+\varsigma_a^{J,+}Z_a^J\bigr\},
\\
\widetilde V_t^{\mathdelayedmethod}(\bar\rho,x_t^-)
&=\E\max_a\bigl\{\mu_a^-+\varsigma_a^G(\bar\rho)Z_a^G+\varsigma_a^{J,-}Z_a^J\bigr\},
\end{aligned}
\]
where
\[
\begin{gathered}
\varsigma_a^G(\bar\rho)^2
=\frac{(\sigma_{t,a}^{(1)})^4\bar\rho_aB_t}{s_{a,\res}^2+(\sigma_{t,a}^{(1)})^2\bar\rho_aB_t},
\\
(\varsigma_a^{J,+})^2
=\lambda_a^2\frac{L_{t+1,a}^2B_{t+1}}{d_a+L_{t+1,a}B_{t+1}},
\qquad
(\varsigma_a^{J,-})^2
=\lambda_a^2\frac{L_{t,a}^2B_t}{d_a+L_{t,a}B_t},
\end{gathered}
\]
none of which depends on \(J_t\). Let
\(\varsigma_a(\bar\rho)^2:=\varsigma_a^G(\bar\rho)^2+(\varsigma_a^{J,+})^2\)
and
\[
\mathcal M_{\bar\rho}(\mu):=\E\max_a\{\mu_a+\varsigma_a(\bar\rho)Z_a\},
\qquad Z_a\overset{\mathrm{iid}}{\sim}\Normal(0,1),
\]
so that \(\widetilde V_t^{\mathmethod}(\bar\rho,x_t^+)=\mathcal M_{\bar\rho}(\mu^-+\Delta)\).
The variance identity in the proof of
Theorem~\ref{thm:rho-return-comparison} reads
\((\varsigma_a^{J,-})^2=\Var(\Delta_a)+(\varsigma_a^{J,+})^2\). Since
\(\Delta\) is independent of \(Z\) and both have independent coordinates,
\(\mu_a^-+\Delta_a+\varsigma_a(\bar\rho)Z_a\) and
\(\mu_a^-+\sqrt{\varsigma_a^G(\bar\rho)^2+(\varsigma_a^{J,-})^2}\,Z_a\) have
the same joint law across arms, and therefore, for every
\(\bar\rho\in\DeltaK\),
\begin{equation}
\label{eq:delayed-as-average}
\widetilde V_t^{\mathdelayedmethod}(\bar\rho,x_t^-)
=\E_\Delta\bigl[\mathcal M_{\bar\rho}(\mu^-+\Delta)\bigr],
\end{equation}
which is \eqref{eq:rhope-delayed-average} in the present notation.

\emph{Step 3: nonnegativity.} For every realization of \(\Delta\) and every
\(\bar\rho\),
\(\max_{\bar\rho'}\mathcal M_{\bar\rho'}(\mu^-+\Delta)\ge\mathcal M_{\bar\rho}(\mu^-+\Delta)\).
Taking expectations, maximizing over \(\bar\rho\), and using
\eqref{eq:delayed-as-average} gives \(\Delta_t^{\mathrm{adv}}(x_t^-)\ge0\).
(The map \(\bar\rho\mapsto\mathcal M_{\bar\rho}(\mu)\) is continuous on the
compact set \(\DeltaK\), so the maximum is attained and is measurable in
\(\Delta\).)

\emph{Step 4: Lipschitz bound.} For any \(\mu,\mu'\),
\(|\max_a(\mu_a+\varsigma_aZ_a)-\max_a(\mu'_a+\varsigma_aZ_a)|\le\|\mu-\mu'\|_\infty\)
pointwise; taking expectations,
\(|\mathcal M_{\bar\rho}(\mu)-\mathcal M_{\bar\rho}(\mu')|\le\|\mu-\mu'\|_\infty\)
uniformly in \(\bar\rho\). Hence
\[
\max_{\bar\rho}\mathcal M_{\bar\rho}(\mu^-+\Delta)
\le\max_{\bar\rho}\mathcal M_{\bar\rho}(\mu^-)+\|\Delta\|_\infty.
\]
On the other hand, with
\(\bar\rho^-\in\argmax_{\bar\rho}\mathcal M_{\bar\rho}(\mu^-)\),
\eqref{eq:delayed-as-average} and the Lipschitz property give
\[
\begin{aligned}
\max_{\bar\rho}\widetilde V_t^{\mathdelayedmethod}(\bar\rho,x_t^-)
&\ge\E_\Delta\mathcal M_{\bar\rho^-}(\mu^-+\Delta)
\\
&\ge\mathcal M_{\bar\rho^-}(\mu^-)-\E\|\Delta\|_\infty
=\max_{\bar\rho}\mathcal M_{\bar\rho}(\mu^-)-\E\|\Delta\|_\infty.
\end{aligned}
\]
Taking the expectation of the first display over \(\Delta\) and subtracting
the second, \(\Delta_t^{\mathrm{adv}}(x_t^-)\le2\,\E\max_a|\Delta_a|\).

\emph{Step 5: the main inequality.} Lemma~\ref{lem:gaussian-max} with
\(N=K\), \(X_a=\Delta_a\), and \(\sigma=\max_a\sqrt{\Var(\Delta_a)}\),
together with
\(\sqrt{\Var(\Delta_a)}=|\lambda_a|L_{t,a}\sqrt{b_t/(d_a+L_{t,a}b_t)}\) from
Step~1, yields the bound in the proposition.

\emph{Step 6: flat prior and order of magnitude.} With zero initial
precisions and \(t\ge1\), let \(C_t:=\sum_{v<t}b_v\) denote the budget
fraction used before batch \(t\). Then
\((\sigma_{t,a}^{(2)})^{-2}=C_t/\tilde s_a^2\), so \(L_{t,a}=\tilde s_a^2/C_t\)
and
\[
\Var(\Delta_a)
=\lambda_a^2\frac{L_{t,a}^2b_t}{d_a+L_{t,a}b_t}
=\lambda_a^2\tilde s_a^2\frac{b_t}{C_t(C_t+b_t)}
=\varrho_a^2s_a^2\frac{b_t}{C_tC_{t+1}}.
\]
With equal batches, \(b_t=1/T\) and \(C_t=t/T\), so
\(\sqrt{\Var(\Delta_a)}=|\varrho_a|s_a\sqrt{T/(t(t+1))}\) and the bound of
Step~5 equals
\(2\sqrt{2\log(2K)}\,\max_a|\varrho_a|s_a\sqrt{T/(t(t+1))}\). Since
\(\sqrt{t/(t+1)}\in[1/\sqrt2,1)\) and \(\log K\le\log(2K)\le2\log K\)
for \(K\ge2\), this equals \(\max_a|\varrho_a|s_a\sqrt{T\log K}/t\) up to
an absolute constant between \(2\) and \(4\), that is, \(\sqrt{T\log K}/t\)
up to a constant depending only on \(\max_a|\varrho_a|s_a\), which is the
order stated in the proposition.
\end{proof}

\subsection{Proof of Proposition~\ref{prop:estimated-moments}}
\label{app:proof-estimated-moments}

Proposition~\ref{prop:estimated-moments} is stated in
Appendix~\ref{app:estimated-moments-theory}; we use its notation
(\(\mathrm{RHOPE}_{\hat\lambda}\), \(\gamma\), \(g\), \(\hat c_{Y\tilde Y}\),
\(n_\lambda\)) and Assumption~\ref{ass:plugin} throughout this subsection,
together with the following abbreviations. \(\mathrm{unif}_{\hat\lambda}\)
and \(\mathrm{unif}_0\) denote the all-uniform static designs that use the
estimator and recommendation rule of \(\mathrm{RHOPE}_{\hat\lambda}\) and of
RHO, respectively. We write
\begin{align*}
w(\hat\lambda)
&:=Ks^2\Bigl\{1-\bigl(1-\tfrac1K\bigr)
\Bigl[\varrho^2-\tfrac{\tilde s^2}{s^2}(\hat\lambda-\lambda)^2\Bigr]\Bigr\},
\qquad
w_0:=w(0)=Ks^2,
\\
\psi
&:=\tfrac{\tilde s^2}{s^2}\,\E(\hat\lambda-\lambda)^2,
\qquad
\bar w_{\hat\lambda}:=\E\,w(\hat\lambda)
=Ks^2\Bigl\{1-\bigl(1-\tfrac1K\bigr)\bigl[\varrho^2-\psi\bigr]\Bigr\},
\\
\psi(n;\varrho,\gamma)
&:=\frac{\gamma-1}{n}+\frac{1}{n-1}-\frac{(n-2)\varrho^2}{n(n-1)},
\end{align*}
so that the first bound of the proposition is \(\tau c_K
g(\bar w_{\hat\lambda}/\tau^2)\), the second is \(\tau c_K g(w_0/\tau^2)\),
and the condition \(\E(\hat\lambda-\lambda)^2<\lambda^2\) reads
\(\psi<\varrho^2\). Here \(w(\hat\lambda)\) is the frequentist variance of
the terminal working-posterior mean \(\mu_{n,T,a}\), as an estimator of
\(h_a\), under \(\mathrm{unif}_{\hat\lambda}\).

\begin{lemma}[Mean squared error of the sample covariance;
\citealp{mani2025nofreelunch}]
\label{lem:mani}
Let \((\epsilon_i,\tilde\epsilon_i)_{i\le n_\lambda}\) be i.i.d.\ with mean zero and
finite fourth moments, \(\Var\epsilon=s^2\), \(\Var\tilde\epsilon=\tilde s^2\),
and \(\Cov(\epsilon,\tilde\epsilon)=\varrho s\tilde s\), and let
\(\hat c:=\frac1{n_\lambda-1}\sum_{i\le n_\lambda}(\epsilon_i-\bar\epsilon)(\tilde\epsilon_i-\bar{\tilde\epsilon})\).
Then
\[
\E(\hat c-\varrho s\tilde s)^2
=\frac1{n_\lambda}\bigl(\E[\epsilon^2\tilde\epsilon^2]-s^2\tilde s^2\bigr)
+\frac{s^2\tilde s^2}{n_\lambda-1}
-\frac{(n_\lambda-2)\varrho^2s^2\tilde s^2}{n_\lambda(n_\lambda-1)}.
\]
\end{lemma}

This is Theorem~4.3 of \citet{mani2025nofreelunch}, whose statement includes
mean terms; the sample covariance is translation invariant, so centering
gives the displayed form. In the notation of
Appendix~\ref{app:estimated-moments-theory}, the lemma applies to the
centered pilot pairs \((\epsilon_i^{\mathrm p},\tilde\epsilon_i^{\mathrm p})\),
which differ from the pilot pairs by a constant vector, so
\(\hat c_{Y\tilde Y}=\hat c\), and with
\(\hat\lambda=\hat c_{Y\tilde Y}/\tilde s^2\) and
\(\lambda=\varrho s\tilde s/\tilde s^2\),
\begin{equation}
\label{eq:psi-closed-form}
\psi
=\frac{\tilde s^2}{s^2}\,\E(\hat\lambda-\lambda)^2
=\frac{\E(\hat c-\varrho s\tilde s)^2}{s^2\tilde s^2}
=\frac{\gamma-1}{n_\lambda}+\frac1{n_\lambda-1}
-\frac{(n_\lambda-2)\varrho^2}{n_\lambda(n_\lambda-1)}
=\psi(n_\lambda;\varrho,\gamma).
\end{equation}

\begin{proof}[Proof of Proposition~\ref{prop:estimated-moments}]
\emph{Step 0: reduction to uniform static designs.}
\(\mathrm{RHOPE}_{\hat\lambda}\) and \(\mathrm{unif}_{\hat\lambda}\) use the
same estimator (the working-posterior mean with zero initial precisions and
coefficient \(\hat\lambda\)) and the same recommendation rule, so
\(\mathrm{unif}_{\hat\lambda}\) is the design \(\pi^{\mathrm{unif}}\) of
Lemma~\ref{lem:rollout}; likewise for RHO and \(\mathrm{unif}_0\). Under
Assumption~\ref{ass:plugin} the information \(\Fcal_t^\pi\) of Lemma~\ref{lem:rollout} also
contains \(\sigma(\hat\lambda)\). Lemma~\ref{lem:rollout} gives
\(\mathfrak R_T(\mathrm{RHOPE}_{\hat\lambda},\nu)\le\mathfrak R_T(\mathrm{unif}_{\hat\lambda},\nu)\)
and \(\mathfrak R_T(\RHO,\nu)\le\mathfrak R_T(\mathrm{unif}_0,\nu)\). It remains to
compute the right-hand sides.

\emph{Step 1: terminal working-posterior mean.} With zero initial
precisions, accumulating the finite-batch updates of
Algorithm~\ref{alg:rhope}, with \(\lambda_a\) replaced by \(\hat\lambda\) in
step~(d) and arbitrary positive variance constants, the variance constants
cancel, and substituting
\(\hat G_{t,a}=\frac1{b_t\sqrt n}\sum_j\xi_{t,j,a}(Y_{t,j,a}-\hat\lambda\tilde Y_{t,j,a})\)
and \(\hat J_{t,a}=\frac1{b_t\sqrt n}\sum_j\tilde Y_{t,j,a}\) yields the
identity of Theorem~\ref{thm:ppi}, which holds for any substituted
coefficient:
\[
\mu_{n,T,a}
=\mu_{n,T,a}^{(1)}+\hat\lambda\mu_{n,T,a}^{(2)}
=\sqrt n\Bigl[\bar Y_{0:T-1,a}^{\mathrm{obs}}
-\hat\lambda\bigl(\bar{\tilde Y}_{0:T-1,a}^{\mathrm{obs}}
-\bar{\tilde Y}_{0:T-1,a}^{\mathrm{all}}\bigr)\Bigr].
\]
Here the balanced assignment makes \(\sum_tb_t\hat\pi_{t,a}=1/K\), so
\(\bar Y_{0:T-1,a}^{\mathrm{obs}}\) and
\(\bar{\tilde Y}_{0:T-1,a}^{\mathrm{obs}}\) average over the \(n/K\) labeled
units of arm \(a\), and \(\bar{\tilde Y}_{0:T-1,a}^{\mathrm{all}}\) over all
\(n\) units.

\emph{Step 2: conditional law given \(\hat\lambda\).} Substituting
\(Y_{t,j,a}=h_a/\sqrt n+\epsilon_{t,j,a}\) and
\(\tilde Y_{t,j,a}=\tilde h_a/\sqrt n+\tilde\epsilon_{t,j,a}\), the
\(\tilde h_a\) terms cancel and
\[
\mu_{n,T,a}-h_a
=\sqrt n\bigl(\bar\epsilon_a^{\mathrm{obs}}-\hat\lambda D_a\bigr),
\qquad
D_a:=\bar{\tilde\epsilon}_a^{\mathrm{obs}}-\bar{\tilde\epsilon}_a^{\mathrm{all}},
\]
where the superscripts obs and all denote averages over the \(n/K\) labeled
units of arm \(a\) and over all \(n\) units, respectively. The assignment
and \(\hat\lambda\) are independent of the noise and of \(h\); given
\(\hat\lambda\) and the assignment, the display is a linear combination of
independent Gaussian noise terms, hence centered Gaussian, independent of
\(h\), and independent across arms. Its variance follows from
\(\Var(\bar\epsilon_a^{\mathrm{obs}})=Ks^2/n\),
\(\Cov(\bar\epsilon_a^{\mathrm{obs}},D_a)=\varrho s\tilde s\,(K-1)/n\), and,
writing
\(D_a=(1-1/K)(\bar{\tilde\epsilon}_a^{\mathrm{obs}}-\bar{\tilde\epsilon}_a^{\mathrm{unobs}})\)
with unobs denoting the \(n(1-1/K)\) unlabeled units of arm \(a\),
\(\Var(D_a)=(1-1/K)^2\tilde s^2\bigl(K/n+1/(n(1-1/K))\bigr)=\tilde s^2(K-1)/n\).
With \(\varrho s\tilde s=\lambda\tilde s^2\),
\[
\begin{aligned}
\Var(\mu_{n,T,a}-h_a\mid\hat\lambda)
&=n\Bigl[\frac{Ks^2}{n}-2\hat\lambda\lambda\tilde s^2\frac{K-1}{n}
+\hat\lambda^2\tilde s^2\frac{K-1}{n}\Bigr]
\\
&=Ks^2-(K-1)\tilde s^2\bigl[\lambda^2-(\hat\lambda-\lambda)^2\bigr]
=w(\hat\lambda),
\end{aligned}
\]
using \((K-1)\tilde s^2\lambda^2=(K-1)\varrho^2s^2=Ks^2(1-1/K)\varrho^2\).
The variance does not depend on which units are labeled, so no further
conditioning on the assignment is needed.

\emph{Step 3: regret of the uniform static designs.} Given \(\hat\lambda\),
\(\mu_{n,T,a}=h_a+\eta_a\) with
\(\eta_a\overset{\mathrm{iid}}{\sim}\Normal(0,w(\hat\lambda))\) independent
of \(h\), and the recommendation is \(\argmax_a\mu_{n,T,a}\).
Lemma~\ref{lem:exchangeable-regret}(b), applied conditionally on
\(\hat\lambda\), and integration over \(\hat\lambda\) give
\[
\mathfrak R_T(\mathrm{unif}_{\hat\lambda},\nu)
=\E_{\hat\lambda}\bigl[\tau c_K\,g\bigl(w(\hat\lambda)/\tau^2\bigr)\bigr].
\]
The terminal statistic of RHO, \(\sqrt n\,\bar Y_{0:T-1,a}^{\mathrm{obs}}\),
is the case \(\hat\lambda\equiv0\), so
\(\mathfrak R_T(\mathrm{unif}_0,\nu)=\tau c_K\,g(w_0/\tau^2)\).

\emph{Step 4: upper bound and threshold.} Since \(g\) is concave, Jensen's
inequality gives
\(\mathfrak R_T(\mathrm{unif}_{\hat\lambda},\nu)
\le\tau c_K\,g\bigl(\E w(\hat\lambda)/\tau^2\bigr)
=\tau c_K\,g(\bar w_{\hat\lambda}/\tau^2)\). Combining with Step~0 proves
the two inequalities. Since \(g\) is strictly increasing,
\(\tau c_K\,g(\bar w_{\hat\lambda}/\tau^2)<\tau c_K\,g(w_0/\tau^2)\) if and
only if \(\bar w_{\hat\lambda}<w_0\), that is,
\(-Ks^2(1-1/K)[\varrho^2-\psi]<0\), that is, \(\psi<\varrho^2\); and
\(\varrho^2=(\tilde s^2/s^2)\lambda^2\) turns this into
\(\E(\hat\lambda-\lambda)^2<\lambda^2\).

\emph{Step 5: closed form of \(\psi\).} For
\(\hat\lambda=\hat c_{Y\tilde Y}/\tilde s^2\), \eqref{eq:psi-closed-form}
gives \(\psi=\psi(n_\lambda;\varrho,\gamma)\), so \(\psi<\varrho^2\) is the
condition on \(\varrho^2\) displayed in the proposition.

\emph{Step 6: bivariate normal pilot pairs.} If
\((\epsilon^{\mathrm p},\tilde\epsilon^{\mathrm p})\) is bivariate normal,
the fourth-moment formula of the normal distribution (Isserlis' theorem)
gives
\(\E[(\epsilon^{\mathrm p})^2(\tilde\epsilon^{\mathrm p})^2]
=s^2\tilde s^2(1+2\varrho^2)\), so \(\gamma=1+2\varrho^2\) and
\[
\psi(n_\lambda;\varrho,1+2\varrho^2)
=\frac{2\varrho^2}{n_\lambda}+\frac1{n_\lambda-1}
-\frac{(n_\lambda-2)\varrho^2}{n_\lambda(n_\lambda-1)}
=\frac{1+\varrho^2}{n_\lambda-1}.
\]
Hence \(\psi<\varrho^2\) if and only if \((n_\lambda-2)\varrho^2>1\).
\end{proof}

\section{Additional Experimental Details}
\label{app:experiments}

\subsection{Synthetic Data Experiments}
\label{app:synthetic-results}

\subsubsection{Details of Synthetic Data Experiments}

Every synthetic experiment has \(K=10\) arms and \(T=10\) equal-sized
batches. The common batch size is \(b_tn\in\{100,1000\}\), so the total
number of experimental units is \(10b_tn\). Conditional on the arm-level
parameters, outcome pairs are independent across batches, units, and arms.
The proxy outcomes are generated synthetically to emulate LLM information;
these experiments do not call an LLM. In all three data-generating processes (DGPs),
\(\varrho\) denotes the Pearson correlation between the true outcome
\(Y_{t,j,a}\) and the proxy outcome \(\tilde Y_{t,j,a}\) for the
same unit and arm; the main correlation grid is
\(\varrho\in\{0.2,0.4,0.6\}\).

\paragraph{Bernoulli DGP.}
For each arm \(a\), we independently draw
\(p_a\sim\mathrm{Beta}(b_tn,b_tn)\). The true and proxy outcomes have the same
Bernoulli marginal distribution:
\begin{align*}
Y_{t,j,a}\mid p_a &\sim \mathrm{Bernoulli}(p_a),
&
\tilde Y_{t,j,a}\mid p_a &\sim \mathrm{Bernoulli}(p_a).
\end{align*}
We first draw \(Y_{t,j,a}\) and then generate \(\tilde Y_{t,j,a}\) according
to
\begin{align*}
\Pr(\tilde Y_{t,j,a}=1\mid Y_{t,j,a}=1,p_a)
&=p_a+\varrho(1-p_a),
&
\Pr(\tilde Y_{t,j,a}=1\mid Y_{t,j,a}=0,p_a)
&=p_a(1-\varrho).
\end{align*}
This construction preserves both Bernoulli marginals and gives
\(\operatorname{Corr}(Y_{t,j,a},\tilde Y_{t,j,a}\mid p_a)=\varrho\).
Thus the true and proxy outcomes share the same arm means and arm
ranking.

\paragraph{Gumbel DGP.}
For each arm \(a\), we independently draw the location parameter \(\ell_a\)
from a Gamma distribution with shape \(b_tn\) and scale \(1/(b_tn)\). Let
\(\beta=\sqrt{6\sigma^2}/\pi\), let \(\gamma_{\mathrm E}\) denote the
Euler--Mascheroni constant, and let \(\Phi\) denote the standard normal
cumulative distribution function. For independent standard normal variables
\(Z_{t,j,a}\) and \(E_{t,j,a}\), set
\begin{align*}
W_{t,j,a}
&=r_{\varrho}Z_{t,j,a}
+\sqrt{1-r_{\varrho}^2}\,E_{t,j,a}, \\
Y_{t,j,a}
&=\ell_a-\beta\log\bigl\{-\log\Phi(Z_{t,j,a})\bigr\},
&
\tilde Y_{t,j,a}
&=\ell_a-\beta\log\bigl\{-\log\Phi(W_{t,j,a})\bigr\}.
\end{align*}
The latent Gaussian correlation \(r_{\varrho}\) is numerically calibrated to
target Pearson correlation \(\varrho\) after the Gumbel transformations.
Both outcomes have marginal distribution
\(\mathrm{Gumbel}(\ell_a,\beta)\), mean
\(\ell_a+\gamma_{\mathrm E}\beta\), and variance \(\sigma^2\). The main
grid uses \(\sigma^2=1\). We additionally consider
\(\sigma^2\in\{1/3,3\}\) at \(b_tn=100\) as a variance ablation.

\paragraph{Bernoulli--Gumbel DGP.}
This mixed DGP allows the true potential outcomes and proxy outcomes to have
different marginal distributions and different arm rankings. Independently
for each arm, we draw
\begin{align*}
p_a &\sim \mathrm{Beta}(b_tn,b_tn),
&
\ell_a &\sim \mathrm{Gamma}(\text{shape}=b_tn,
\text{scale}=1/(b_tn)),
&
p_a &\perp \ell_a.
\end{align*}
Using independent standard normal variables \(Z_{t,j,a}\) and
\(E_{t,j,a}\), we then set
\begin{align*}
Y_{t,j,a}
&=\mathbf{1}\!\left\{Z_{t,j,a}>\Phi^{-1}(1-p_a)\right\}, \\
W_{t,j,a}
&=r_{\varrho,a}Z_{t,j,a}
+\sqrt{1-r_{\varrho,a}^2}\,E_{t,j,a}, \\
\tilde Y_{t,j,a}
&=\ell_a-\beta\log\bigl\{-\log\Phi(W_{t,j,a})\bigr\},
\end{align*}
where \(\beta=\sqrt{6}/\pi\). Because the Bernoulli threshold depends on
\(p_a\), the latent Gaussian correlation \(r_{\varrho,a}\) is numerically
calibrated separately for each arm to target
\(\operatorname{Corr}(Y_{t,j,a},\tilde Y_{t,j,a}\mid p_a,\ell_a)
=\varrho\). The resulting marginal distributions are
\(\mathrm{Bernoulli}(p_a)\) and \(\mathrm{Gumbel}(\ell_a,\beta)\), with
means \(p_a\) and \(\ell_a+\gamma_{\mathrm E}\beta\), respectively.
Because \(p_a\) and \(\ell_a\) are drawn independently, the ranking implied
by the proxy outcomes need not match the true arm ranking.

\paragraph{Experimental protocol.}
The first batch uses balanced uniform assignment for every method, with
arm counts differing by at most one. For the optimization-based policies,
TS, and TTTS, including their LLM-assisted versions, subsequent posterior
updates use realized allocation proportions. Uniform and uniform-LLM also
use realized allocation proportions in every posterior update, while keeping
their allocation probabilities uniform. Under the flat prior, uniform's
terminal posterior mean on the outcome scale equals the observed sample mean
for each arm.

For each configuration and each policy, we average results over 5,000
independent replications, using random seed 42 and 12 parallel workers. In
one replication, we generate the true potential outcomes and synthetic
proxy outcomes prescribed by the corresponding DGP and run the policy over
all ten batches. Posterior-based policies update their working posteriors
after each batch and recommend the arm with the largest terminal posterior
mean. Sequential Halving, TS-SPAS, and MLA-UCB use their respective
recommendation rules described below. The DGP-specific variances and
covariances are treated as known by policies that require them. At every
batch, relative simple regret is 100 times the ratio of the policy's mean
simple regret across replications to that of uniform allocation in the same
configuration and batch. Thus, uniform is 100 by construction, and smaller
values are better. The value after \(\pm\) in every table is the standard
error of the policy's mean simple regret, that is, its standard deviation
across replications divided by the square root of the number of
replications, expressed on the same scale as the entry, as a percentage of
the mean simple regret of uniform.

For RHO, both RHOPE variants, and the two myopic policies, we optimize the allocation on
the simplex through a direct softmax parameterization and Adam. We use 1,024
Sobol Monte Carlo draws, 300 optimization steps, learning rate 0.005, three
restarts, a simplex floor of \(10^{-6}\), and early stopping with patience 40
and minimum improvement \(10^{-5}\). Both components of the
LLM-assisted working posterior have zero initial precision, corresponding
to flat priors. The
TS-SPAS implementation uses a two-stage screening-and-allocation procedure;
its finite-support saddle solver has support size 32, 64 Monte Carlo draws,
12 inner steps, 12 outer steps, and tolerance 0.005.

\paragraph{Benchmark policies.}
The synthetic experiments compare the following policies.
\begin{itemize}
\item \textbf{uniform} uses equal allocation probabilities across arms and
the true-outcome-only posterior.
\item \textbf{TS} draws arm means from the true-outcome-only posterior and allocates
according to the resulting Thompson-sampling decision.
\item \textbf{TTTS} applies top-two Thompson sampling with top-two mixing
probability \(0.5\), again using the true-outcome-only posterior.
\item \textbf{myopic} chooses the next-batch allocation by optimizing a
one-step objective under the true-outcome-only posterior, without explicitly planning
over all remaining batches.
\item \textbf{RHO} uses the true-outcome-only posterior and the residual horizon
optimization objective of \citet{che2023adaptive} to select the allocation.
\item \textbf{Sequential Halving} is a non-LLM active-set elimination
method. It halves the active set over batch stages and retains at least two
active arms before recommending the active arm with the largest sample mean.
\item \textbf{TS-SPAS} is the non-LLM implementation of the two-stage
screening plus saddle-point allocation approach of \citet{kato2025minimax}.
It first screens arms and then computes a second-stage allocation and terminal
decision rule on the retained set.
\item \textbf{MLA-UCB} uses the machine-learning-assisted (MLA) mean estimator
of \citet{ji2025surrogate} and an upper-confidence-bound allocation rule
based on true and proxy outcomes. Its terminal recommendation maximizes the
MLA mean estimate.
\end{itemize}
The policies uniform-LLM, TS-LLM, TTTS-LLM, myopic-LLM, and \method{} are,
respectively, the LLM-assisted versions of uniform, TS, TTTS, myopic, and
RHO. They combine proxy outcomes with observed true outcomes through
finite-batch working-posterior versions of the updates in
Section~\ref{subsec:bayes}, with the allocation-proportion conventions above; the difference
from their non-LLM counterparts is therefore the information used for posterior
updates and the terminal recommendation, and, where applicable, allocation.

\subsubsection{Additional Synthetic Results}

\begin{table}[tbp]
\centering
\caption{Supplementary final relative simple regret results for the
Bernoulli--Gumbel synthetic setting. Each entry is a percentage relative to
uniform, which is 100 in every column; lower is better. Entries are mean
\(\pm\) standard error of the mean over 5,000 replications, both as
percentages of the mean simple regret of uniform. Boldface marks the
lowest value in each column. Dashes omit values for no-proxy methods that are
unchanged across \(\varrho\).}
\label{tab:synthetic-bernoulli-gumbel}
\small\setlength{\tabcolsep}{3pt}

\begin{tabular}{lrrrrrr}
\toprule
& \multicolumn{3}{c}{\(b_tn=100\)} & \multicolumn{3}{c}{\(b_tn=1000\)} \\
\cmidrule(lr){2-4}\cmidrule(lr){5-7}
Policy / \(\varrho\) & \(0.2\) & \(0.4\) & \(0.6\) & \(0.2\) & \(0.4\) & \(0.6\) \\
\midrule
\multicolumn{7}{l}{\underline{\text{Without proxy}}} \\
uniform & 100.00\sem{1.61} & -- & -- & 100.00\sem{1.61} & -- & -- \\
TS & 79.75\sem{1.43} & -- & -- & 82.64\sem{1.46} & -- & -- \\
TTTS & 79.08\sem{1.43} & -- & -- & 80.00\sem{1.43} & -- & -- \\
Sequential Halving & 76.12\sem{1.39} & -- & -- & 79.37\sem{1.40} & -- & -- \\
TS-SPAS & 112.11\sem{1.72} & -- & -- & 113.85\sem{1.72} & -- & -- \\
myopic & 74.55\sem{1.37} & -- & -- & 76.68\sem{1.40} & -- & -- \\
RHO & 74.66\sem{1.37} & -- & -- & 73.05\sem{1.37} & -- & -- \\
\addlinespace[1pt]
\multicolumn{7}{l}{\underline{\text{With proxy}}} \\
uniform-LLM & 100.84\sem{1.64} & 95.00\sem{1.59} & 85.98\sem{1.50} & 99.06\sem{1.59} & 93.26\sem{1.54} & 82.52\sem{1.43} \\
TS-LLM & 79.67\sem{1.41} & 71.62\sem{1.36} & 64.32\sem{1.28} & 78.88\sem{1.41} & 72.94\sem{1.35} & 62.64\sem{1.24} \\
TTTS-LLM & 76.87\sem{1.41} & 71.86\sem{1.33} & 60.62\sem{1.23} & 80.46\sem{1.44} & 72.09\sem{1.35} & 64.75\sem{1.28} \\
MLA-UCB & 97.30\sem{1.59} & 91.54\sem{1.55} & 82.17\sem{1.45} & 94.56\sem{1.57} & 88.51\sem{1.51} & 80.45\sem{1.41} \\
myopic-LLM & 73.05\sem{1.36} & 69.64\sem{1.30} & 61.91\sem{1.22} & 75.32\sem{1.37} & 69.21\sem{1.32} & 60.65\sem{1.21} \\
\delayedmethod{} & \textbf{72.19}\sem{1.33} & 65.76\sem{1.28} & 58.11\sem{1.18} & \textbf{71.99}\sem{1.34} & \textbf{66.62}\sem{1.28} & 58.32\sem{1.19} \\
\method{} & 73.21\sem{1.35} & \textbf{64.96}\sem{1.26} & \textbf{57.33}\sem{1.15} & 73.89\sem{1.37} & 66.69\sem{1.27} & \textbf{57.78}\sem{1.17} \\
\bottomrule
\end{tabular}
\end{table}

Table~\ref{tab:synthetic-bernoulli-gumbel} reports the main-grid results for
the Bernoulli--Gumbel setting. Table~\ref{tab:gumbel-variance} reports the
Gumbel outcome-variance ablations for (a) \(\sigma^2=1/3\) and
(b) \(\sigma^2=3\). Together with
Table~\ref{tab:synthetic} in the main text, these tables give the final,
tenth-batch relative simple regret for the synthetic configurations considered.

\begin{figure}[tbp]
\centering
\includegraphics[width=0.98\linewidth]{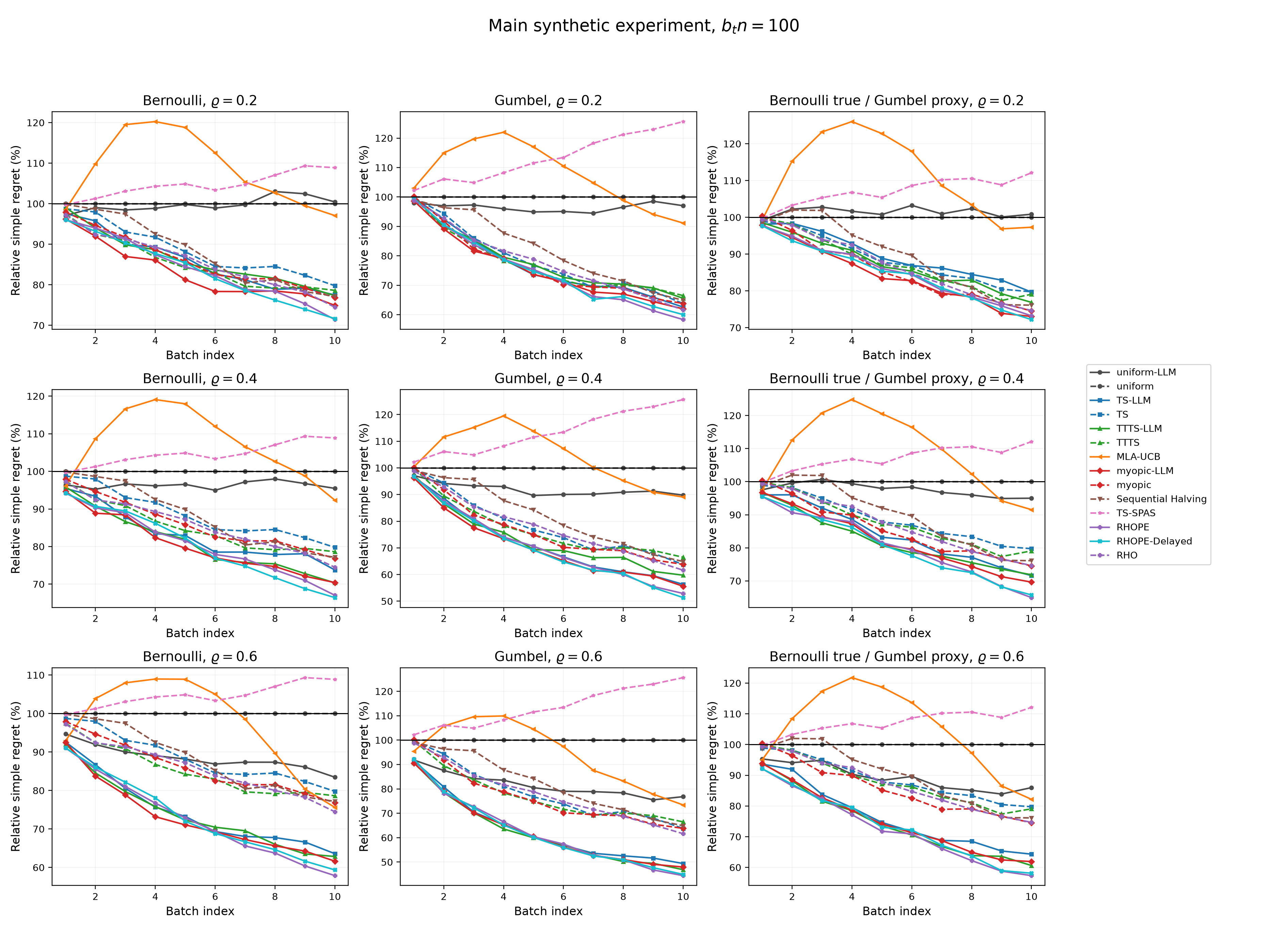}
\caption{Batch-level relative simple regret trajectories in the main synthetic
experiments with \(b_tn=100\). From top to bottom, the rows have
\(\varrho=0.2,0.4,0.6\); the columns report Bernoulli, Gumbel, and
Bernoulli--Gumbel results, respectively.}
\label{fig:synthetic-main-nt100}
\end{figure}

\begin{figure}[tbp]
\centering
\includegraphics[width=0.98\linewidth]{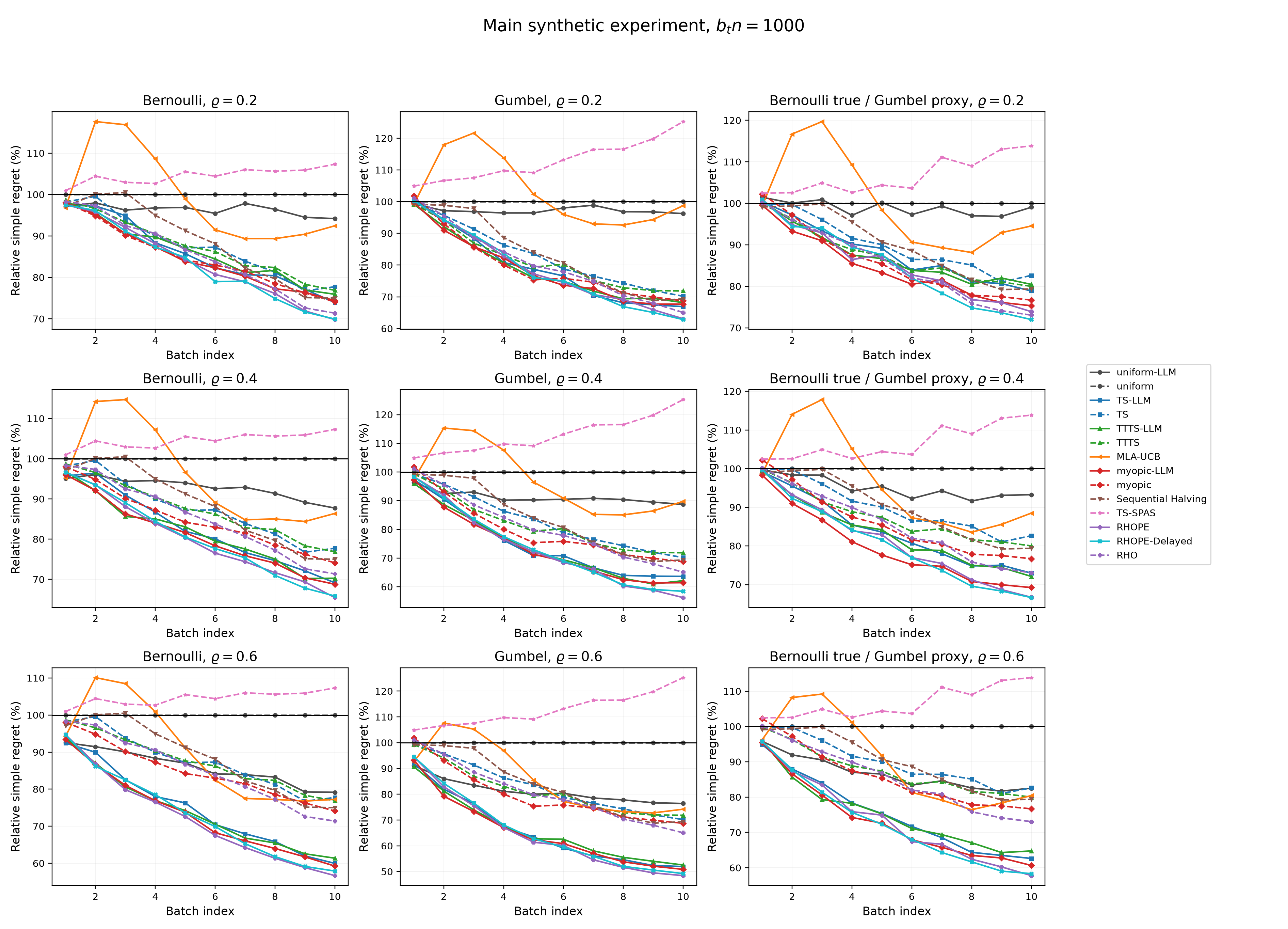}
\caption{Batch-level relative simple regret trajectories in the main synthetic
experiments with \(b_tn=1000\). From top to bottom, the rows have
\(\varrho=0.2,0.4,0.6\); the columns report Bernoulli, Gumbel, and
Bernoulli--Gumbel results, respectively.}
\label{fig:synthetic-main-nt1000}
\end{figure}

\begin{figure}[tbp]
\centering
\includegraphics[width=0.98\linewidth]{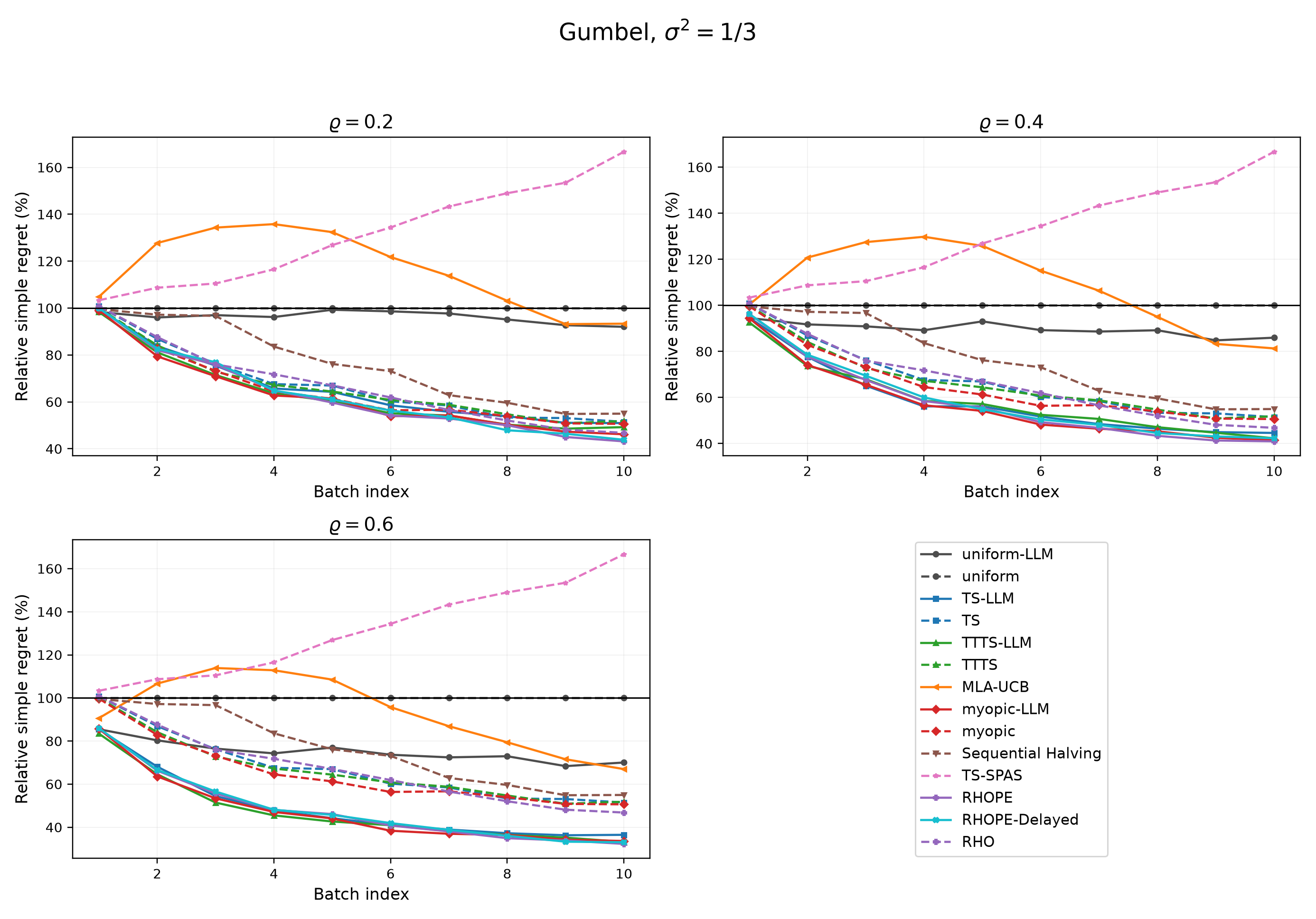}
\caption{Batch-level relative simple regret trajectories in the Gumbel
variance-ablation experiment with \(b_tn=100\) and \(\sigma^2=1/3\). The
three trajectory panels correspond to \(\varrho=0.2,0.4,0.6\); the lower-right
panel contains the shared legend.}
\label{fig:synthetic-var1over3}
\end{figure}

\begin{figure}[tbp]
\centering
\includegraphics[width=0.98\linewidth]{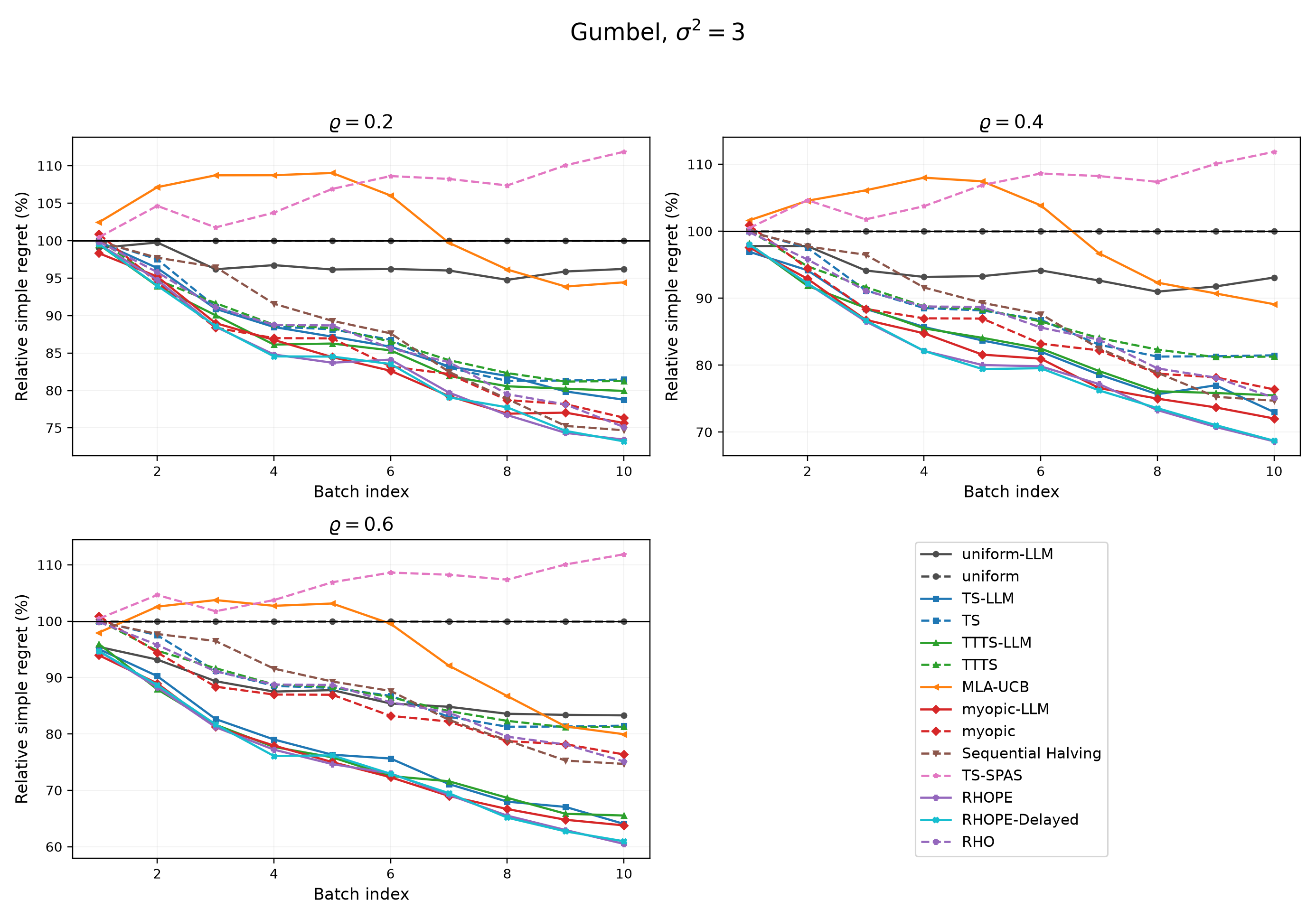}
\caption{Batch-level relative simple regret trajectories in the Gumbel
variance-ablation experiment with \(b_tn=100\) and \(\sigma^2=3\). The three
trajectory panels correspond to \(\varrho=0.2,0.4,0.6\); the lower-right
panel contains the shared legend.}
\label{fig:synthetic-var3}
\end{figure}

Figures~\ref{fig:synthetic-main-nt100}--\ref{fig:synthetic-var3}
show how relative simple regret evolves across the ten batches. Each curve is
normalized by uniform in the same configuration and batch, so values below
100 indicate lower simple regret than uniform at that point in the experiment.
In the main-grid figures, each image contains Bernoulli, Gumbel, and
Bernoulli--Gumbel panels. The variance-ablation figures contain the Gumbel
setting only.

\begin{table}[tbp]
\centering
\caption{Gumbel outcome-variance ablations for (a) \(\sigma^2=1/3\) and
(b) \(\sigma^2=3\). Entries are final simple regret percentages
relative to uniform, mean \(\pm\) standard error of the mean over 5,000
replications; lower is better. Boldface marks column minima;
dashes denote no-proxy results unchanged across \(\varrho\).}
\label{tab:gumbel-variance}
\small
\begin{minipage}{\textwidth}
\centering
\textbf{(a) \(\sigma^2=1/3\)}\par\vspace{1pt}
\begin{tabular}{lrrr}
\toprule
Policy / \(\varrho\) & \(0.2\) & \(0.4\) & \(0.6\) \\
\midrule
\multicolumn{4}{l}{\underline{\text{Without proxy}}} \\
uniform & 100.00\sem{2.46} & -- & -- \\
TS & 51.50\sem{1.62} & -- & -- \\
TTTS & 51.85\sem{1.65} & -- & -- \\
Sequential Halving & 54.99\sem{1.68} & -- & -- \\
TS-SPAS & 166.67\sem{3.56} & -- & -- \\
myopic & 50.64\sem{1.53} & -- & -- \\
RHO & 46.83\sem{1.47} & -- & -- \\
\addlinespace[1pt]
\multicolumn{4}{l}{\underline{\text{With proxy}}} \\
uniform-LLM & 92.02\sem{2.35} & 85.97\sem{2.25} & 70.05\sem{1.94} \\
TS-LLM & 50.97\sem{1.60} & 44.61\sem{1.43} & 36.49\sem{1.29} \\
TTTS-LLM & 49.24\sem{1.56} & 42.30\sem{1.39} & 33.19\sem{1.14} \\
MLA-UCB & 93.36\sem{2.39} & 81.31\sem{2.16} & 66.99\sem{1.89} \\
myopic-LLM & 46.20\sem{1.47} & 41.46\sem{1.35} & 33.63\sem{1.15} \\
\delayedmethod{} & 43.85\sem{1.40} & 42.32\sem{1.37} & 33.06\sem{1.15} \\
\method{} & \textbf{43.17}\sem{1.39} & \textbf{40.95}\sem{1.37} & \textbf{32.29}\sem{1.13} \\
\bottomrule
\end{tabular}

\end{minipage}\par\medskip
\begin{minipage}{\textwidth}
\centering
\textbf{(b) \(\sigma^2=3\)}\par\vspace{1pt}
\begin{tabular}{lrrr}
\toprule
Policy / \(\varrho\) & \(0.2\) & \(0.4\) & \(0.6\) \\
\midrule
\multicolumn{4}{l}{\underline{\text{Without proxy}}} \\
uniform & 100.00\sem{1.46} & -- & -- \\
TS & 81.46\sem{1.34} & -- & -- \\
TTTS & 81.27\sem{1.33} & -- & -- \\
Sequential Halving & 74.69\sem{1.25} & -- & -- \\
TS-SPAS & 111.86\sem{1.56} & -- & -- \\
myopic & 76.37\sem{1.26} & -- & -- \\
RHO & 75.12\sem{1.26} & -- & -- \\
\addlinespace[1pt]
\multicolumn{4}{l}{\underline{\text{With proxy}}} \\
uniform-LLM & 96.23\sem{1.43} & 93.08\sem{1.40} & 83.30\sem{1.31} \\
TS-LLM & 78.77\sem{1.30} & 72.98\sem{1.25} & 64.04\sem{1.16} \\
TTTS-LLM & 79.94\sem{1.30} & 75.48\sem{1.27} & 65.53\sem{1.17} \\
MLA-UCB & 94.44\sem{1.41} & 89.10\sem{1.37} & 79.95\sem{1.30} \\
myopic-LLM & 75.66\sem{1.27} & 72.03\sem{1.22} & 63.78\sem{1.15} \\
\delayedmethod{} & \textbf{73.20}\sem{1.25} & 68.71\sem{1.20} & 60.97\sem{1.13} \\
\method{} & 73.44\sem{1.24} & \textbf{68.61}\sem{1.20} & \textbf{60.50}\sem{1.12} \\
\bottomrule
\end{tabular}

\end{minipage}
\end{table}

\subsection{Semi-Synthetic Data Experiments}
\label{app:semi-synthetic-details}

The semi-synthetic experiment starts from the GDSC response matrix and its
cell-line covariates. The source data contain 896 cell lines, cell-line-level
responses for seven cancer drugs, and 18,522 features per cell line, including
mutation indicators, copy-number alterations, phenotype and tissue covariates,
and expression features. The raw response is the natural logarithm of the
\(\mathrm{IC}_{50}\) minus the drug's median over all screened cell lines, so
a smaller value corresponds to a more sensitive cell line. In the source
matrix, many entries are imputed values rather than measurements, and all
imputed entries of a drug carry the same value. Such entries distort the
correlation between true outcomes and proxy outcomes and would mask
the benefit that proxy outcomes bring to adaptive experimentation.
We therefore remove every cell line with an imputed entry, which leaves 649
cell lines. From these, 22 randomly selected cell lines form the pool of
in-context examples, and 16 randomly selected cell lines from this pool
serve as the in-context examples and as the training set of the non-LLM
predictors. The remaining 627 cell lines form the fixed semi-synthetic
instance with the seven drugs as \(K=7\) treatment arms.

For each experimental cell line \(c\), synthetic patient \(i\), and treatment
arm \(a\), we construct the true raw potential outcome as
\[
Y^{\mathrm{raw}}_{c,i,a}
=
Y^{\mathrm{raw}}_{c,a}+\eta_{c,i,a},
\]
where \(Y^{\mathrm{raw}}_{c,a}\) is the observed cell-line-level raw response
in GDSC and \(\eta_{c,i,a}\sim\Normal(0,4)\) is the synthetic patient noise
(\(\sigma=2\)). The adaptive policies operate on outcomes
\(Y_{c,i,a}=-Y^{\mathrm{raw}}_{c,i,a}\), so a larger outcome corresponds to
greater drug sensitivity. Thus, the patient-level variation is synthetic while
the cell-line profiles and the seven treatment-specific cell-line responses
remain anchored in the real dataset. Each of the 627 cell lines contributes
one patient to each of \(T=10\) batches, so every batch contains
\(b_tn=627\) patients and the experiment \(n=6{,}270\) patients in total.

\paragraph{Stronger LLM-generated proxies.}
We use Claude Opus 5.5 (API model ``claude-opus-5-5''), GPT-6.1 Sol
(``gpt-6.1-sol''), and DeepSeek V4 Pro (``deepseek-v4-pro'') to produce
proxy outcomes from one shared prompt. The system prompt defines the response
scale, lists the names, synonyms, targets, and pathways of the seven drugs,
describes the coverage of the profile features, instructs the model not to
consult external data and not to try to identify the cell line, and contains
the complete profiles and observed responses, rounded to two decimals, of the
16 in-context cell lines. The user prompt describes one cell line by its
cancer type, growth mode in culture, microsatellite status, the detected or
not-detected status of 18 drug-relevant mutations and copy-number
alterations (SMO is listed as not assessed for every cell line), its coding mutations in a panel of 300 cancer genes, its
copy-number gains and losses in 126 recurrent regions, each listed with the
genes it contains, and the percentiles, among all 896 cell lines, of the
baseline expression of 18 genes selected in advance for their relation to
the mechanisms of the seven drugs (EGFR, ERBB2, DUSP6, SPRY2, SPRY4, ETV4,
ETV5, PHLDA1, RB1, CDKN2A, CCND1, CCNE1, SLFN11, BRCA1, GLI1, PTCH1, CDH1,
and VIM). Within this coverage, an alteration that is not listed was not
detected. Profiles carry no cell-line identifier, and no response of an
experimental cell line appears in any prompt. Each request returns a JSON
object with the seven raw responses, enforced through structured outputs for
Claude and GPT and through JSON mode with validation for DeepSeek. Reasoning
is enabled, with effort ``high'' for Claude Opus 5.5 and GPT-6.1 Sol and
``low'' for DeepSeek V4 Pro, under an output limit of 65,536 tokens; no
response was truncated, and the sampling parameters are the API defaults.
The mean numbers of output tokens per request were 370 for Claude Opus 5.5,
1,772 for GPT-6.1 Sol (of which 1,693 reasoning tokens), and 7,136 for
DeepSeek V4 Pro (7,054 reasoning tokens). For every experimental cell line
we collect 20 independent draws, one request each, so that each model
produces \(627\times20=12{,}540\) parsed response vectors, and we use the
mean of the 20 raw responses, negated to the reward scale, as the cell line's
proxy vector. The following is the actual user prompt of one experimental
cell line.

\begin{center}
\setlength{\fboxsep}{6pt}
\fbox{\begin{minipage}{0.94\linewidth}
\raggedright
Cell-line profile:\par
- Cancer type: haematopoietic neoplasm other.\par
- Growth in culture: suspension.\par
- Microsatellite status: microsatellite stable.\par
- Key biomarker status: BRAF mutation: not detected; KRAS mutation: not
detected; NRAS mutation: not detected; HRAS mutation: detected; EGFR
mutation: not detected; EGFR region gain: not detected; ERBB2 mutation: not
detected; ERBB2 region gain: not detected; RB1 mutation: not detected; RB1
region loss: not detected; CDKN2A mutation: not detected; CDKN2A region
loss: not detected; CCND1 region gain: not detected; CCNE1 region gain: not
detected; BRCA1 mutation: not detected; BRCA2 mutation: not detected; TP53
mutation: not detected; PTCH1 mutation: not detected; SMO: not assessed.\par
- Genes with coding mutations: MLL2, HRAS, TET2.\par
- Copy-number gains affecting: none.\par
- Copy-number losses affecting: CASP1; SDHD; SYNCRIP, ZNF292; ATRX.\par
- Baseline mRNA expression percentile (0 = lowest, 100 = highest):\par
\hspace*{1em}EGFR 19; ERBB2 16; DUSP6 17; SPRY2 84; SPRY4 55; ETV4 44; ETV5 46;
PHLDA1 37; RB1 100; CDKN2A 41; CCND1 6; CCNE1 66; SLFN11 61; BRCA1 77;
GLI1 34; PTCH1 87; CDH1 25; VIM 57.\par
\par
Task: Estimate the response of this cell line to each of the seven
treatments, on the response scale defined in the system instructions. Return
only JSON with exactly the required raw\_responses keys.
\end{minipage}}
\end{center}

\paragraph{Weaker LLM-generated proxies.}
As a weaker LLM, DeepSeek V4 Flash (API model ``deepseek-v4-flash'')
generates proxies with reasoning disabled, temperature 1, a 256-token
output limit, and JSON mode. Its prompt describes a cell line only by 785
binary mutation, copy-number, phenotype, and tissue indicators: it contains
no expression features, no biomarker status lines, and no gene annotations
of the copy-number regions, and it identifies the cell line by a profile ID.
The system prompt (with the same seven drugs, the same response scale, and
the same 16 in-context cell lines as complete examples), the 627
experimental cell lines, the number of draws, and the averaging are
otherwise as above.

\paragraph{Non-LLM proxies.}
As references that use the same information and the same 16 labeled
in-context cell lines, ridge regression and gradient-boosted trees (GBT)
predict each drug's response from a numeric encoding of the prompt profile:
indicators for cancer type, growth mode, microsatellite status, each mutated
gene, and each copy-number region, together with the 18 expression
percentiles, standardized over the 896 cell lines (499 non-constant
features). Both predictors are fitted separately for each drug on the 16
in-context cell lines only, with hyperparameters selected by leave-one-out
cross-validation within these 16 lines: the ridge penalty from a grid of 25
values, and for GBT the tree depth in \(\{1,2\}\) and the number of trees in
\(\{50,100,200\}\), with learning rate 0.05, subsampling rate 0.8, and half of
the features considered per tree. Their predictions for the 627 experimental
cell lines, negated to the reward scale, are used exactly like the LLM
proxies.

\paragraph{Proxy moments and correlations.}
Every proxy is a deterministic function of the cell line, so the ten patients
of a cell line share its proxy vector, and the proxy carries no patient-level
variation. The known-moment policies use the moments of the fixed instance:
for arm \(a\), the outcome variance is the variance of
\(-Y^{\mathrm{raw}}_{c,a}\) across the 627 cell lines plus the noise variance
4, and the proxy variance and the proxy--outcome covariance are the
corresponding across-cell-line moments; the implied within-arm correlation
is \(\varrho_a=\operatorname{Corr}(Y_{c,i,a},\tilde Y_{c,a})\).
Table~\ref{tab:semi-synthetic-correlations} reports these correlations for
the six proxies. Averaged over arms, the stronger LLMs reach 0.307 (Claude
Opus 5.5), 0.290 (GPT-6.1 Sol), and 0.255 (DeepSeek V4 Pro), DeepSeek V4
Flash reaches 0.201, and GBT and ridge reach 0.188 and 0.164.

\begin{table}[tbp]
\centering
\caption{Arm-wise proxy--outcome correlations \(\varrho_a\) implied by the
known moments of the semi-synthetic instance (627 cell lines, patient-noise
variance 4). Arms 1--7 are cisplatin, vismodegib, palbociclib, trametinib,
dabrafenib, afatinib, and olaparib, respectively.}
\label{tab:semi-synthetic-correlations}
\small\setlength{\tabcolsep}{4pt}
\begin{tabular}{lrrrrrrrr}
\toprule
Proxy & Arm 1 & Arm 2 & Arm 3 & Arm 4 & Arm 5 & Arm 6 & Arm 7 & Mean \\
\midrule
Claude Opus 5.5 & 0.201 & 0.164 & 0.315 & 0.478 & 0.467 & 0.357 & 0.168 & 0.307 \\
GPT-6.1 Sol & 0.190 & 0.132 & 0.290 & 0.468 & 0.449 & 0.358 & 0.147 & 0.290 \\
DeepSeek V4 Pro & 0.186 & 0.056 & 0.251 & 0.424 & 0.433 & 0.311 & 0.125 & 0.255 \\
DeepSeek V4 Flash & 0.097 & 0.130 & 0.148 & 0.388 & 0.442 & 0.147 & 0.053 & 0.201 \\
GBT & 0.173 & 0.085 & 0.182 & 0.358 & 0.140 & 0.266 & 0.109 & 0.188 \\
Ridge & 0.078 & 0.108 & 0.191 & 0.356 & 0.160 & 0.203 & 0.053 & 0.164 \\
\bottomrule
\end{tabular}
\end{table}

We evaluate every policy in 10,000 replications with seed 42. Each
replication regenerates the patient noise, a balanced batch assignment that
places one patient of every cell line in each batch, and the policies'
internal randomness, while the cell-line responses and the proxies are fixed.
All runs with the same moment setting (known or estimated) use the same replications, so comparisons
between policies and between proxies are paired, and the results of the
non-LLM policies, whose inputs do not depend on the proxy, are reported once.
As in the synthetic experiments, the first batch is uniformly assigned,
posterior-based policies recommend the arm with the largest terminal
posterior mean, the other baselines retain the recommendation rules
described above, and subsequent optimization-based updates use realized
allocation proportions.

\section{Ablation Experiments with Estimated Moments}
\label{app:unknown-moments}

The known-moment conditions in Sections~\ref{subsec:synthetic}
and~\ref{subsec:semi} supply policies with the true arm-level variances and
covariances. In the ablations reported here, these moments are instead
estimated from experimental data, for both the synthetic and the GDSC
experiments.
The first batch remains a forced-uniform batch, so
neither the unknown moments nor their estimates affect its allocation. After
observing that batch, we construct arm-specific plug-in estimates, recompute
the first posterior update with those estimates, and hold them fixed for all
subsequent posterior updates and allocation decisions.

Specifically, let \(\mathcal I_{0,a}=\{j:A_{0,j}=a\}\) and
\(N_{0,a}=|\mathcal I_{0,a}|\ge2\). The barred quantities below are the
sample means over the indicated assigned or all-unit samples. We estimate the true-outcome variance from the
assigned observations, the proxy variance from all units in
the batch, and their covariance from the paired assigned observations:
\begin{align*}
\hat s_{a,Y}^2
&=\frac{1}{N_{0,a}-1}\sum_{j\in\mathcal I_{0,a}}
\left(Y_{0,j,a}-\bar Y_{0,a}^{\mathrm{obs}}\right)^2,\\
\hat s_{a,\tilde Y}^2
&=\frac{1}{b_0n-1}\sum_{j=1}^{b_0n}
\left(\tilde Y_{0,j,a}-\bar{\tilde Y}_{0,a}^{\mathrm{all}}\right)^2,\\
\hat c_{a,Y\tilde Y}
&=\frac{1}{N_{0,a}-1}\sum_{j\in\mathcal I_{0,a}}
\left(Y_{0,j,a}-\bar Y_{0,a}^{\mathrm{obs}}\right)
\left(\tilde Y_{0,j,a}-\bar{\tilde Y}_{0,a}^{\mathrm{obs}}\right).
\end{align*}
We then replace each displayed variance estimate by its maximum with
\(10^{-6}\), retaining the same notation. To ensure a valid plug-in covariance
matrix, we project the covariance so that the implied correlation has an
absolute value of at most \(0.999\). Denoting the projected covariance by
\(\hat c_{a,Y\tilde Y}^{\mathrm{proj}}\), the orthogonal-decomposition
parameters are
\begin{align*}
\hat\lambda_a
&=\frac{\hat c_{a,Y\tilde Y}^{\mathrm{proj}}}
{\hat s_{a,\tilde Y}^2},
&
\hat s_{a,\res}^2
&=\max\left\{
\hat s_{a,Y}^2-
\frac{(\hat c_{a,Y\tilde Y}^{\mathrm{proj}})^2}
{\hat s_{a,\tilde Y}^2},10^{-6}
\right\}.
\end{align*}
The retained non-LLM policies require only \(\hat s_{a,Y}^2\). MLA-UCB, Sequential
Halving, and TS-SPAS do not use the oracle moments replaced in this exercise,
so this ablation reports the remaining 11 policies. All relative simple
regret values continue to use uniform as the 100\% reference.

\subsection{Synthetic Experiments}
\label{app:unknown-synthetic}

The ablation uses the three main DGPs, \(K=10\), \(T=10\),
\(b_tn\in\{100,1000\}\), and
\(\varrho\in\{0.2,0.4,0.6\}\). Each policy--configuration pair uses 5,000
replications. Table~\ref{tab:unknown-synthetic} reports the terminal results;
no oracle moments are used in the recomputed first-batch posterior update or
in subsequent batches.

{\small\setlength{\tabcolsep}{3pt}%
\begin{longtable}{lrrrrrr}
\caption{Final relative simple regret in the synthetic experiments with
moments estimated from the first batch: (a) Bernoulli, (b) Gumbel with
\(\sigma^2=1\), and (c) Bernoulli--Gumbel settings. Each entry is a percentage
relative to uniform, mean \(\pm\) standard error of the mean over 5,000
replications; lower is better. Boldface marks the lowest value in each
column when that value is displayed. Dashes omit values for no-proxy methods
that are unchanged across \(\varrho\).}\label{tab:unknown-synthetic}\\
\endfirsthead
\multicolumn{7}{c}{Table~\thetable{} (continued)}\\
\endhead
\bottomrule
\endfoot
\multicolumn{7}{c}{\textbf{(a) Bernoulli setting}}\\

\toprule
& \multicolumn{3}{c}{\(b_tn=100\)} & \multicolumn{3}{c}{\(b_tn=1000\)} \\
\cmidrule(lr){2-4}\cmidrule(lr){5-7}
Policy / \(\varrho\) & \(0.2\) & \(0.4\) & \(0.6\) & \(0.2\) & \(0.4\) & \(0.6\) \\
\midrule
\multicolumn{7}{l}{\underline{\text{Without proxy}}} \\
uniform & 100.00\sem{1.60} & -- & -- & 100.00\sem{1.55} & -- & -- \\
TS & 78.57\sem{1.41} & -- & -- & 77.59\sem{1.39} & -- & -- \\
TTTS & 80.24\sem{1.42} & -- & -- & 74.73\sem{1.35} & -- & -- \\
myopic & 77.06\sem{1.36} & -- & -- & 73.42\sem{1.35} & -- & -- \\
RHO & \textbf{74.27}\sem{1.35} & -- & -- & 70.23\sem{1.31} & -- & -- \\
\addlinespace[1pt]
\multicolumn{7}{l}{\underline{\text{With proxy}}} \\
uniform-LLM & 104.40\sem{1.63} & 97.92\sem{1.57} & 86.72\sem{1.45} & 93.74\sem{1.51} & 88.16\sem{1.45} & 79.96\sem{1.38} \\
TS-LLM & 80.99\sem{1.44} & 77.07\sem{1.38} & 65.83\sem{1.26} & 77.19\sem{1.35} & 70.61\sem{1.29} & 61.10\sem{1.20} \\
TTTS-LLM & 82.25\sem{1.44} & 74.58\sem{1.37} & \textbf{65.63}\sem{1.26} & 74.77\sem{1.36} & 70.47\sem{1.31} & 62.32\sem{1.21} \\
myopic-LLM & 81.28\sem{1.45} & 77.65\sem{1.41} & 74.17\sem{1.39} & 73.71\sem{1.34} & 69.49\sem{1.29} & 59.36\sem{1.15} \\
\delayedmethod{} & 77.45\sem{1.40} & 75.96\sem{1.39} & 71.00\sem{1.36} & \textbf{68.28}\sem{1.27} & \textbf{63.75}\sem{1.22} & 57.94\sem{1.14} \\
\method{} & 77.85\sem{1.39} & 76.67\sem{1.40} & 69.30\sem{1.35} & 70.93\sem{1.29} & 64.91\sem{1.23} & \textbf{55.41}\sem{1.12} \\
\midrule

\multicolumn{7}{c}{\textbf{(b) Gumbel setting (\(\sigma^2=1\))}}\\

\toprule
& \multicolumn{3}{c}{\(b_tn=100\)} & \multicolumn{3}{c}{\(b_tn=1000\)} \\
\cmidrule(lr){2-4}\cmidrule(lr){5-7}
Policy / \(\varrho\) & \(0.2\) & \(0.4\) & \(0.6\) & \(0.2\) & \(0.4\) & \(0.6\) \\
\midrule
\multicolumn{7}{l}{\underline{\text{Without proxy}}} \\
uniform & 100.00\sem{1.85} & -- & -- & 100.00\sem{1.87} & -- & -- \\
TS & 67.62\sem{1.50} & -- & -- & 69.51\sem{1.49} & -- & -- \\
TTTS & 71.66\sem{1.54} & -- & -- & 72.68\sem{1.54} & -- & -- \\
myopic & 68.16\sem{1.49} & -- & -- & 69.01\sem{1.48} & -- & -- \\
RHO & \textbf{66.88}\sem{1.47} & -- & -- & 64.64\sem{1.44} & -- & -- \\
\addlinespace[1pt]
\multicolumn{7}{l}{\underline{\text{With proxy}}} \\
uniform-LLM & 104.05\sem{1.91} & 97.84\sem{1.84} & 85.70\sem{1.70} & 96.90\sem{1.84} & 89.56\sem{1.73} & 77.44\sem{1.57} \\
TS-LLM & 72.08\sem{1.55} & 67.74\sem{1.50} & 61.01\sem{1.42} & 69.52\sem{1.47} & 64.88\sem{1.41} & 53.81\sem{1.25} \\
TTTS-LLM & 72.94\sem{1.56} & 67.01\sem{1.50} & \textbf{60.04}\sem{1.41} & 69.65\sem{1.52} & 64.64\sem{1.43} & 54.31\sem{1.27} \\
myopic-LLM & 73.71\sem{1.59} & 74.54\sem{1.61} & 69.11\sem{1.58} & 66.55\sem{1.45} & 61.10\sem{1.39} & 53.53\sem{1.26} \\
\delayedmethod{} & 71.03\sem{1.55} & 72.09\sem{1.61} & 69.07\sem{1.60} & 63.94\sem{1.42} & \textbf{59.33}\sem{1.33} & 51.33\sem{1.22} \\
\method{} & 70.60\sem{1.52} & 69.22\sem{1.56} & 68.23\sem{1.55} & \textbf{62.94}\sem{1.43} & 59.60\sem{1.36} & \textbf{50.46}\sem{1.22} \\
\midrule

\multicolumn{7}{c}{\textbf{(c) Bernoulli--Gumbel setting}}\\

\toprule
& \multicolumn{3}{c}{\(b_tn=100\)} & \multicolumn{3}{c}{\(b_tn=1000\)} \\
\cmidrule(lr){2-4}\cmidrule(lr){5-7}
Policy / \(\varrho\) & \(0.2\) & \(0.4\) & \(0.6\) & \(0.2\) & \(0.4\) & \(0.6\) \\
\midrule
\multicolumn{7}{l}{\underline{\text{Without proxy}}} \\
uniform & 100.00\sem{1.61} & -- & -- & 100.00\sem{1.61} & -- & -- \\
TS & 79.36\sem{1.43} & -- & -- & 83.72\sem{1.46} & -- & -- \\
TTTS & 78.27\sem{1.41} & -- & -- & 80.40\sem{1.42} & -- & -- \\
myopic & \textbf{76.21}\sem{1.38} & -- & -- & 75.48\sem{1.38} & -- & -- \\
RHO & 76.47\sem{1.40} & -- & -- & 74.01\sem{1.39} & -- & -- \\
\addlinespace[1pt]
\multicolumn{7}{l}{\underline{\text{With proxy}}} \\
uniform-LLM & 105.86\sem{1.67} & 100.30\sem{1.63} & 88.91\sem{1.53} & 99.42\sem{1.60} & 94.31\sem{1.55} & 81.89\sem{1.43} \\
TS-LLM & 83.04\sem{1.47} & 77.57\sem{1.40} & 66.78\sem{1.28} & 79.93\sem{1.44} & 75.48\sem{1.40} & 63.78\sem{1.26} \\
TTTS-LLM & 80.25\sem{1.44} & 75.69\sem{1.40} & \textbf{64.93}\sem{1.27} & 78.70\sem{1.44} & 71.74\sem{1.35} & 64.13\sem{1.26} \\
myopic-LLM & 79.48\sem{1.44} & 77.23\sem{1.40} & 70.93\sem{1.34} & 75.89\sem{1.38} & 72.27\sem{1.35} & 60.12\sem{1.20} \\
\delayedmethod{} & 77.95\sem{1.42} & \textbf{73.40}\sem{1.37} & 68.90\sem{1.34} & 73.01\sem{1.35} & \textbf{67.82}\sem{1.29} & 59.07\sem{1.19} \\
\method{} & 79.01\sem{1.43} & 73.81\sem{1.38} & 69.48\sem{1.35} & \textbf{71.89}\sem{1.34} & 68.96\sem{1.31} & \textbf{57.92}\sem{1.19} \\
\end{longtable}}

Comparing Table~\ref{tab:unknown-synthetic} with the known-moment results in
Tables~\ref{tab:synthetic} and
\ref{tab:synthetic-bernoulli-gumbel} reveals a clear role for batch size. When
\(b_tn=1000\), the forced-uniform first batch provides approximately 100
observed outcomes per arm, enabling sufficiently accurate
plug-in estimates of the arm-level variances and covariances. Consequently,
either \delayedmethod{} or \method{} achieves the best performance in every
setting, as in the known-moment experiments. When \(b_tn=100\), however, the
first batch provides only approximately 10 observed outcomes per arm, making
the variance and covariance estimates substantially less accurate; other
methods then often outperform the two RHOPE variants. This
comparison indicates that strong practical performance from our methods when
the required moments are unknown depends on a finite-batch structure with
sufficiently large batches. This is precisely the regime targeted in
Section~\ref{sec:setup}. As explained there, updating the allocation policy can
entail nontrivial operational costs and engineering difficulties, particularly
in medical and platform experiments, so the number of reallocations is often
finite. In the target applications, this finite number of reallocations is
paired with many units in each batch, allowing data from the initial uniform batch
to support sufficiently accurate estimates of the required variances and covariances for
subsequent allocation decisions.

\subsection{Semi-Synthetic Experiment}
\label{app:unknown-semi}

We apply the same first-batch estimation procedure to the semi-synthetic GDSC
experiment with each of the six proxies. The fixed instance has seven arms,
ten batches, and 627 patients per batch; each run uses 10,000 replications.
The first batch therefore provides approximately 90 assigned true outcomes per
arm and all 627 proxy outcomes for each arm. Non-LLM results are identical
across proxies and are shown once. Table~\ref{tab:unknown-semi} reports the
terminal relative simple regret of the 11 policies retained in this ablation.

\begin{table}[tbp]
\centering
\caption{GDSC relative simple regret (\%) with moments estimated from the
first batch. Entries are mean \(\pm\) standard error of the mean over 10,000
replications, both as percentages of the mean simple regret of uniform;
lower is better. Boldface marks column minima among the displayed values.}
\label{tab:unknown-semi}
\fontsize{8}{9.6}\selectfont\setlength{\tabcolsep}{3pt}
\begin{tabular}{@{}lrrrrrr@{}}
\toprule
Policy / proxy & Claude Opus 5.5 & GPT-6.1 Sol & DeepSeek V4 Pro & DeepSeek V4 Flash & GBT & Ridge \\
\midrule
\multicolumn{7}{l}{\underline{\text{Without proxy}}} \\
uniform & 100.00\sem{1.87} & -- & -- & -- & -- & -- \\
TS & 50.00\sem{1.26} & -- & -- & -- & -- & -- \\
TTTS & 46.69\sem{1.23} & -- & -- & -- & -- & -- \\
myopic & 47.34\sem{1.22} & -- & -- & -- & -- & -- \\
RHO & 43.24\sem{1.15} & -- & -- & -- & -- & -- \\
\addlinespace[1pt]
\multicolumn{7}{l}{\underline{\text{With proxy}}} \\
uniform-LLM & 92.44\sem{1.73} & 91.91\sem{1.71} & 93.48\sem{1.75} & 94.65\sem{1.74} & 96.68\sem{1.83} & 96.89\sem{1.81} \\
TS-LLM & 50.33\sem{1.24} & 48.99\sem{1.25} & 52.24\sem{1.27} & 49.65\sem{1.24} & 51.30\sem{1.28} & 50.12\sem{1.25} \\
TTTS-LLM & 43.10\sem{1.17} & 44.07\sem{1.18} & 46.54\sem{1.19} & 45.36\sem{1.19} & 46.71\sem{1.21} & \textbf{45.38}\sem{1.19} \\
myopic-LLM & 45.64\sem{1.19} & 46.08\sem{1.20} & 46.79\sem{1.20} & 47.42\sem{1.20} & 49.36\sem{1.22} & 49.21\sem{1.24} \\
\delayedmethod{} & \textbf{40.14}\sem{1.14} & 40.52\sem{1.16} & 40.68\sem{1.15} & \textbf{41.45}\sem{1.17} & 44.12\sem{1.18} & 53.19\sem{1.16} \\
\method{} & 40.56\sem{1.16} & \textbf{39.94}\sem{1.15} & \textbf{40.65}\sem{1.15} & 41.65\sem{1.17} & \textbf{43.45}\sem{1.17} & 53.56\sem{1.17} \\
\bottomrule
\end{tabular}
\end{table}

\subsection{A Theoretical Complement: When Does a Plug-in Coefficient Help?}
\label{app:estimated-moments-theory}

Proposition~\ref{prop:estimation-gap} compares the regret guarantees of RHO
and \delayedmethod{} when the projection coefficient \(\lambda\) is known. In the
ablations of this appendix, \(\lambda\) is replaced by an estimate
\(\hat\lambda\), and the estimation error enters the posterior updates and
the terminal recommendation of \method{}. The following result compares the
two guarantees in this situation and gives the condition under which the
guarantee of \method{} with the estimated coefficient remains below that of
RHO. Its proof is in Appendix~\ref{app:proof-estimated-moments}.

We use the exchangeable-arm setting of Proposition~\ref{prop:estimation-gap}:
the moments \((s,\tilde s,\varrho)\) are common to all arms, so
\(\lambda=\varrho s/\tilde s\), the arm means are
\(h_a\overset{\mathrm{iid}}{\sim}\Normal(m_0,\tau^2)\) for some \(m_0\in\mathbb R\), and \(c_K\) is the
expected maximum of \(K\) independent standard normal variables. The
coefficient is estimated as \(\hat\lambda:=\hat c_{Y\tilde Y}/\tilde s^2\),
where \(\hat c_{Y\tilde Y}\) is the unbiased sample covariance of
\(n_\lambda\ge2\) i.i.d.\ pilot pairs \((Y^{\mathrm p},\tilde Y^{\mathrm p})\)
of a true and a proxy outcome, whose variances and correlation are
\((s^2,\tilde s^2,\varrho)\), as for the experimental noise, and whose
fourth moments are finite; the same \(\hat\lambda\) is used for all arms.
Let \(\mathrm{RHOPE}_{\hat\lambda}\) denote Algorithm~\ref{alg:rhope} with
\(\lambda_a\) replaced by \(\hat\lambda\) throughout, let
\(g(x):=1-(1+x)^{-1/2}\), and let
\(\gamma:=\E[(\epsilon^{\mathrm p})^2(\tilde\epsilon^{\mathrm p})^2]/(s^2\tilde s^2)\)
be the standardized fourth cross-moment of the centered pilot pairs
\((\epsilon^{\mathrm p},\tilde\epsilon^{\mathrm p})
:=(Y^{\mathrm p}-\E Y^{\mathrm p},\tilde Y^{\mathrm p}-\E\tilde Y^{\mathrm p})\).

\begin{assumption}[Plug-in coefficient]
\label{ass:plugin}
\begin{enumerate}
\item[(i)]\hypertarget{assumption.2}{}The unit-level noise \((\epsilon_{t,j,a},\tilde\epsilon_{t,j,a})\)
of the experiment is bivariate normal with the common moments \((s,\tilde s,\varrho)\),
independent across units and arms and of the parameters.
\item[(ii)] \(\mathrm{RHOPE}_{\hat\lambda}\) and RHO use the flat working
prior of Theorem~\ref{thm:ppi}, and within each batch the allocation is
implemented in balanced form: arm \(a\) receives exactly \(\rho_{t,a}b_tn\)
units, chosen at random independently of the noise.
\item[(iii)] The pilot pairs defining \(\hat\lambda\) are independent of
the experimental data and of \(h\).
\item[(iv)] \(\mathrm{RHOPE}_{\hat\lambda}\) and RHO are monotone in the
sense of Definition~\ref{def:monotone}.
\end{enumerate}
\end{assumption}

\begin{proposition}[Plug-in coefficient and the regret guarantee]
\label{prop:estimated-moments}
Under Assumption~\ref{ass:plugin},
\begin{align*}
\mathfrak R_T(\mathrm{RHOPE}_{\hat\lambda},\nu)
&\le\tau c_K\,
g\!\left(\frac{Ks^2}{\tau^2}\Bigl\{1-\Bigl(1-\frac1K\Bigr)
\Bigl[\varrho^2-\frac{\tilde s^2}{s^2}\,\E(\hat\lambda-\lambda)^2\Bigr]\Bigr\}\right),
\\
\mathfrak R_T(\RHO,\nu)&\le\tau c_K\,g\!\left(\frac{Ks^2}{\tau^2}\right),
\end{align*}
and the first bound is smaller than the second if and only if
\(\E(\hat\lambda-\lambda)^2<\lambda^2\), that is, if and only if
\[
\varrho^2
>\frac{\gamma-1}{n_\lambda}+\frac{1}{n_\lambda-1}
-\frac{(n_\lambda-2)\varrho^2}{n_\lambda(n_\lambda-1)}.
\]
If the pilot pairs are bivariate normal, then \(\gamma=1+2\varrho^2\) and
this condition reduces to \((n_\lambda-2)\varrho^2>1\), that is,
\(\varrho^2>1/(n_\lambda-2)\) for \(n_\lambda>2\).
\end{proposition}

Proposition~\ref{prop:estimated-moments} explains the pattern of the
ablations. Replacing the oracle coefficient by an estimate adds the
estimation error \(\E(\hat\lambda-\lambda)^2\) to the terminal uncertainty of
\method{}, so part of the precision gain from the proxies is lost; the
guarantee of \method{} stays below that of RHO only when the proxy--outcome
correlation exceeds a positive threshold. This is why the advantage of
\method{} over RHO in Table~\ref{tab:unknown-semi} is weaker than in
Table~\ref{tab:semi} for every LLM. The threshold is governed by the size of
the sample from which the coefficient is estimated, here the first batch:
the smaller the first batch, the higher the correlation required. This is
why, in Appendix~\ref{app:unknown-synthetic}, \method{} mostly falls behind RHO at
low correlations when the batch size is small, while it mostly keeps its
advantage when the batch size is large.

\section{Robustness to the Covariance Restrictions}
\label{app:covariance-robustness}

The posterior updates of \method{} rest on a working model that imposes two
restrictions on the joint covariance of the true and proxy outcomes
(Assumption~\ref{ass:gse}). This appendix restates the two restrictions
(Appendix~\ref{app:working-unrestricted}), explains why we impose them
(Appendix~\ref{app:working-motivation}), and verifies in the semi-synthetic
experiment that they change the posterior little
(Appendix~\ref{app:working-validation}).

\subsection{The Two Restrictions}
\label{app:working-unrestricted}

Let \(\epsilon_{t,j}\) and \(\tilde\epsilon_{t,j}\) be the \(K\)-dimensional
noise vectors of a unit, and let \(\Lambda=\diag\{\lambda_a\}_{a=1}^K\), as in
Assumption~\ref{ass:gse}. The covariance restrictions in
Assumption~\ref{ass:gse} are the following.
\begin{enumerate}
\item[(R1)] \emph{Diagonal proxy covariance:}
\(\Cov(\tilde\epsilon_{t,j})=\diag\{\tilde s_a^2\}_{a=1}^K\). The proxies of
different arms are uncorrelated.
\item[(R2)] \emph{Joint orthogonality:}
\(\Cov(\tilde\epsilon_{t,j},\epsilon_{t,j}-\Lambda\tilde\epsilon_{t,j})=0\).
The residual of each arm is uncorrelated with the proxies of all arms; for
the arm's own proxy this holds by the definition of \(\lambda_a\), so the
restriction concerns the proxies of the other arms.
\end{enumerate}
We call the model with (R1) and (R2) the \emph{working model}, and
the model with neither restriction the \emph{unrestricted model}.

\subsection{Motivation for the Working Model}
\label{app:working-motivation}

The posterior updates and the allocation objective of \method{} involve
moments of the true and proxy outcomes. These moments are unknown in
practice and have to be estimated from the first batch, as in
Appendix~\ref{app:unknown-moments}.

The number of moments to estimate depends on the model. The unrestricted
model requires the \(K\) true-outcome variances, the \(K(K+1)/2\) distinct
elements of the proxy covariance matrix, and the \(K^2\) covariances between
the true outcomes and the proxies, \(3K(K+1)/2\) moments in total. Under
(R1) and (R2), only three moments per arm remain: the variance of the true
outcome, the variance of the proxy, and their covariance, \(3K\) moments in
total. The number thus grows quadratically in \(K\) without the restrictions
and linearly with them.

The first batch is too small to estimate the \(3K(K+1)/2\) moments of the
unrestricted model accurately, because a covariance that involves the true
outcome of an arm can only be estimated from the units assigned to that arm.
Estimating so many moments from so few observations would overfit, and
inaccurate moments propagate into the posterior and the allocation, where
they can hurt the performance of \method{} instead of improving it.
Appendix~\ref{app:unknown-moments} shows that even the \(3K\) moments of the
working model are costly to estimate when the first batch is small.

We therefore adopt the working model to reduce the problem to a
low-dimensional one: we accept a small misspecification in exchange for
moments that can be estimated accurately, and hence for better finite-sample
performance of \method{}. Deliberately simplified covariance structures of
this kind are common in statistics. For example, discriminant analysis with
a diagonal covariance can outperform its counterpart with the full estimated
covariance when the dimension is large relative to the sample size
\citep{bickel2004some}.

\subsection{Validation of the Working Model}
\label{app:working-validation}

The synthetic experiments satisfy (R1) and (R2) by construction, so we
validate the working model in the semi-synthetic experiment of
Section~\ref{subsec:semi}, which violates both. For each of the six
proxies, we take the 10,000 replications of \method{} and, for the sampling trajectory of each replication,
compute the terminal joint posterior of the arm means under the working
model and under the unrestricted model. The unrestricted model uses the full
proxy covariance matrix and the full outcome--proxy covariance matrix of
the instance, computed across the
627 cell lines in the same way as the known moments of the working model
(Appendix~\ref{app:semi-synthetic-details}).

Table~\ref{tab:offdiag-moments} shows how far the instance is from the two
restrictions. On the correlation scale, the off-diagonal elements of the
proxy covariance matrix, which (R1) sets to zero, are 0.16--0.37 on average
in absolute value and reach 0.80--0.95, so (R1) is clearly violated. The
off-diagonal elements of the residual--proxy covariance matrix, which (R2)
sets to zero, are 0.03--0.06 on average and reach 0.09--0.20, so (R2) is
violated as well, although to a smaller extent.

\begin{table}[htbp]
\centering
\caption{Mean and maximum absolute off-diagonal elements of the proxy
covariance matrix \(\Cov(\tilde\epsilon_{t,j})\) and of the residual--proxy
covariance matrix
\(\Cov(\tilde\epsilon_{t,j},\epsilon_{t,j}-\Lambda\tilde\epsilon_{t,j})\) in
the semi-synthetic instance, on the correlation scale.}
\label{tab:offdiag-moments}
\small
\begin{tabular}{lrrrr}
\toprule
& \multicolumn{2}{c}{Proxy--proxy} & \multicolumn{2}{c}{Residual--proxy} \\
\cmidrule(lr){2-3}\cmidrule(lr){4-5}
Proxy & Mean & Max & Mean & Max \\
\midrule
Claude Opus 5.5 & 0.274 & 0.954 & 0.034 & 0.128 \\
GPT-6.1 Sol & 0.193 & 0.880 & 0.037 & 0.113 \\
DeepSeek V4 Pro & 0.160 & 0.886 & 0.042 & 0.095 \\
DeepSeek V4 Flash & 0.332 & 0.799 & 0.035 & 0.167 \\
GBT & 0.298 & 0.817 & 0.056 & 0.198 \\
Ridge & 0.370 & 0.799 & 0.040 & 0.131 \\
\bottomrule
\end{tabular}
\end{table}

Table~\ref{tab:posterior-comparison} compares the two posteriors over the
10,000 replications. The two posterior means of an arm differ by 0.07--0.11
posterior standard deviations on average. The posterior variance of an arm
under the unrestricted model is 1.0\%--2.5\% smaller on average than under
the working model, and at most 7.1\% smaller. The posterior correlation
between two arms, which the working model sets to zero, is 0.003--0.005 on
average and never exceeds 0.065. Hence, although both restrictions are violated, the violations have little effect on
the posterior.

\begin{table}[htbp]
\centering
\caption{Terminal posteriors of the arm means under the unrestricted model
(mean \(\mu^{\mathrm U}\), covariance \(V^{\mathrm U}\)) and the working
model (\(\mu^{\mathrm W}\), \(V^{\mathrm W}\)), over arms (or pairs of arms)
and the 10,000 replications. Mean difference:
\(|\mu_a^{\mathrm U}-\mu_a^{\mathrm W}|/(V_{aa}^{\mathrm W})^{1/2}\).
Variance ratio: \(V_{aa}^{\mathrm U}/V_{aa}^{\mathrm W}\). Correlation:
\(|V_{ab}^{\mathrm U}|/(V_{aa}^{\mathrm U}V_{bb}^{\mathrm U})^{1/2}\),
\(a\ne b\).}
\label{tab:posterior-comparison}
\small
\begin{tabular}{lrrrrrr}
\toprule
& \multicolumn{2}{c}{Mean difference} & \multicolumn{2}{c}{Variance ratio} & \multicolumn{2}{c}{Correlation} \\
\cmidrule(lr){2-3}\cmidrule(lr){4-5}\cmidrule(lr){6-7}
Proxy & Mean & 95\% & Mean & Min & Mean & Max \\
\midrule
Claude Opus 5.5 & 0.069 & 0.191 & 0.990 & 0.969 & 0.005 & 0.064 \\
GPT-6.1 Sol & 0.076 & 0.203 & 0.989 & 0.970 & 0.005 & 0.065 \\
DeepSeek V4 Pro & 0.083 & 0.216 & 0.987 & 0.972 & 0.004 & 0.061 \\
DeepSeek V4 Flash & 0.079 & 0.249 & 0.985 & 0.956 & 0.004 & 0.054 \\
GBT & 0.110 & 0.312 & 0.975 & 0.929 & 0.003 & 0.027 \\
Ridge & 0.093 & 0.250 & 0.983 & 0.952 & 0.003 & 0.028 \\
\bottomrule
\end{tabular}
\end{table}

\paragraph{Why the proxy covariance matters little.}
The posterior variance of an arm is the sum of a residual part and a proxy
part, \(s_{a,\res}^2/\kappa_a+\lambda_a^2\tilde s_a^2\), where \(\kappa_a\) is the
fraction of patients assigned to arm \(a\). The off-diagonal elements of the
proxy covariance matrix act only on the proxy part, and this part is small
for two reasons. First, the proxy explains only a small share of the outcome
variance: the correlation between an outcome and its proxy is at most 0.48
(Table~\ref{tab:semi-synthetic-correlations}), so
\(\lambda_a^2\tilde s_a^2=\varrho_a^2s_a^2\) is at most 23\% of \(s_a^2\).
Second, the proxy means are estimated from all patients, whereas the
residual means are estimated only from the patients assigned to the arm, so
the residual part is inflated by \(1/\kappa_a\) and the proxy part is not. As a
result, over the replications, the proxy part accounts for 0.6\%--2.5\% of
the posterior variance of an arm on average, and for at most 15\%. The
correlation between the proxies of two arms links only these small parts, so
the posterior correlation between the two arms stays small.

\paragraph{Why the residual--proxy covariance matters little.}
The off-diagonal elements of the residual--proxy covariance are small to
begin with (Table~\ref{tab:offdiag-moments}). The proxies of a cell line
come from the same model and the same profile, and the proxy of drug \(a\)
is already that model's prediction of the response to drug \(a\). The proxy
of another drug \(b\) is correlated with this response mainly through the
proxy of drug \(a\); once the latter is removed, the former has little left
to explain.

Taken together, these results show that the posterior of \method{} is robust
to violations of the two covariance restrictions in this experiment. The
working model avoids the overfitting that the unrestricted model would incur
when its moments are estimated, and it sacrifices little accuracy in the
posterior. We therefore regard it as a reasonable simplification of the
unrestricted model.

\section{Algorithms}
\label{app:rhope-algorithm}

The subscript \(n\) distinguishes finite-batch working-posterior quantities
from their Gaussian-limit counterparts. Algorithms~\ref{alg:rhope}--\ref{alg:rhope-delayed}
implement allocation rules derived within the Gaussian sequential experiment,
which provides their asymptotic framework. In that model, \(\pi_{t,a}\)
determines the residual posterior precision increment. In the finite-batch
updates below, \(\pi_t=\rho_t\) is the chosen allocation, whereas
\(\hat\pi_{t,a}\) is the realized proportion used to account for the actual
sample count. The vectors \(Z^J,Z^G\)
have mutually independent standard normal coordinates. Ties in the terminal
recommendation are broken by choosing the smallest arm index.

The displayed optimization steps assume proper current posteriors.
For the flat-prior experiments, both initial precisions and
precision-weighted means are zero. We replace the first optimization and
categorical assignment by balanced uniform assignment, with at least one
observation per arm, and accumulate the displayed update increments in
precision form. Optimization starts after this initialization, when the
working posteriors are proper. Experiments with estimated moments additionally use
the first-batch estimates in Appendix~\ref{app:unknown-moments}.

\begin{algorithmbox}{alg:rhope}{\method{}}
\textbf{Input.} Total budget \(n\); batch fractions \(b_0,\ldots,b_{T-1}\);
variances \(s_a^2,\tilde s_a^2\) and correlations \(\varrho_a\), \(a\in[K]\); prior means and standard
deviations
\((\boldsymbol\mu_0^{(1)},\boldsymbol\sigma_0^{(1)},
\boldsymbol\mu_0^{(2)},\boldsymbol\sigma_0^{(2)})\).

\textbf{Initialize.} Set the initial MDP state
\(
(\boldsymbol\mu_{n,0}^{(1)},\boldsymbol\sigma_{n,0}^{(1)},
\boldsymbol\mu_{n,0}^{(2)},\boldsymbol\sigma_{n,0}^{(2)})
=
(\boldsymbol\mu_0^{(1)},\boldsymbol\sigma_0^{(1)},
\boldsymbol\mu_0^{(2)},\boldsymbol\sigma_0^{(2)}).
\)

\begin{enumerate}
\item For \(t=0,\ldots,T-1\):
  \begin{enumerate}
  \item For every \(a\in[K]\), form the all-unit proxy aggregate estimator
  \(
  \hat J_{t,a}=\frac1{b_t\sqrt n}\sum_{j=1}^{b_tn}\tilde Y_{t,j,a},
  \)
  and update the proxy posterior by
  \[
  \begin{aligned}
  &(\sigma_{n,t+1,a}^{(2)})^{-2}
  =
  (\sigma_{n,t,a}^{(2)})^{-2}+\frac{b_t}{\tilde s_a^2},
  \\
  &\mu_{n,t+1,a}^{(2)}
  =
  (\sigma_{n,t+1,a}^{(2)})^2
  \left[
  (\sigma_{n,t,a}^{(2)})^{-2}\mu_{n,t,a}^{(2)}
  +\frac{b_t}{\tilde s_a^2}\hat J_{t,a}
  \right].
  \end{aligned}
  \]
  \item Let \(B_t:=\sum_{v=t}^{T-1}b_v\). Solve
  \[
  \begin{aligned}
  \rho_t
  &\in
  \argmax_{\bar\rho\in\DeltaK}
  \E_{Z^G,Z^J}\Bigg[
  \max_{a\in[K]}
  \Bigg\{
  \mu_{n,t,a}^{(1)}
  +
  \sqrt{
  \frac{(\sigma_{n,t,a}^{(1)})^4\bar\rho_a B_t}
  {s_{a,\res}^2+(\sigma_{n,t,a}^{(1)})^2\bar\rho_aB_t}
  }
  Z_a^G
  \\
  &\qquad
  +\lambda_a\mu_{n,t+1,a}^{(2)}
  +
  \lambda_a
  \sqrt{
  \frac{(\sigma_{n,t+1,a}^{(2)})^4B_{t+1}}
  {\tilde s_a^2+(\sigma_{n,t+1,a}^{(2)})^2B_{t+1}}
  }
  Z_a^J
  \Bigg\}
  \Bigg].
  \end{aligned}
  \]
  \item For every \(j=1,\ldots,b_tn\), assign
  \(A_{t,j}\sim\mathrm{Categorical}(\rho_t)\) and set
  \(\xi_{t,j,a}=\ind\{A_{t,j}=a\}\) for \(a\in[K]\).
\item For every \(a\in[K]\), update the residual posterior with the allocation-weighted residual aggregate estimator
  \(
  \hat G_{t,a}
  =
  \frac1{b_t\sqrt n}
  \sum_{j=1}^{b_tn}\xi_{t,j,a}
  \left(Y_{t,j,a}-\lambda_a\tilde Y_{t,j,a}\right)
  \):
  \[
  \begin{aligned}
  &(\sigma_{n,t+1,a}^{(1)})^{-2}
  =
  (\sigma_{n,t,a}^{(1)})^{-2}
  +\frac{\hat\pi_{t,a}b_t}{s_{a,\res}^2},
  \\
  &\mu_{n,t+1,a}^{(1)}
  =
  (\sigma_{n,t+1,a}^{(1)})^2
  \bigg[
  (\sigma_{n,t,a}^{(1)})^{-2}\mu_{n,t,a}^{(1)}
  +\frac{b_t}{s_{a,\res}^2}\hat G_{t,a}
  \bigg].
  \end{aligned}
  \]
  Here
  \(
  \hat\pi_{t,a}
  =
  \frac1{b_tn}\sum_{j=1}^{b_tn}\xi_{t,j,a}
  \)
  is the realized allocation proportion for arm \(a\) in batch \(t\).
  \end{enumerate}
\item Return
\[
\hat a_T\in\argmax_{a\in[K]}
\left\{\mu_{n,T,a}^{(1)}+\lambda_a\mu_{n,T,a}^{(2)}\right\}.
\]
\end{enumerate}
\end{algorithmbox}

\begin{algorithmbox}{alg:rhope-delayed}{\delayedmethod{}}
\textbf{Input.} Total budget \(n\); batch fractions \(b_0,\ldots,b_{T-1}\);
variances \(s_a^2,\tilde s_a^2\) and correlations \(\varrho_a\), \(a\in[K]\); prior means and standard
deviations
\((\boldsymbol\mu_0^{(1)},\boldsymbol\sigma_0^{(1)},
\boldsymbol\mu_0^{(2)},\boldsymbol\sigma_0^{(2)})\).

\textbf{Initialize.} Set the initial MDP state
\(
(\boldsymbol\mu_{n,0}^{(1)},\boldsymbol\sigma_{n,0}^{(1)},
\boldsymbol\mu_{n,0}^{(2)},\boldsymbol\sigma_{n,0}^{(2)})
=
(\boldsymbol\mu_0^{(1)},\boldsymbol\sigma_0^{(1)},
\boldsymbol\mu_0^{(2)},\boldsymbol\sigma_0^{(2)}).
\)

\begin{enumerate}
\item For \(t=0,\ldots,T-1\):
  \begin{enumerate}
  \item Let \(B_t:=\sum_{v=t}^{T-1}b_v\). Before using any proxy outcomes
  from batch \(t\), solve
  \[
  \begin{aligned}
  \rho_t
  &\in
  \argmax_{\bar\rho\in\DeltaK}
  \E_{Z^G,Z^J}\Bigg[
  \max_{a\in[K]}
  \Bigg\{
  \mu_{n,t,a}^{(1)}
  +\lambda_a\mu_{n,t,a}^{(2)}
  \\
  &\qquad
  +\sqrt{
  \frac{(\sigma_{n,t,a}^{(1)})^4\bar\rho_a B_t}
  {s_{a,\res}^2+(\sigma_{n,t,a}^{(1)})^2\bar\rho_a B_t}
  }Z_a^G
  +\lambda_a\sqrt{
  \frac{(\sigma_{n,t,a}^{(2)})^4 B_t}
  {\tilde s_a^2+(\sigma_{n,t,a}^{(2)})^2 B_t}
  }Z_a^J
  \Bigg\}
  \Bigg].
  \end{aligned}
  \]
  \item For every \(j=1,\ldots,b_tn\), assign
  \(A_{t,j}\sim\mathrm{Categorical}(\rho_t)\) and set
  \(\xi_{t,j,a}=\ind\{A_{t,j}=a\}\) for \(a\in[K]\).
  \item Once the allocation is fixed, incorporate the current batch's
  all-arm proxy outcomes. For every \(a\in[K]\), form
  \(
  \hat J_{t,a}=\frac1{b_t\sqrt n}\sum_{j=1}^{b_tn}\tilde Y_{t,j,a}
  \)
  and update the proxy posterior by
  \[
  \begin{aligned}
  &(\sigma_{n,t+1,a}^{(2)})^{-2}
  =
  (\sigma_{n,t,a}^{(2)})^{-2}+\frac{b_t}{\tilde s_a^2},
  \\
  &\mu_{n,t+1,a}^{(2)}
  =
  (\sigma_{n,t+1,a}^{(2)})^2
  \left[
  (\sigma_{n,t,a}^{(2)})^{-2}\mu_{n,t,a}^{(2)}
  +\frac{b_t}{\tilde s_a^2}\hat J_{t,a}
  \right].
  \end{aligned}
  \]
  \item For every \(a\in[K]\), use the assigned true outcomes and their paired
  proxies to form the residual aggregate estimator
  \(
  \hat G_{t,a}
  =
  \frac1{b_t\sqrt n}
  \sum_{j=1}^{b_tn}\xi_{t,j,a}
  \left(Y_{t,j,a}-\lambda_a\tilde Y_{t,j,a}\right)
  \)
  and update the residual posterior by
  \[
  \begin{aligned}
  &(\sigma_{n,t+1,a}^{(1)})^{-2}
  =
  (\sigma_{n,t,a}^{(1)})^{-2}
  +\frac{\hat\pi_{t,a}b_t}{s_{a,\res}^2},
  \\
  &\mu_{n,t+1,a}^{(1)}
  =
  (\sigma_{n,t+1,a}^{(1)})^2
  \bigg[
  (\sigma_{n,t,a}^{(1)})^{-2}\mu_{n,t,a}^{(1)}
  +\frac{b_t}{s_{a,\res}^2}\hat G_{t,a}
  \bigg].
  \end{aligned}
  \]
  Here
  \(\hat\pi_{t,a}=\frac1{b_tn}\sum_{j=1}^{b_tn}\xi_{t,j,a}\)
  is the realized allocation proportion.
  \end{enumerate}
\item Return
\[
\hat a_T\in\argmax_{a\in[K]}
\left\{\mu_{n,T,a}^{(1)}+\lambda_a\mu_{n,T,a}^{(2)}\right\}.
\]
\end{enumerate}
\end{algorithmbox}

\clearpage
\end{document}